\documentclass[10pt,twocolumn,letterpaper]{article}
\usepackage{iccv}              
\usepackage{array}
\usepackage{float}
\usepackage{arydshln} 
\usepackage{amsmath}  
\usepackage{xcolor} 
\usepackage{caption}

\definecolor{iccvblue}{rgb}{0.21,0.49,0.74}
\usepackage[pagebackref,breaklinks,colorlinks,allcolors=iccvblue]{hyperref}
\usepackage{graphicx}
\usepackage{amsthm}
\usepackage{multirow}
\usepackage{ulem}
\usepackage{float}
\usepackage{cuted}
\def\paperID{12879} 
\def\confName{ICCV}
\def\confYear{2025}

\title{LOCATEdit: Graph Laplacian Optimized Cross Attention for Localized Text-Guided Image Editing}

\author{
Achint Soni$^{1}$\quad Meet Soni$^{2}$\quad Sirisha Rambhatla$^{1}$\\
$^{1}$University of Waterloo\quad $^{2}$Stony Brook University\\
\\
{\tt\small \{a2soni, sirisha.rambhatla\}@uwaterloo.ca, meet.soni@stonybrook.edu}
}

\renewcommand{\arraystretch}{0.5} 
\theoremstyle{plain} 
\newtheorem{theorem}{Theorem}

\newtheorem{lemma}{Lemma}

\def\method{LOCATEdit\xspace}

\begin{document}
\maketitle
\begin{strip}
    \centering
    \includegraphics[width=\textwidth]{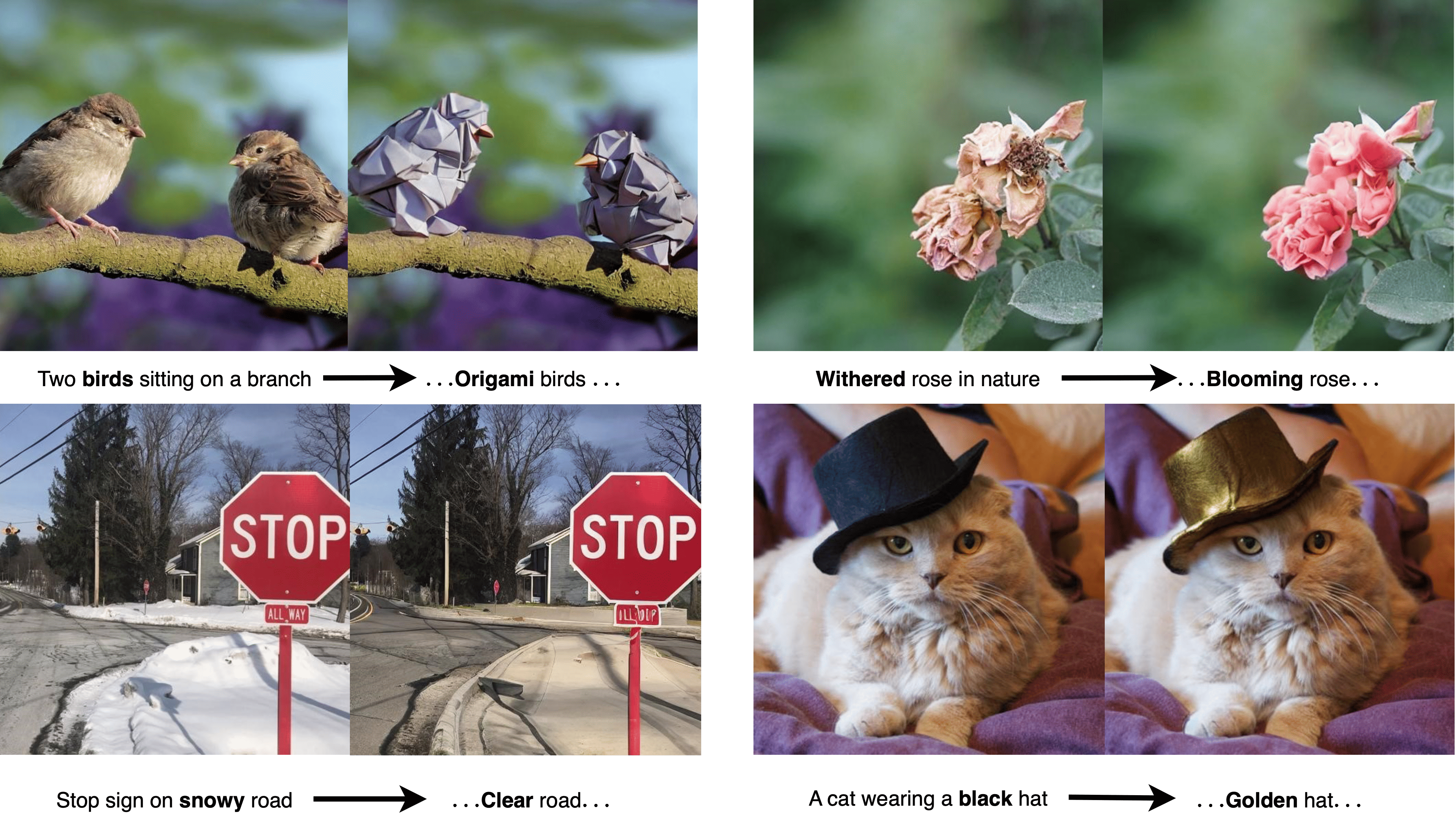}
    \captionsetup{hypcap=false} 
    \captionof{figure}{Our \method demonstrates strong performance on various complex image editing tasks.
    \label{fig:feature-graphic}}
\end{strip}

\begin{abstract}


Text-guided image editing aims to modify specific regions of an image according to natural language instructions while maintaining the general structure and the background fidelity. Existing methods utilize masks derived from cross-attention maps generated from diffusion models to identify the target regions for modification. However, since cross-attention mechanisms focus on semantic relevance, they struggle to maintain the image integrity. As a result, these methods often lack spatial consistency, leading to editing artifacts and distortions.  In this work, we address these limitations and introduce \textbf{\method}, which enhances cross-attention maps through a graph-based approach utilizing self-attention-derived patch relationships to maintain smooth, coherent attention across image regions, ensuring that alterations are limited to the designated items while retaining the surrounding structure. \method consistently and substantially outperforms existing baselines on PIE-Bench, demonstrating its state-of-the-art performance and effectiveness on various editing tasks. Code can be found on \url{https://github.com/LOCATEdit/LOCATEdit/}

\end{abstract}

\vspace{-10pt}
\section{Introduction}

Diffusion models have become popular for image generation, yet practical applications demand precise control for editing. Text-guided editing techniques \cite{yang2023dynamic, wang2024vision, cao2023masactrl} have emerged as powerful tools to facilitate such modifications across domains, from digital art \cite{huang2024creativesynth, gao2023textpainter, ko2023large, sun2024anycontrol} to medical imaging \cite{wang20243d, nafi2024diffusion}, enabling more intuitive image manipulation through natural language prompts. However, prompt-driven editing is often imprecise \cite{ju2024pnp, brack2024ledits++, couairon2022diffedit}.

To attain precise control in text-guided image editing, recent studies use masks derived from cross-attention maps; however, inaccuracies in these maps can result in edits spilling over unintended regions, causing problems such as object identity loss \cite{kohstructure, shen2024rethinking, wang2023mdp} and background drift \cite{liu2024towards, yang2023dynamic}.  Because of this, techniques that depend exclusively on cross-attention could make global changes when only localized modifications are required \cite{hertz2022prompt, ju2024pnp}. These problems highlight the necessity for a method that precisely identifies editing areas without jeopardizing the integrity of the overall image.

\begin{figure}
    \centering
    \includegraphics[width=\linewidth]{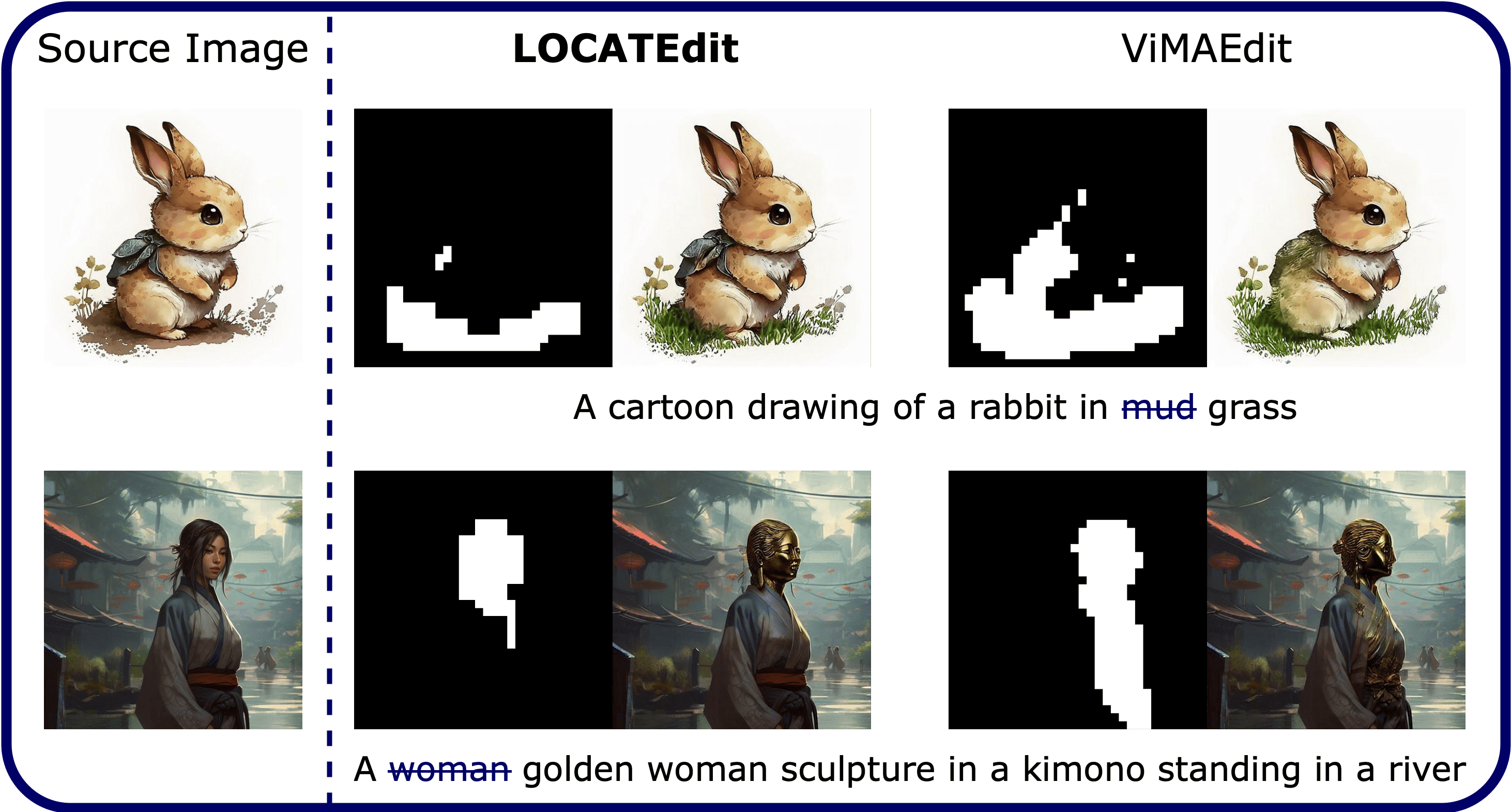}
    \caption{Example of over-editing caused due to imprecise masks.}
    \label{fig:over-editing}
\end{figure}

Recent methods have demonstrated improved mask accuracy through the utilization of cross- and self-attention masks, while simultaneously adopting the dual branch editing paradigm \cite{wang2024vision, yang2023dynamic}.  Additionally, they incorporate target image embeddings as auxiliary guidance derived from source image embeddings and the editing information contained in source-target prompt pairs. Despite these, challenges such as unintended spills continue to be a problem, which can be seen in Figure \ref{fig:over-editing}. Our key observation is that naively combining cross-attention and self-attention results in significant information loss. Consequently, we propose to induce spatial consistency and precise identification of regions to be edited via a graph-based approach. Given graphs \(\mathcal{G}_{\text{src}}\) and \(\mathcal{G}_{\text{tgt}}\) for the source and target branch, respectively, each of these graphs is constructed using the respective Cross and Self-Attention, hence CASA graphs, encoding cross-attention maps as nodes and self-attention relationships as weighted edges. With this abstraction, these graphs intrinsically depict the structure of the image, thereby connecting local and global contexts, while also maintaining the semantic relevance.

We explicitly enforce graph structure by proposing a graph Laplacian regularizer on $\mathcal{G}_{src}$ and $\mathcal{G}_{tgt}$ to impose spatial consistency, motivated by the effectiveness of Laplacian regularization in image denoising and mesh editing \cite{pang2017graph, sorkine2004laplacian}.  Prior works on segmentation and spatial regularization \cite{uchinoura2022graph, zeng2019deep} also demonstrate that this Laplacian constraint effectively maintains object boundaries and preserves local detail, thus \textit{disentangling} the areas of interest from unrelated regions. Furthermore, \citet{belkin2003laplacian} and \citet{lim2024graph} illustrate that this regularization also enhances the separation of semantic characteristics. By integrating a Laplacian smoothness factor into the diffusion process, \method optimizes the attention values across interconnected patches without any additional training, hence reducing background drift and limiting global changes as can be seen in Figure \ref{fig:feature-graphic}. This ensures that modifications are confined to designated areas while preserving the overall structural integrity of the original image. Notably, this optimization admits a closed-form solution, hence eliminating the need for iterative refinement \cite{wang2024vision}.
Overall, our contributions can be summarized as follows:
\begin{itemize}
    \item \textbf{CASA Graph:} We introduce \textbf{\method}, a method which encapsulates word-to-pixel relevance through pixel-to-pixel relationships by modeling attention maps with CASA graph.
    \item \textbf{Improved spatial consistency:} By optimizing masks through graph Laplacian regularization on the CASA graph, we maintain object structure and confine changes to intended regions, hence minimizing distortions.
    \item \textbf{Disentangled and faithful editing:} Leveraging Laplacian smoothing, \method achieves precise semantic modifications while preserving the original image context, ensuring disentangled editing.
\end{itemize}

\section{Related Work}

Recent advances in image editing have leveraged a range of conditioning modalities—including text, reference images, and segmentation maps—to drive semantic, structural, and stylistic modifications \cite{huang2024diffusion}. In this work, we focus specifically on text-guided image editing with an emphasis on preserving the original content and ensuring effective foreground-background disentanglement.

\subsection{Text-guided Image Editing}
Early methods exploited the power of CLIP \cite{pmlr-v139-radford21a} to align images and text. For example, \cite{Kim_2022_CVPR} fine-tuned diffusion models during reverse diffusion using a CLIP loss to adjust image attributes, though these approaches were limited to global changes and often suffered from degraded image quality. Later works such as DiffuseIT \cite{kwon2022diffusion} and StyleDiffusion \cite{Wang_2023_ICCV} improved performance by introducing semantic or style disentanglement losses; however, they are computationally expensive and typically confined to specific style modifications. More recent frameworks like InstructPix2Pix \cite{Brooks_2023_CVPR} preserve source content using text instructions, yet require carefully curated instruction-image pair datasets and supervised training. Additionally, methods that manipulate text embeddings for disentangled editing have been explored \cite{wu2023uncovering}, though they often yield only marginal improvements over earlier approaches.

Collectively, these studies underscore both the promise and limitations of text-guided editing, motivating our work on refining attention maps to achieve spatially consistent and localized modifications.

\subsection{Training Free Image Editing}

Recent advances in text-to-image synthesis \cite{gal2022image, nichol2021glide, ramesh2022hierarchical, rombach2022high, saharia2022photorealistic} have enabled high-quality photorealistic image generation from text prompts. Building on these advances, several studies have proposed dual-branch, training-free approaches that leverage rich feature and attention maps from pre-trained diffusion models for image editing. These methods exploit signals from the source image’s diffusion process to drive content modification, obviating the need for additional model training while achieving remarkable success in altering image content.

Notably, PRedITOR \cite{ravi2023preditor} generates a target CLIP embedding via a diffusion prior model but struggles with fine detail and background consistency. Other methods enhance structural control: P2P \cite{hertz2022prompt} replaces cross-attention maps to maintain spatial alignment, and PnP \cite{tumanyan2023plug} injects spatial features and self-attention maps into decoder layers. Approaches like MasaCtrl \cite{cao2023masactrl} preserve structure through mutual self-attention, while editing-area grounding techniques and attention regularization losses are employed in DPL \cite{yang2023dynamic} and refined further in ViMAEdit \cite{wang2024vision}. Despite these advances, challenges in achieving precise localization and consistent edits persist, motivating our work.

\subsection{Graph Laplacian}

In optimization and semi-supervised learning, Laplacian regularization promotes smooth variation along a graph, similar to how Conditional Random Fields (CRFs) refine segmentation by enforcing spatial and color consistency \cite{lafferty2001conditional}. Unlike CRFs, Laplacian smoothing is fully differentiable and easily integrated into neural networks. Its effectiveness has been demonstrated in tasks such as image matting \cite{levin2007closed}, where the matting Laplacian preserves edges while interpolating unknown regions, and in action localization \cite{park2019graph}, where it refines class activation maps for more coherent predictions.

\section{Background}

\begin{figure*}
    \centering
    \includegraphics[width=\linewidth]{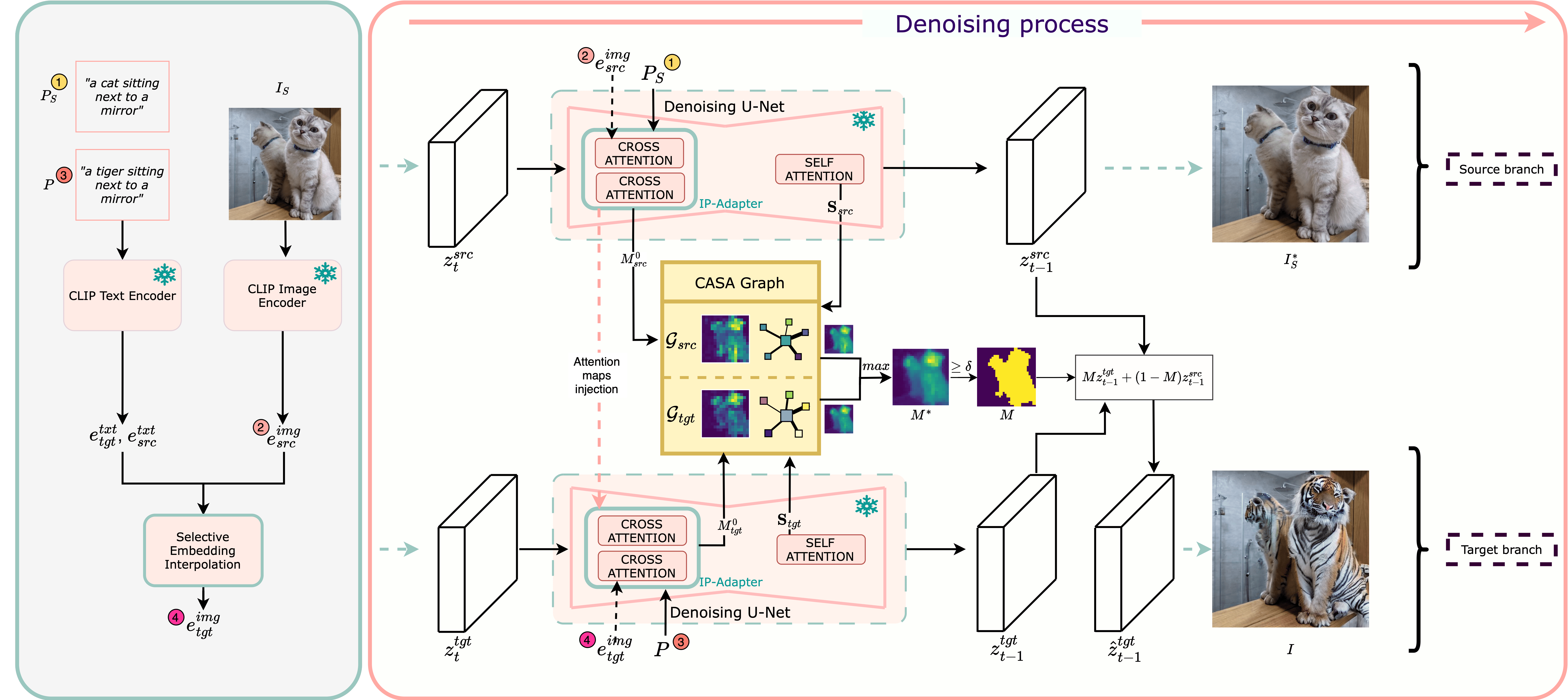}
    \caption{\textbf{Overview of our text-guided image editing pipeline.} \method refines cross-attention maps with graph Laplacian regularization for spatial consistency, uses an IP-Adapter for additional guidance, and employs selective pruning on text embeddings to suppress noise, ensuring the edited image preserves key structural details.}
    \label{fig:pipeline}
\end{figure*}
\subsection{Diffusion models}
Diffusion models \cite{ho2020denoising, song2020denoising, nichol2021glide} constitute a class of generative approaches that operate through two 
complementary processes---forward and backward diffusion. In the forward 
diffusion process, starting from an original clean sample $\mathbf{z}_{0}$ (drawn from the 
data distribution), Gaussian noise is iteratively added at each timestep $t=1,2,\dots,T$. 
Specifically, one obtains
\[
\mathbf{z}_{t} = \sqrt{\alpha_{t}} \,\mathbf{z}_{0} \;+\; \sqrt{1 - \alpha_{t}} \,\boldsymbol{\epsilon}_{t}, 
\quad t = 1,\ldots,T,
\]
where $\boldsymbol{\epsilon}_{t} \sim \mathcal{N}(\mathbf{0}, \mathbf{I})$ is an independent 
Gaussian noise term injected at timestep $t$. The sequence $\{\alpha_{t}\}_{t=1}^{T}$ governs 
the noise variance at each stage, ensuring that after $T$ diffusion steps, $\mathbf{z}_{T}$ 
is approximately distributed as a standard Gaussian.

The backward diffusion process reverses this corruption procedure by progressively 
denoising the noisy sample $\mathbf{z}_{T}$ into a cleaner sample $\mathbf{z}_{T-1}$, then 
$\mathbf{z}_{T-2}$, and so forth, converging to a final clean reconstruction $\mathbf{z}_{0}$. 
To accomplish this, one samples $\mathbf{z}_{t-1}$ from a conditional distribution over 
$\mathbf{z}_{t}$, typically parameterized by a learnable denoising function. Formally, 
the update rule may be expressed as
\[
\mathbf{z}_{t-1} 
= 
\boldsymbol{\mu}_{t}\bigl(\mathbf{z}_{t},\theta\bigr)
+ 
\sigma_{t}\,\tilde{\boldsymbol{\epsilon}}_{t},
\quad t = T, \ldots, 1,
\]
where $\tilde{\boldsymbol{\epsilon}}_{t}$ is a random Gaussian 
noise , $\boldsymbol{\mu}_{t}$ and $\sigma_{t}$ represents the mean and variance of distribution that $\mathbf{z_{t-1}}$ can be sampled from, and 
$\theta$ encapsulates the learned parameters. In the DDIM formulation \cite{song2020denoising}, 
one often employs a deterministic variant by modifying the variance schedule, making the sampling 
process more efficient while maintaining high sample quality.

A pivotal component in modern diffusion models is the noise prediction network 
$\boldsymbol{\epsilon}_{\theta}(\mathbf{z}_{t}, t)$. Rather than predicting $\mathbf{z}_{0}$ 
or $\mathbf{z}_{t-1}$ directly, the network estimates the noise present in the corrupted sample 
$\mathbf{z}_{t}$. Once trained, this noise predictor effectively guides the reverse diffusion 
steps to iteratively remove the injected Gaussian noise.

\subsection{Attention mechanism}

In practice, the noise prediction model is frequently instantiated by a U-Net architecture, chosen for its efficacy in pixel-level prediction tasks. 
Each U-Net block typically consists of (i)~a residual convolutional sub-block that refines the spatial representation of the intermediate feature maps, and (ii)~a self-attention sub-block that captures long-range patch-to-patch dependencies. (iii) cross-attention sub-block that aligns the image to textual information

In the mechanism, feature tensors are first projected into three distinct embeddings---queries $Q$, keys $K$, and 
values $V$. Attention is computed as
\[
\text{Attention}(Q, K, V) 
= 
\text{Softmax}\Bigl(\frac{QK^{\top}}{\sqrt{d}}\Bigr)\,V,
\]
where $d$ is the dimensionality of the query/key vectors, In both self-attention and cross-attention layers, $Q$ is projected from spatial features. In self-attention, $K$ and $V$ also come from spatial features, whereas in cross-attention, they are projected from textual embeddings. These projections use learned metrics that are optimized during training.

\section{\method}
In this section, we present \method for precise, localized text-guided image editing that refines the cross-attention maps. Our approach integrates two complementary modules. First,  we utilize the CASA graph to impose spatial coherence and ensure that edits are restricted to the designated areas. Second, building upon previous work \cite{wang2024vision}, we integrate an image embedding-enhanced denoising process augmented by a selective pruning operator applied to the text embedding offsets. This operator eliminates minor semantic variations, therefore minimizing unwanted changes and avoiding unnecessary editing of non-target regions. Together, these modules allow \method\ to maintain the structural integrity of the original image while precisely implementing the desired edits.

\subsection{Dual-Branch Editing Paradigm}

Our pipeline employs a dual-branch design in which a source branch reconstructs the original image and a target branch generates the edited output. To maintain structural consistency, both branches start from the same initial noise \(\mathbf{z}_T\) and share intermediate latent variables. Crucially, we inject the cross-attention maps from the source branch into the target branch \cite{hertz2022prompt} to maintain the spatial structure. Formally, if \(Q^{\text{src}}\) and \(K^{\text{src}}\) are the query and key embeddings from the source branch and \(V^{\text{tgt}}\) denotes the value embeddings from the target branch, then the target cross-attention is computed as
\[
\text{Attention}\bigl(Q^{\text{src}}, K^{\text{src}}, V^{\text{tgt}}\bigr)
= \text{Softmax}\Bigl(\frac{Q^{\text{src}} (K^{\text{src}})^\top}{\sqrt{d}}\Bigr)\,V^{\text{tgt}}.
\]

\begin{figure*}[t]
    \centering
    \includegraphics[width=\linewidth]{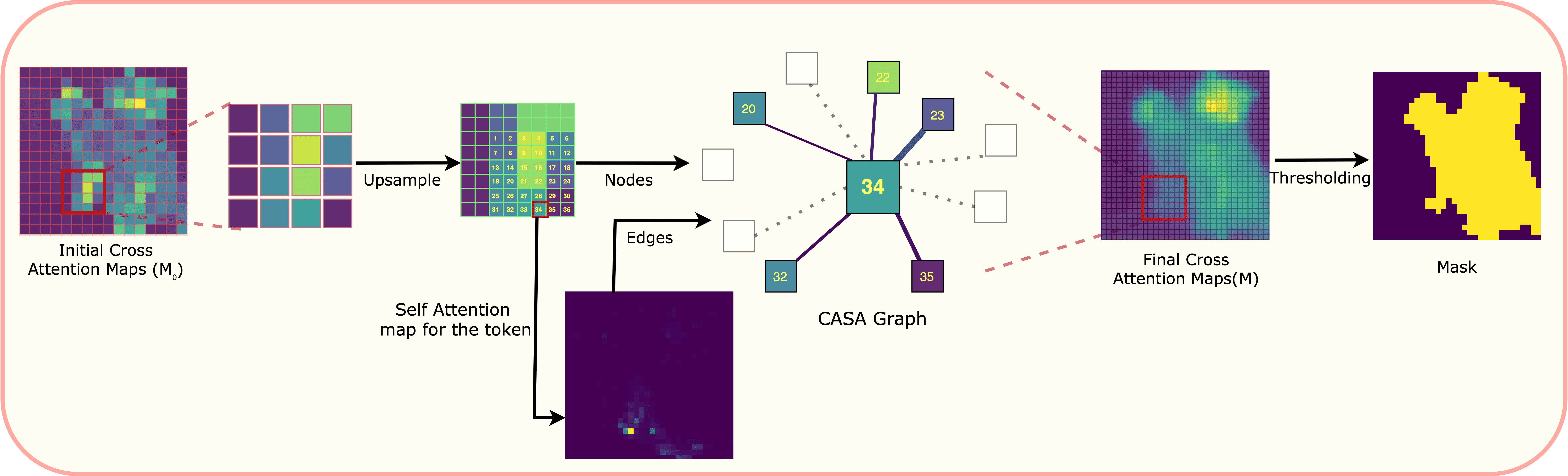}
    \caption{C\textbf{ASA (Cross and Self-Attention) Graph Construction workflow.} The initial cross-attention maps are upsampled to form a patch-level adjacency graph, then Laplacian regularization enforces spatial consistency. Thresholding the refined maps yields final, more robust attention masks.}
    \label{fig:graph_laplacian}
\end{figure*}

\subsection{Selective Embedding Interpolation}

Following previous work \cite{wang2024vision}, we employ an IP-Adapter \cite{ye2023ip} to provide explicit guidance for target image generation. After extracting the source image embedding \(e_{\text{src}}^{img}\) and the CLIP-based text embeddings \(e_{\text{src}}^{txt}\) and \(e_{\text{tgt}}^{txt}\) corresponding to the source and target prompts respectively, the conventional target image embedding is computed as
\begin{equation}\label{eq:embedding_inter}
    e_{\text{tgt}}^{img} = e_{\text{src}}^{img} + \Bigl(e_{\text{src}}^{txt} - e_{\text{tgt}}^{txt}\Bigr).
\end{equation}
This embedding is then processed by the IP-Adapter, which projects it into a latent feature space that is integrated into the diffusion model’s cross-attention mechanism. Specifically, given the query \(Q\) (derived from the noisy latent), the IP-Adapter produces additional key and value features \(K^{\text{IP}}\) and \(V^{\text{IP}}\) from the projected target embedding. These are then combined with the original key \(K\) and value \(V\) features to form the final cross-attention:
\[
 Z = \text{Attention}(Q, K, V)  + \lambda\text{Attention}(Q, K^\text{IP}, V^\text{IP}) 
\]
thereby incorporating semantic guidance into the diffusion process without requiring any additional training.\\

A limitation of directly using the difference \(e_{\text{src}}^T - e_{\text{tgt}}^T\) in Equation~\eqref{eq:embedding_inter} is that low-magnitude components, inherent in the entangled nature of CLIP text embeddings \cite{materzynska2022disentangling}, can lead to unintended edits. To mitigate this, we introduce a selective pruning operator \(\mathcal{H}\) that thresholds the text difference, retaining only the dominant semantic offsets. Formally, we replace Equation~\eqref{eq:embedding_inter} with
\begin{equation}\label{eq:pruned_embedding}
    e_{\text{tgt}}^I = e_{\text{src}}^I + \mathcal{H}\Bigl(e_{\text{src}}^T - e_{\text{tgt}}^T\Bigr),
\end{equation}
where \(\mathcal{H} : \mathbb{R}^d \to \mathbb{R}^d\) is defined elementwise as
\begin{equation}\label{eq:pruning_function}
\bigl[\mathcal{H}(\mathbf{y})\bigr]_i =
\begin{cases}
y_i, & \text{if } |y_i| \ge \tau, \\[6pt]
0,   & \text{otherwise}.
\end{cases}
\end{equation}
Here, \(d\) is the embedding dimension and \(\tau\) is determined via a percentile threshold on the absolute values of the difference. This selective pruning ensures that only significant semantic shifts contribute to the target image embedding, thereby reducing the risk of global edits and preserving the structural consistency of non-target regions. The pruned embedding is then processed through the IP-Adapter as described above, ensuring that the final diffusion process is both semantically guided and robust to minor, spurious variations.

\subsection{Formulating CASA Graph}
While the IP-Adapter provides explicit semantic guidance, the cross-attention maps extracted during denoising may still contain spills that lead to unintended edits. To address this, we refine these attention maps by modeling them as a CASA graph, where each node represents an image patch and the edges capture patch-to-patch relationships obtained from self-attention as can be seen in Figure \ref{fig:graph_laplacian}. The graph Laplacian regularization enforces a smoothness constraint across the CASA graph, penalizing abrupt differences in attention between strongly connected patches. In effect, this smoothing suppresses isolated high responses that can cause over-editing, ensuring that only spatially coherent regions receive significant modifications. By harmonizing the attention values over connected patches, \method robustly confines edits to the intended regions and preserves the overall spatial consistency.

Formally, within each U-Net block, each prompt word is linked to a cross-attention map; however, only the cross-attention maps related to the blend word(s) are necessary.  Following previous studies \cite{chefer2023attend, yang2023dynamic, hertz2022prompt}, we compute the average of the cross-attention maps obtained from multiple U-Net blocks to get initial maps. We obtain initial attention maps for both the source and target branches, denoted as \(M_0^{\text{src}} \in \mathbb{R}^{r\times r}\) and \(M_0^{\text{tgt}} \in \mathbb{R}^{r\times r}\). These masks are then upsampled to a higher resolution of \(R\times R\) (where \(R=\gamma r\) and \(\gamma>1\)) to capture fine spatial details, and subsequently flattened to yield the initial saliency maps \(\mathbf{m}_0^{\text{src}},\, \mathbf{m}_0^{\text{tgt}} \in \mathbb{R}^{R^2}\).

To prioritize high-confidence regions, we compute a weight for each patch by applying the sigmoid function \(\sigma(\cdot)\) to the corresponding element of \(\mathbf{m}_0^{\text{src}}\) and then squaring the output. Squaring the sigmoid output emphasizes larger values while further suppressing lower ones, thereby enhancing the reliability of high-confidence regions. These weights are assembled into a diagonal confidence matrix with a scaling factor \(\alpha\):

\begin{equation}\label{eq:sigmoid_weights} 
\mathbf{\Lambda}^{\text{src}} = \operatorname{diag}\Bigl(\sigma\Bigl(\alpha \mathbf{m}_0[1]\Bigr)^2,\, \dots,\, \sigma\Bigl(\alpha\mathbf{m}_0[R^2]\Bigr)^2\Bigr).
\end{equation}

and similarly for \(\mathbf{\Lambda}^{\text{tgt}}\).

Next, we extract self-attention maps \(\mathbf{S}^{\text{src}} \in \mathbb{R}^{R^2 \times R^2}\) and \(\mathbf{S}^{\text{tgt}} \in \mathbb{R}^{R^2 \times R^2}\) for the source and target branches, respectively. To ensure mutual relationships are treated uniformly and to guarantee the convexity of the optimization, we symmetrize both the maps as
\begin{equation}\label{eq:symmetric_q}
    \mathbf{S}_{\mathrm{sym}} = \frac{1}{2}\Bigl(\mathbf{S} + \mathbf{S}^\top\Bigr).
\end{equation}

Now, for each branch we construct CASA graph \(\mathcal{G} = (V, E)\) where each node \(v_i \in V\) corresponds to a patch in the flattened saliency map \(\mathbf{m}_0\). The edge weight between nodes \(v_i\) and \(v_j\) is given by the symmetrized self-attention map \(\mathbf{S}_{\mathrm{sym}}\). This graph structure, with nodes representing the initial saliency values and edges capturing inter-patch relationships, serves as the foundation for the CASA graph.

\subsection{Graph Laplacian Regularization}
After initializing CASA graphs \(\mathcal{G}_{\text{src}}\) and \(\mathcal{G}_{\text{tgt}}\) for both branches, we optimize for the value of their nodes using graph Laplacian optimization.

Formally, graph Laplacian is defined by:
\[
\mathbf{L} = \mathbf{D} - \mathbf{S}_{\mathrm{sym}}.
\]
where \(\mathbf{D}\) is a degree matrix for \(\mathbf{S}_{\mathrm{sym}}\), which is computed as
\[
\mathbf{D}(i,i)=\sum_{j=1}^{R^2} \mathbf{S}_{\mathrm{sym}}(i,j), \quad \mathbf{D}(i,j)=0 \quad \text{for } i\neq j,
\]

\begin{lemma}
The graph Laplacian \(\mathbf{L} \in \mathbb{R}^{R^2\times R^2}\) is positive semidefinite.
\end{lemma}
Detailed proof is provided in Appendix~\ref{proof:lemma1}.\\

We then optimize the initial saliency maps \(\mathbf{m}_0^{\text{src}}\text{ and } \mathbf{m}_0^{\text{tgt}}\)for both branches through the following convex optimization problem:
\begin{theorem}
Let \(\mathbf{m}_0 \in \mathbb{R}^{R^2}\) be the initial saliency map, and let \(\mathbf{\Lambda}\) and \(\mathbf{L}\) be defined as above. The optimal saliency map \(\mathbf{m}^* \in \mathbb{R}^{R^2}\) is the unique minimizer of
\[
J(\mathbf{m}) = (\mathbf{m} - \mathbf{m}_0)^\top \mathbf{\Lambda} (\mathbf{m} - \mathbf{m}_0) + \lambda\, \mathbf{m}^\top L\, \mathbf{m},
\]
with the solution
\[
\mathbf{m}^* = \left(\mathbf{\Lambda} + \lambda\, L\right)^{-1}\mathbf{\Lambda}\,\mathbf{m}_0.
\]
\end{theorem}
Detailed proof is provided in Appendix~\ref{proof:theorem1}.

The refined saliency maps \(\mathbf{m}^{*\text{src}}\) and \(\mathbf{m}^{*\text{tgt}}\) are then reshaped back to \(M^{*src} \text{ and } M^{*tgt}\), which are then used to obtain \(M^*\) by taking the element-wise maximum of the two maps:
\[
M^* = \max\{M_{\text{src}}^*,\, M_{\text{tgt}}^*\},
\]
Thresholding \(M^*\) with \(\delta\) gives the final spatial mask \(M\). An optimized CASA graph enforces a smooth, spatially consistent mask that preserves high-confidence regions and mitigates over-editing in less reliable areas.

Moreover, to maintain background consistency and prevent unintended changes outside the editing region, this optimized mask is used to replace the target branch's latent representation at each denoising step:
\[
\hat{\mathbf{z}}_{t-1}^{\text{tgt}} = M \odot \mathbf{z}_{t-1}^{\text{tgt}} \;+\; (1 - M) \odot \mathbf{z}_{t-1}^{\text{src}}, \quad t = T,\dots,1.
\]
Here, \(\odot\) denotes Hadamard product. This step ensures that the background and non-editable regions of the source image remain unchanged throughout the iterative denoising process.

\begin{figure}
    \centering
    \includegraphics[width=\linewidth]{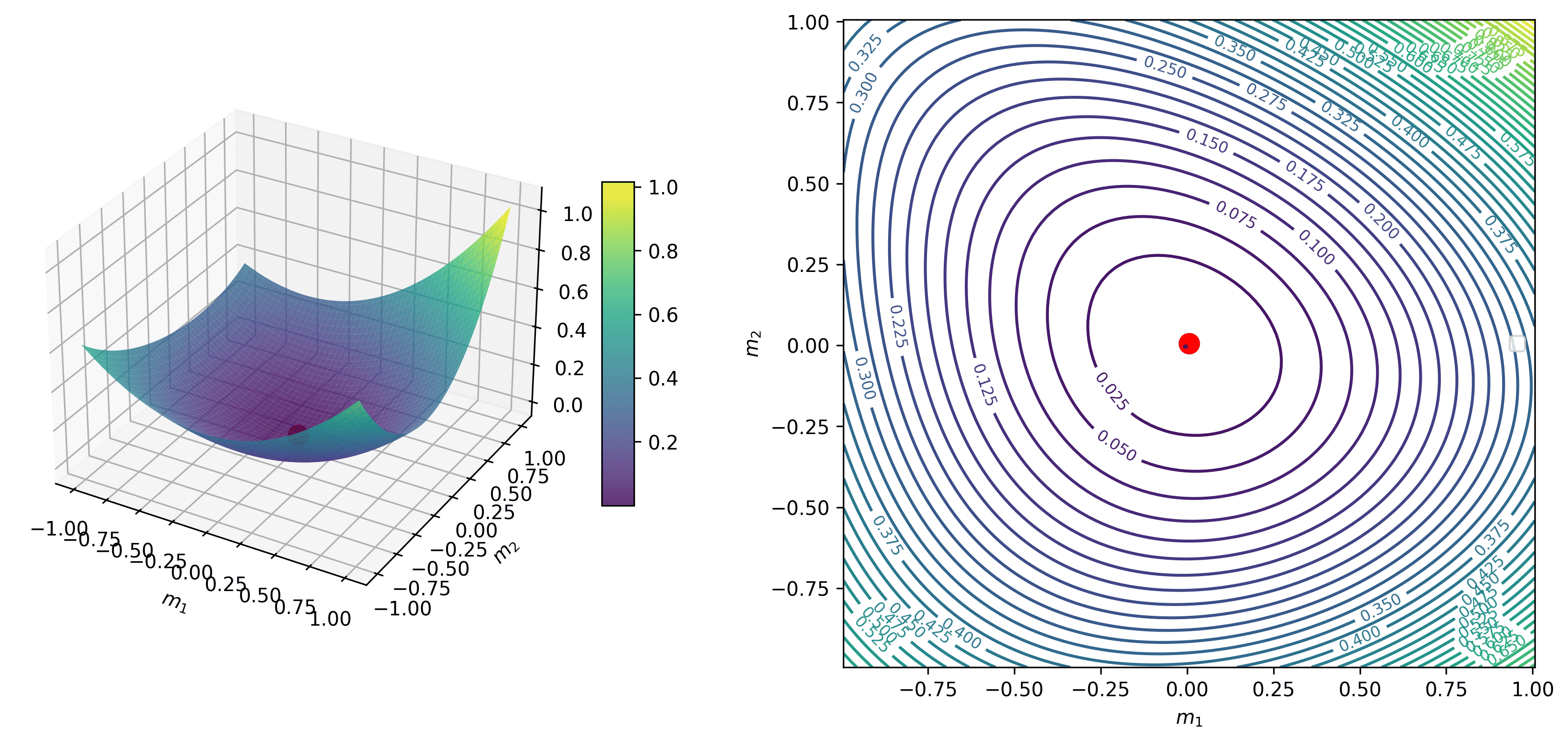}
    \caption{Illustration of the convex objective \(J(\textbf{m})\) in a 2D slice of the higher-dimensional space. The single global minimum, marked in red, highlights the function’s convex nature.}
    \label{fig:convexity}
\end{figure}

\section{Experiments}
\begin{table*}[!ht]
\centering
\resizebox{\textwidth}{!}{ 
\begin{tabular}{>{\centering\arraybackslash}m{3cm}
                >{\centering\arraybackslash}m{2.5cm}:
                >{\centering\arraybackslash}m{2.5cm}
                >{\centering\arraybackslash}m{2.5cm}
                >{\centering\arraybackslash}m{2.5cm}
                >{\centering\arraybackslash}m{2.5cm}
                >{\centering\arraybackslash}m{2.5cm}}
 & \textbf{Source Image} & \textbf{LOCATEdit} & \textbf{ViMAEdit} & \textbf{InfEdit} & \textbf{MasaCtrl} & \textbf{LEDITS++} \\ 

an \textcolor{red}{\sout{orange}} \textbf{black} cat sitting on top of a fence
& \includegraphics[width=\linewidth,height=2.5cm,keepaspectratio]{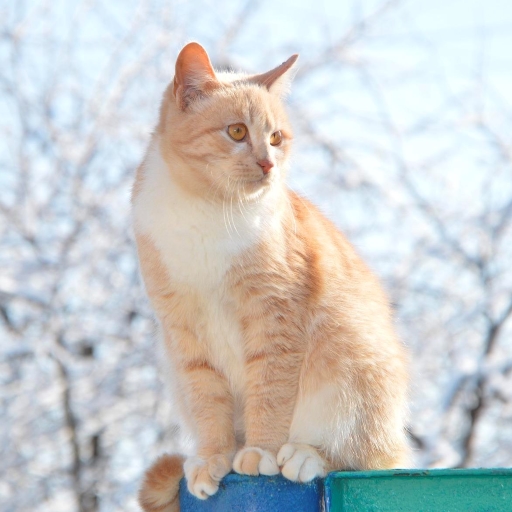} 
& \includegraphics[width=\linewidth,height=2.5cm,keepaspectratio]{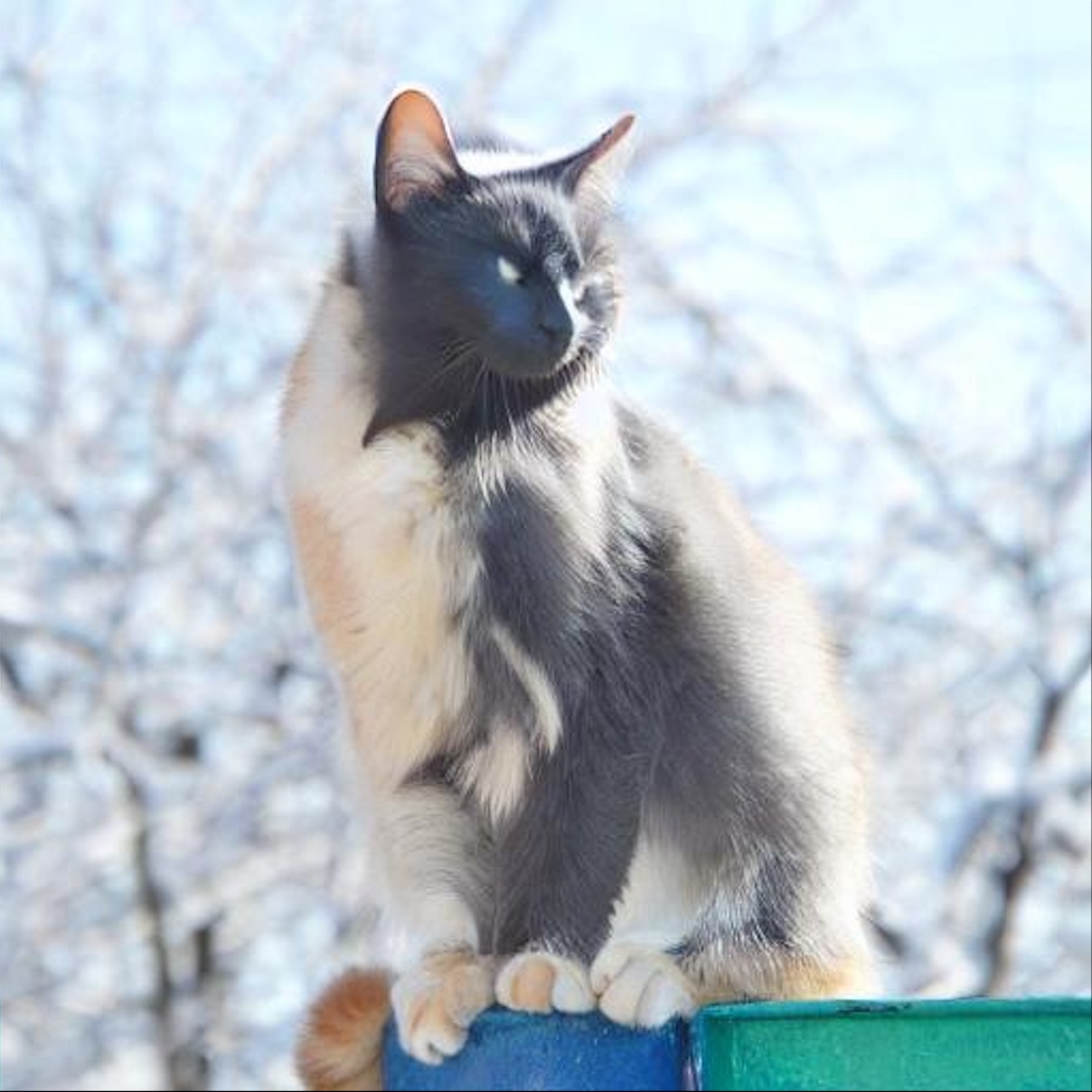} 
& \includegraphics[width=\linewidth,height=2.5cm,keepaspectratio]{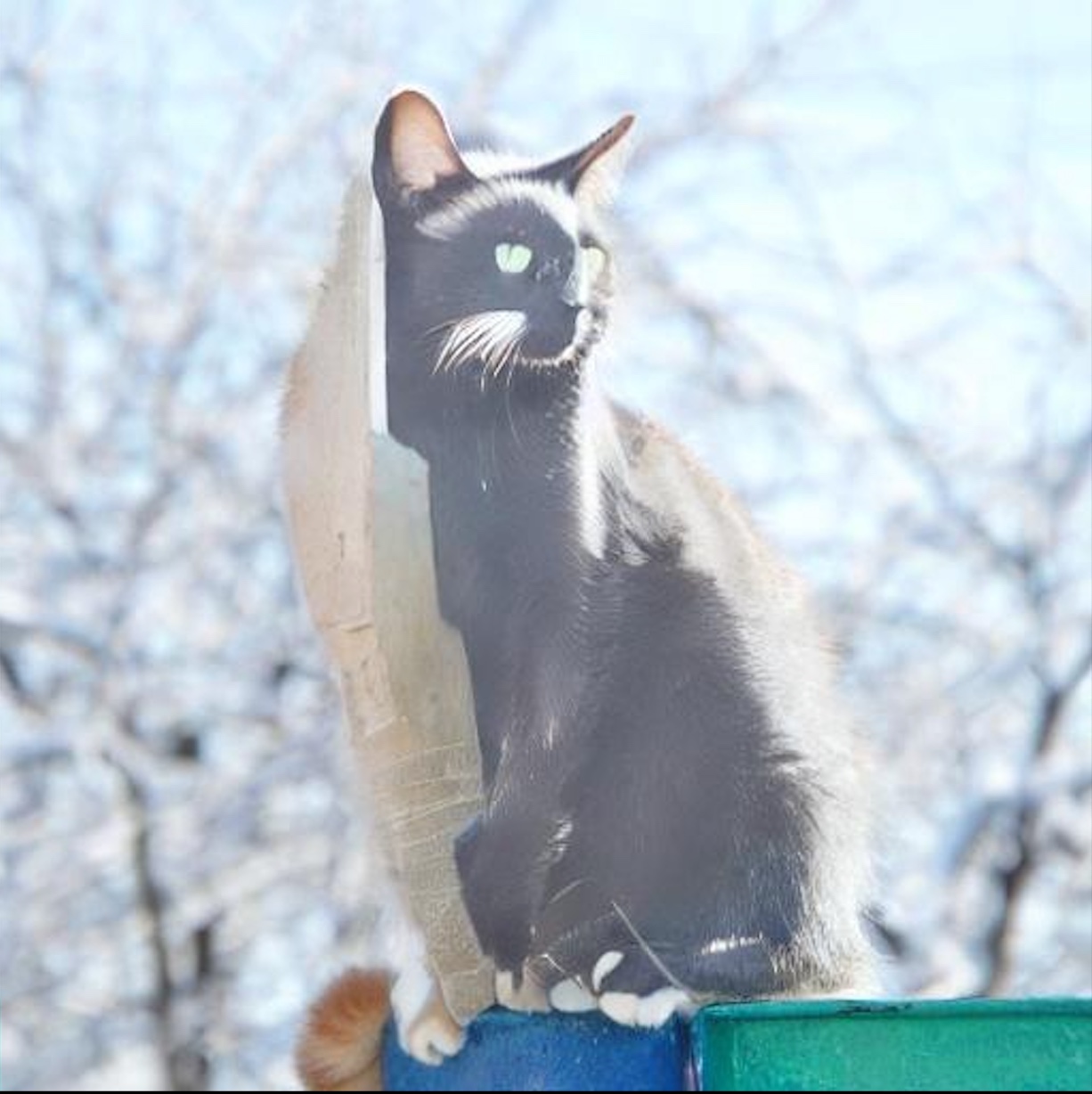} 
& \includegraphics[width=\linewidth,height=2.5cm,keepaspectratio]{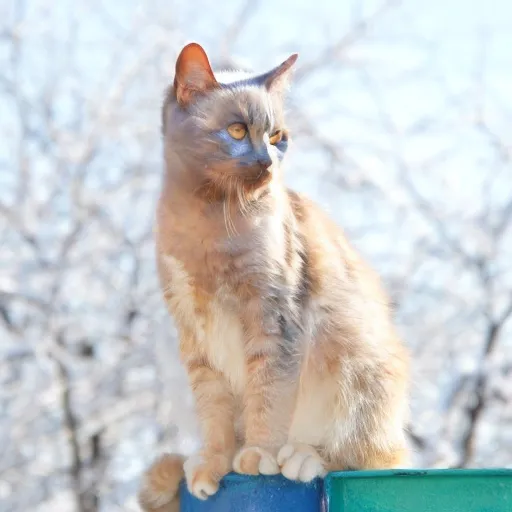} 
& \includegraphics[width=\linewidth,height=2.5cm,keepaspectratio]{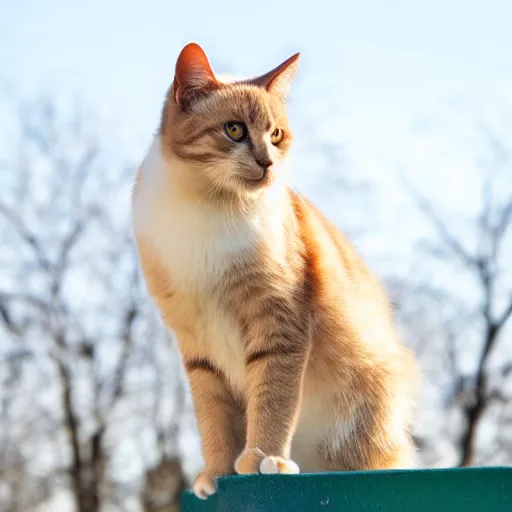} 
& \includegraphics[width=\linewidth,height=2.5cm,keepaspectratio]{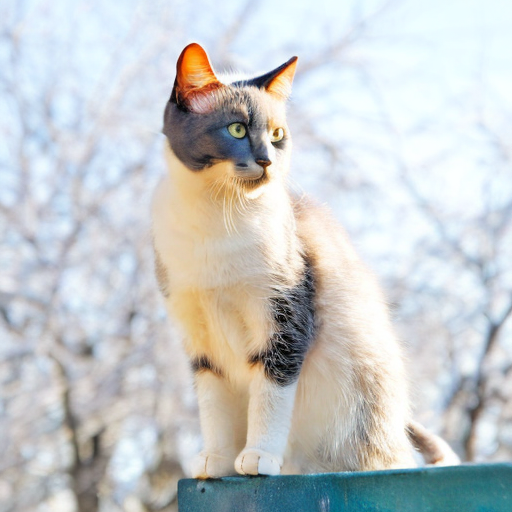} \\ 

a \textcolor{red}{\sout{cat}} \textbf{tiger} sitting next to a mirror
& \includegraphics[width=\linewidth,height=2.5cm,keepaspectratio]{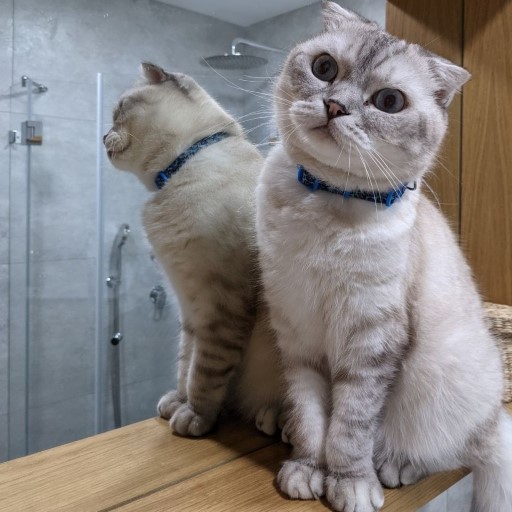} 
& \includegraphics[width=\linewidth,height=2.5cm,keepaspectratio]{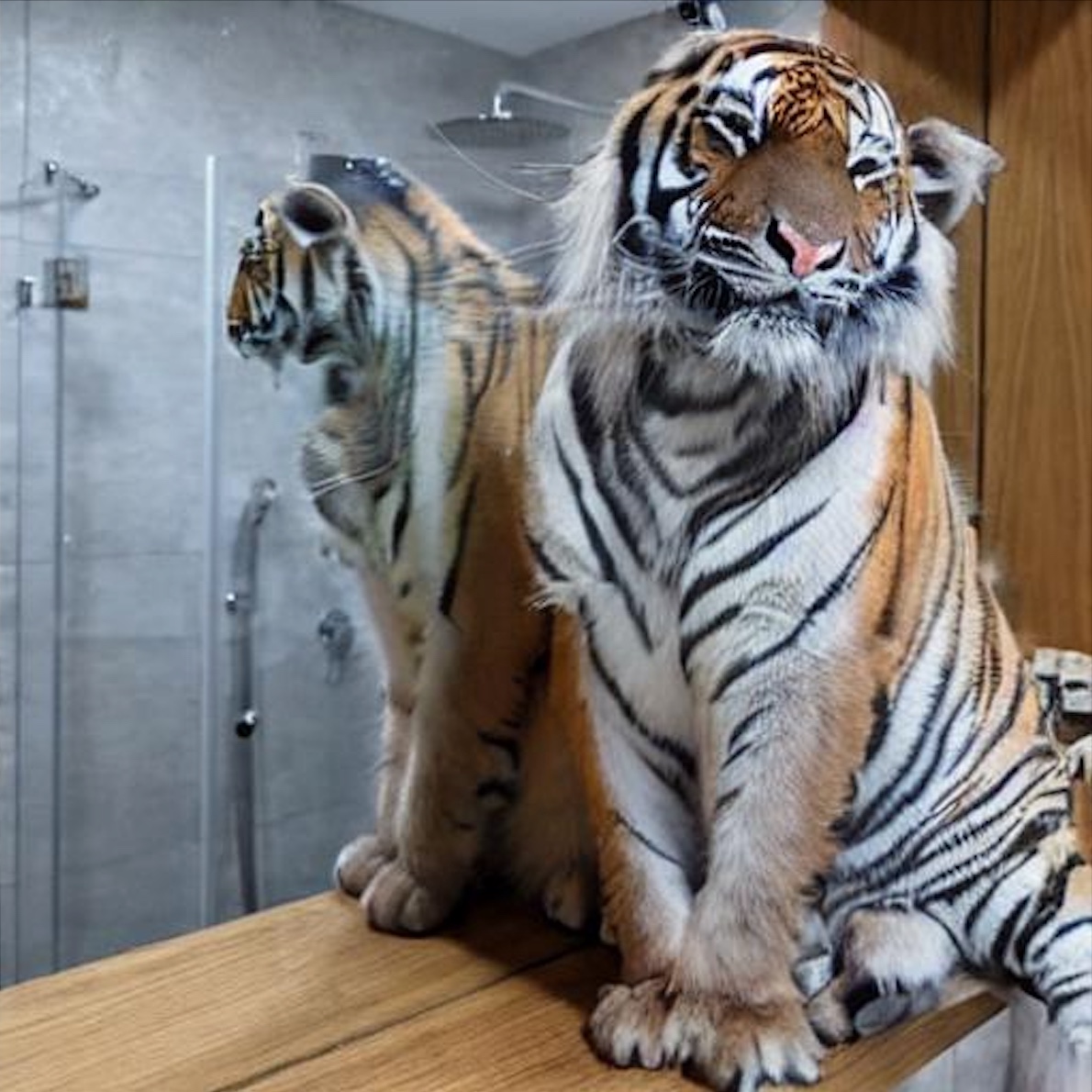} 
& \includegraphics[width=\linewidth,height=2.5cm,keepaspectratio]{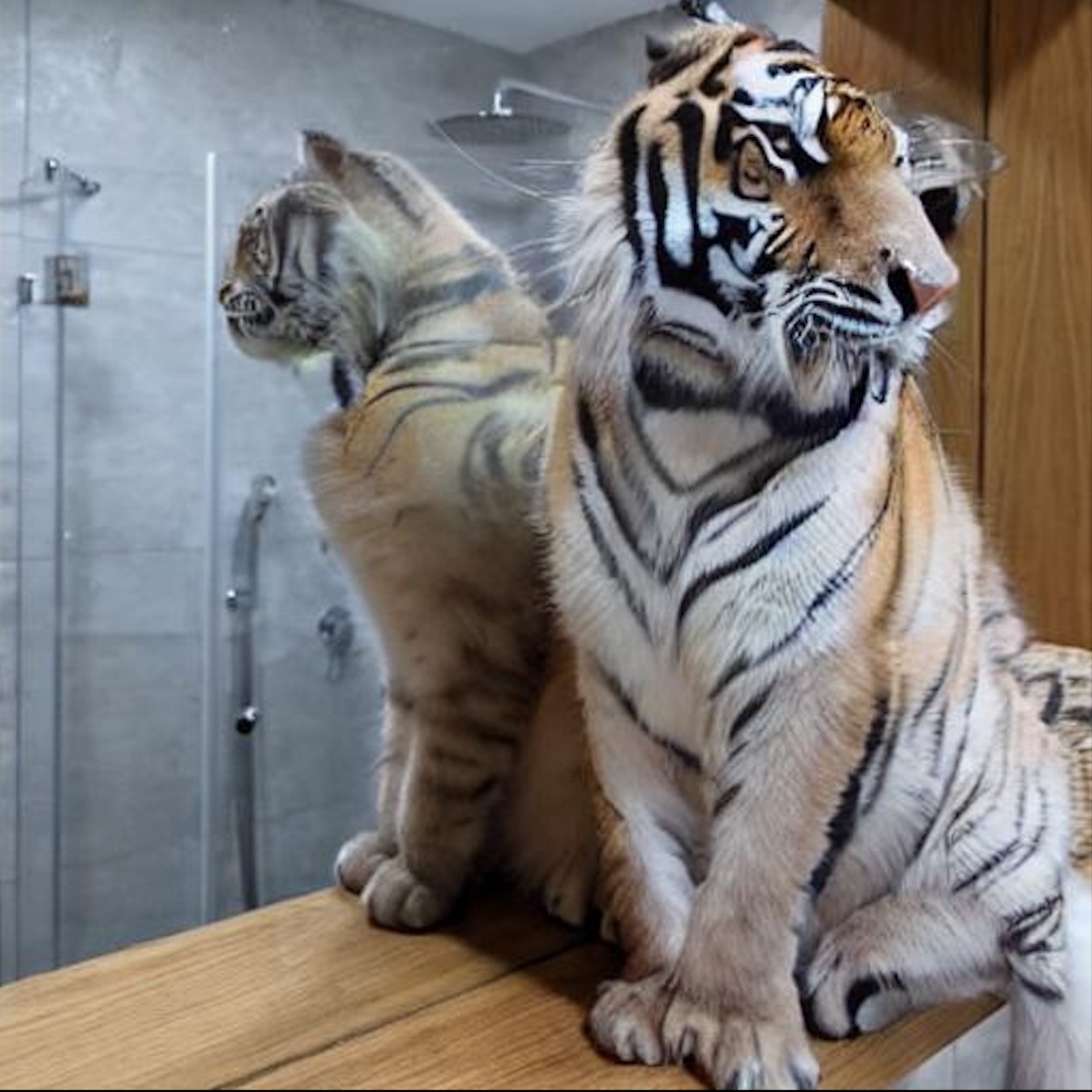} 
& \includegraphics[width=\linewidth,height=2.5cm,keepaspectratio]{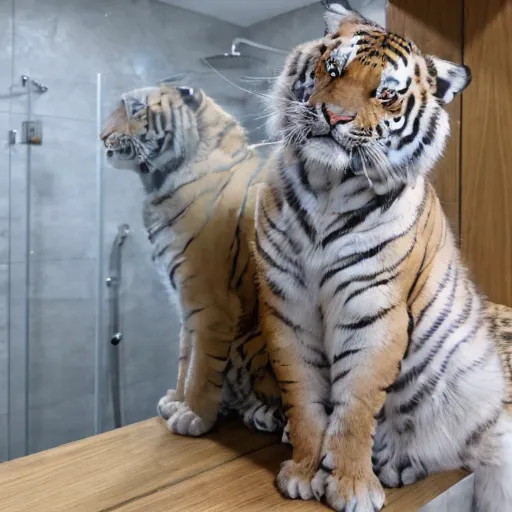} 
& \includegraphics[width=\linewidth,height=2.5cm,keepaspectratio]{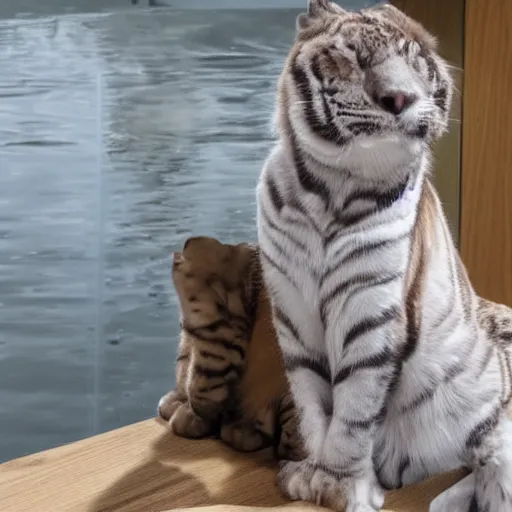} 
& \includegraphics[width=\linewidth,height=2.5cm,keepaspectratio]{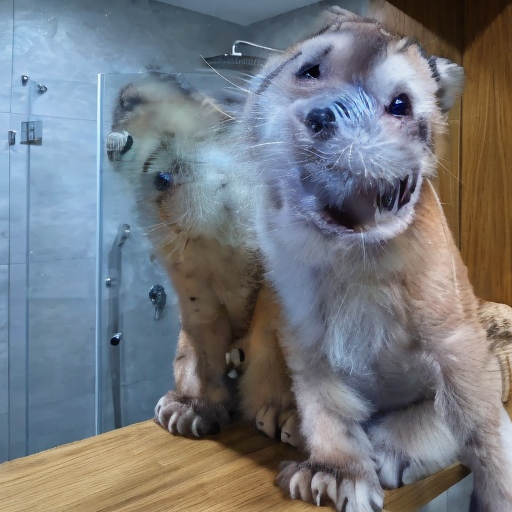}  \\ 

the \textcolor{red}{\sout{crescent moon}} \textbf{golden crescent moon} and stars are seen in the night sky
& \includegraphics[width=\linewidth,height=2.5cm,keepaspectratio]{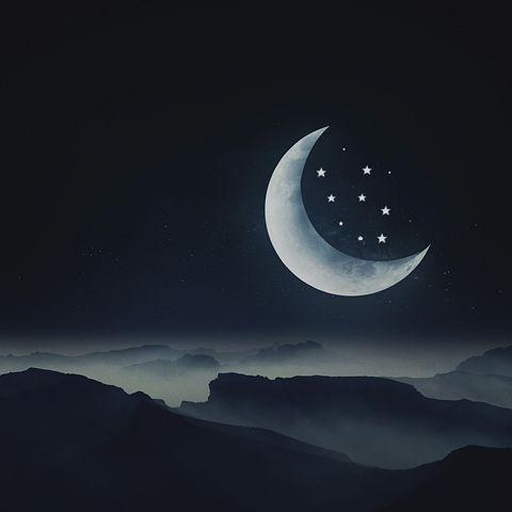} 
& \includegraphics[width=\linewidth,height=2.5cm,keepaspectratio]{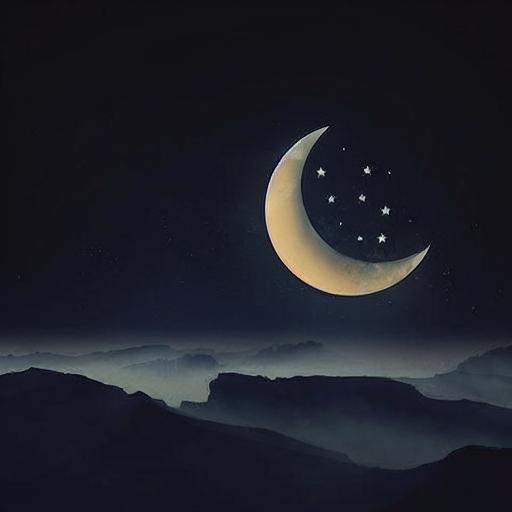} 
& \includegraphics[width=\linewidth,height=2.5cm,keepaspectratio]{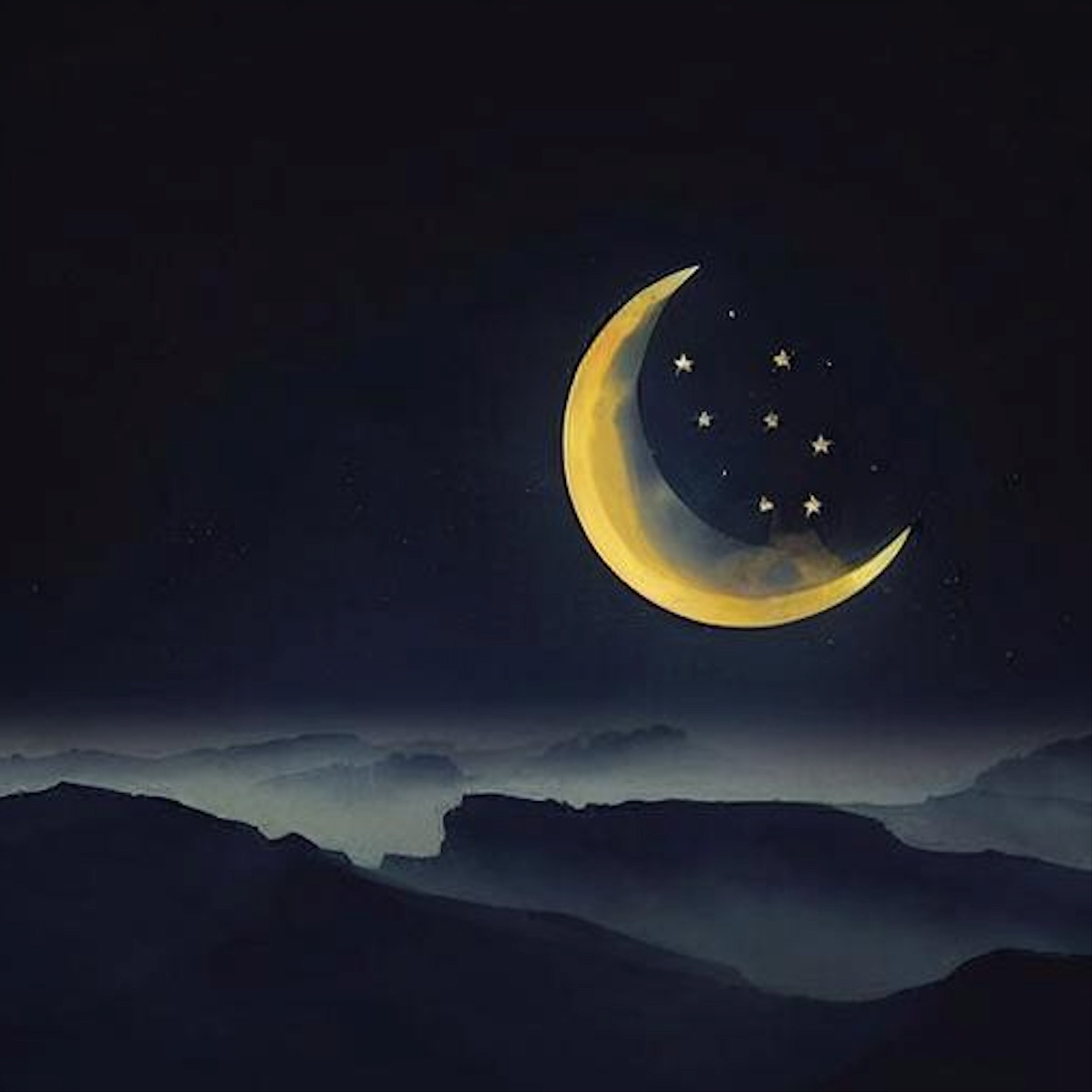} 
& \includegraphics[width=\linewidth,height=2.5cm,keepaspectratio]{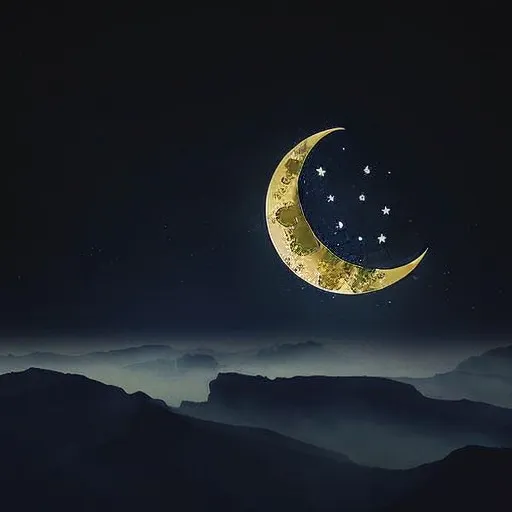} 
& \includegraphics[width=\linewidth,height=2.5cm,keepaspectratio]{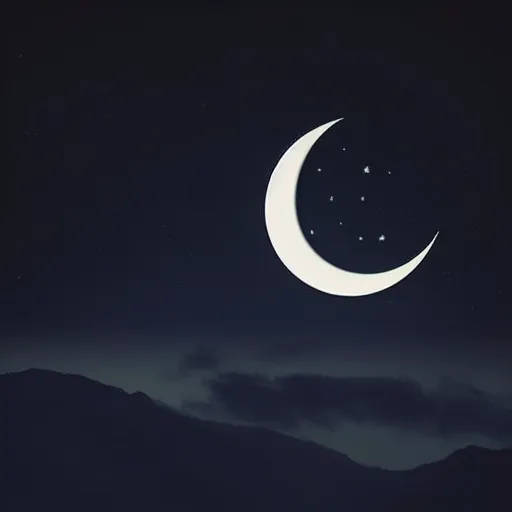} 
& \includegraphics[width=\linewidth,height=2.5cm,keepaspectratio]{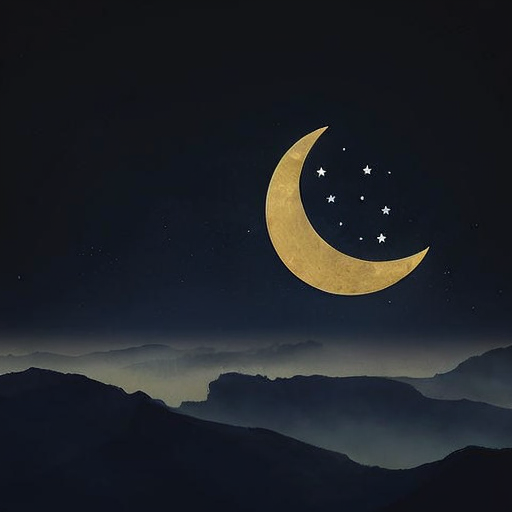}  \\ 

A \textcolor{red}{\sout{white}} \textbf{golden} horse running in the sunset
& \includegraphics[width=\linewidth,height=2.5cm,keepaspectratio]{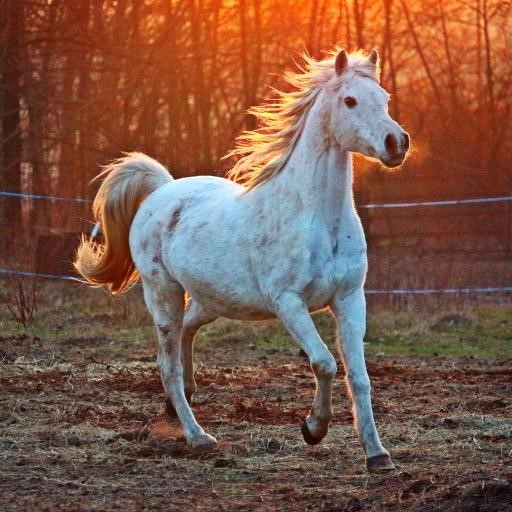} 
& \includegraphics[width=\linewidth,height=2.5cm,keepaspectratio]{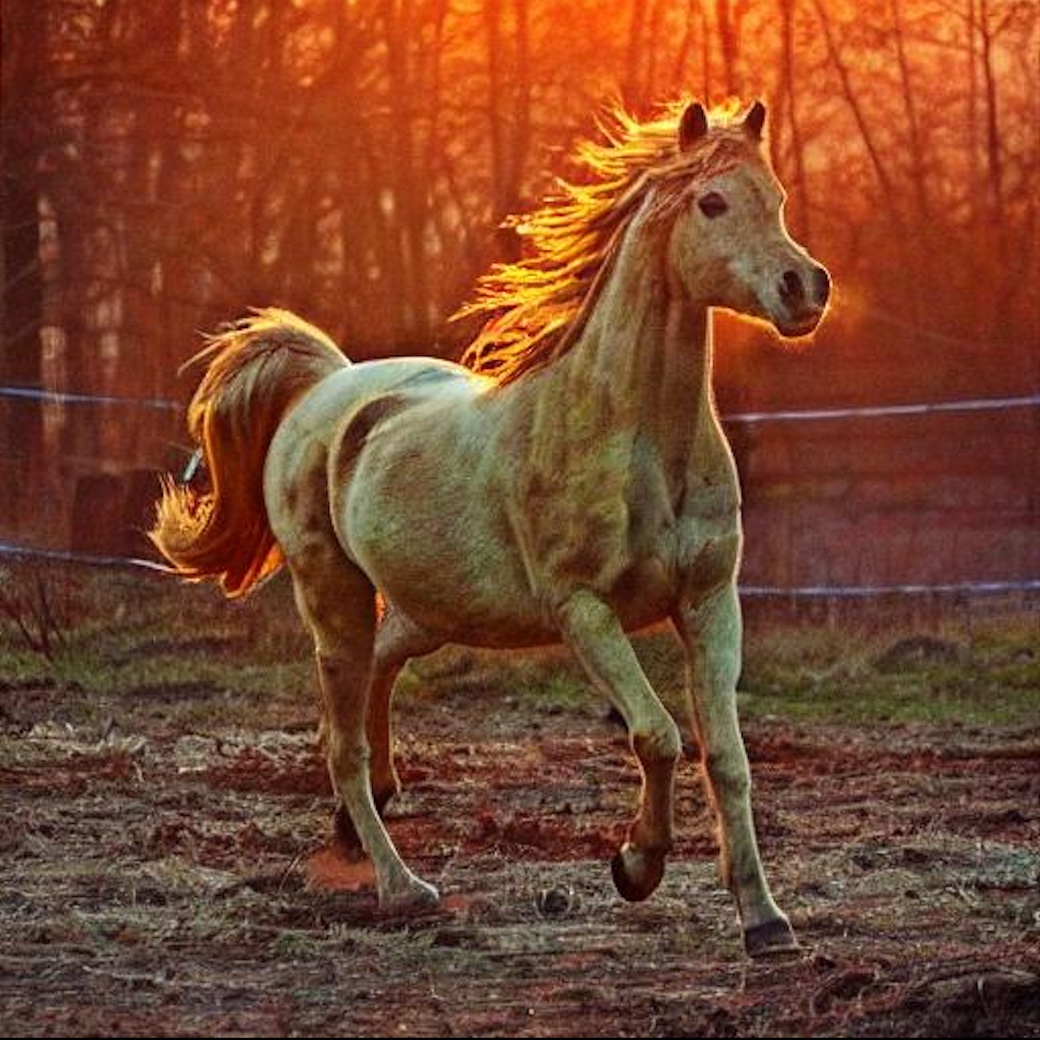} 
& \includegraphics[width=\linewidth,height=2.5cm,keepaspectratio]{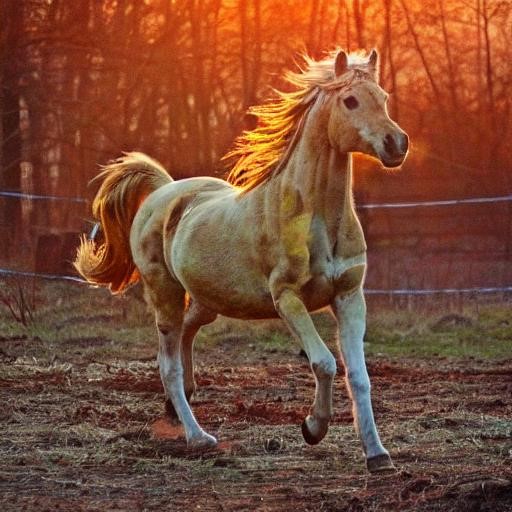} 
& \includegraphics[width=\linewidth,height=2.5cm,keepaspectratio]{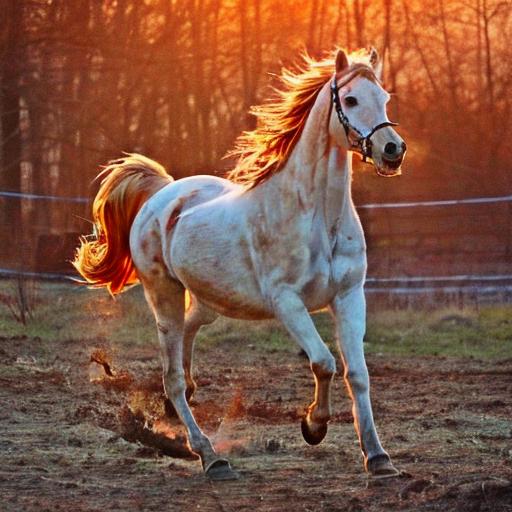} 
& \includegraphics[width=\linewidth,height=2.5cm,keepaspectratio]{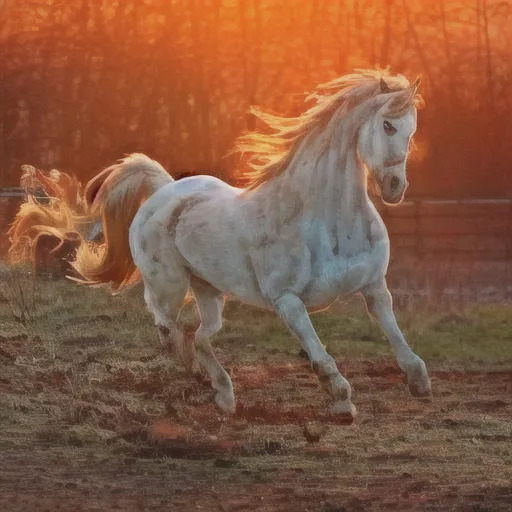} 
& \includegraphics[width=\linewidth,height=2.5cm,keepaspectratio]{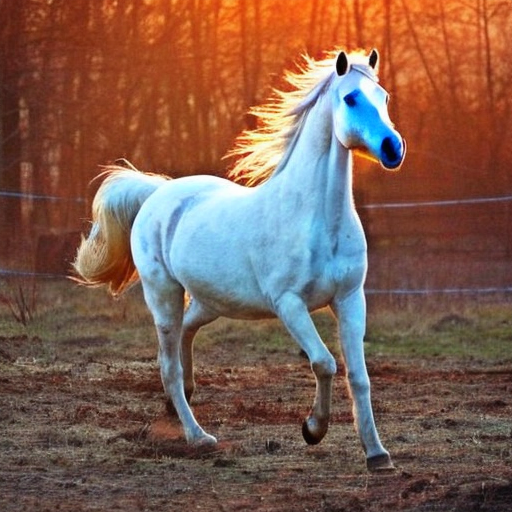}  \\ 

a \textcolor{red}{\sout{open}}  \textbf{closed} eyes cat sitting on wooden floor
& \includegraphics[width=\linewidth,height=2.5cm,keepaspectratio]{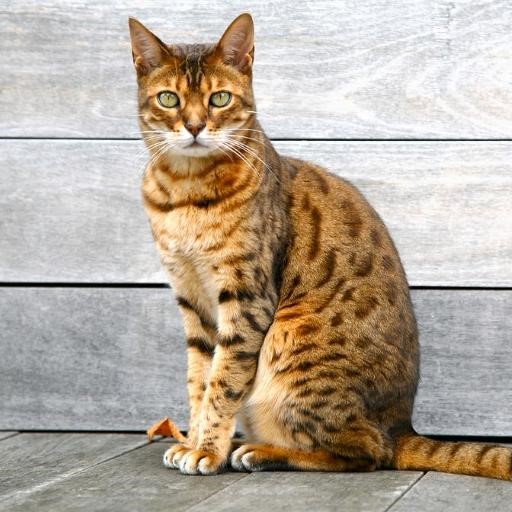} 
& \includegraphics[width=\linewidth,height=2.5cm,keepaspectratio]{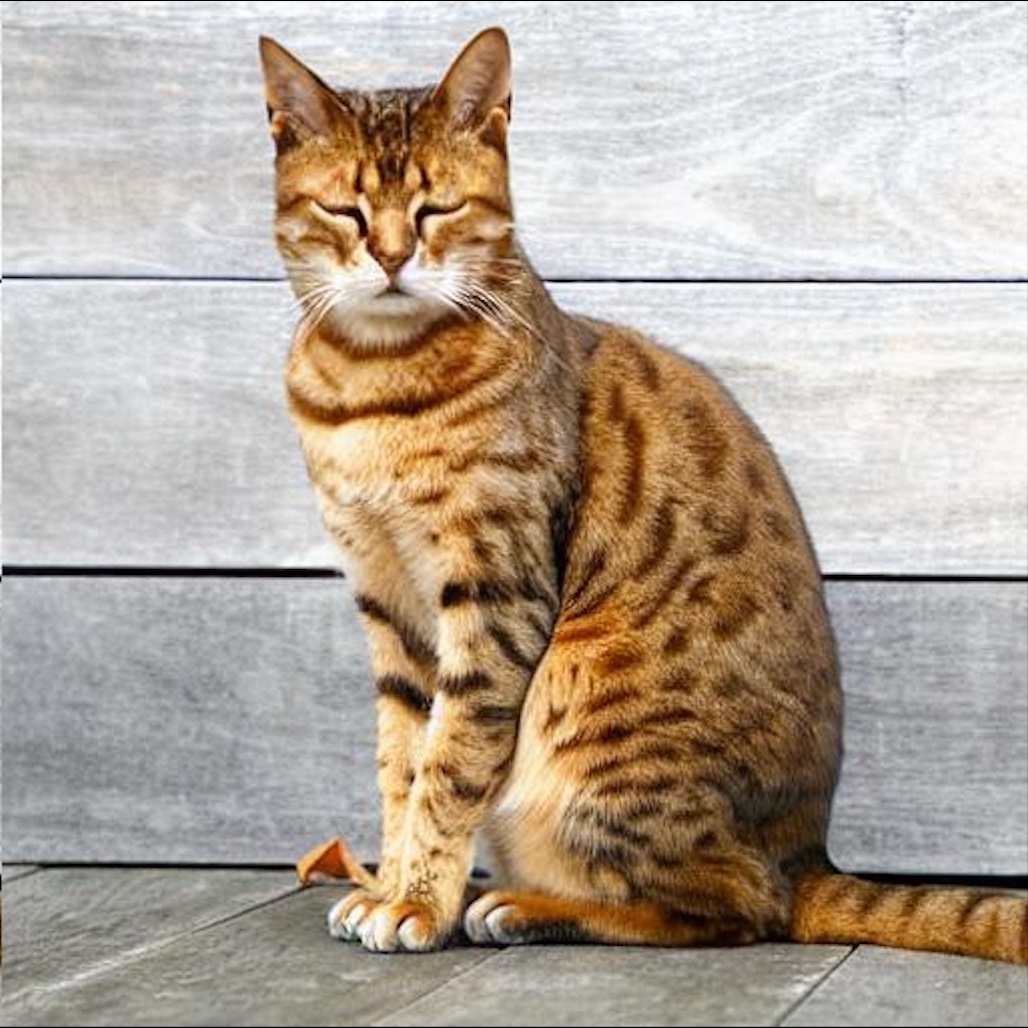} 
& \includegraphics[width=\linewidth,height=2.5cm,keepaspectratio]{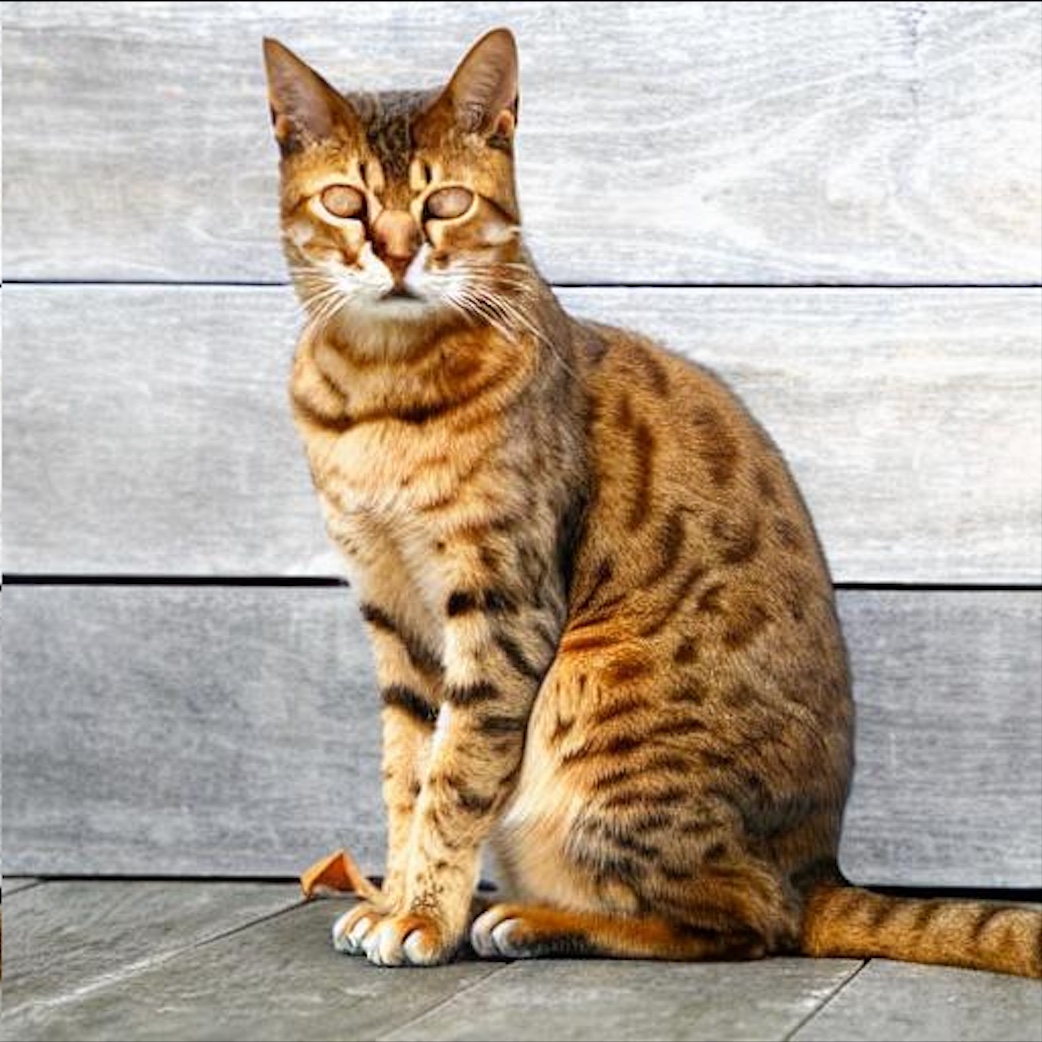} 
& \includegraphics[width=\linewidth,height=2.5cm,keepaspectratio]{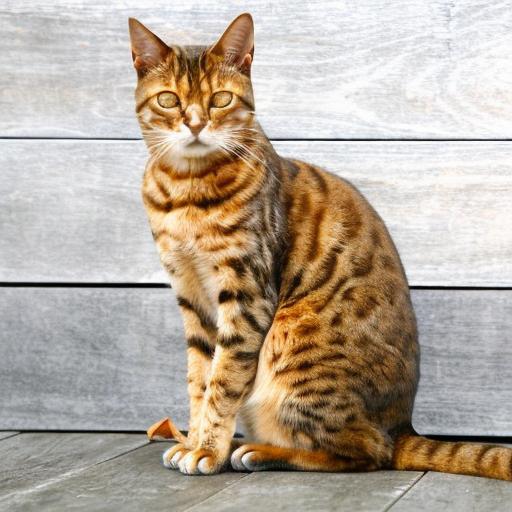} 
& \includegraphics[width=\linewidth,height=2.5cm,keepaspectratio]{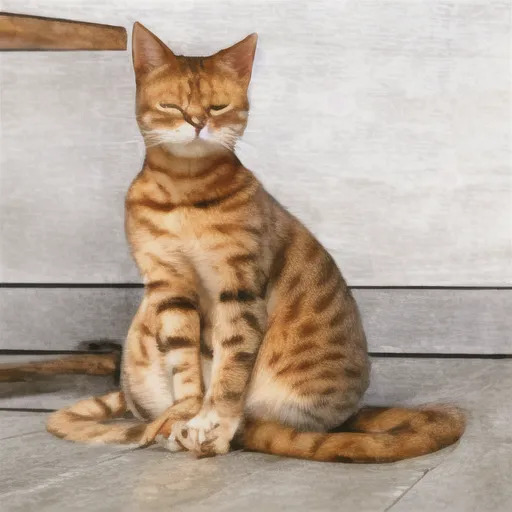} 
& \includegraphics[width=\linewidth,height=2.5cm,keepaspectratio]{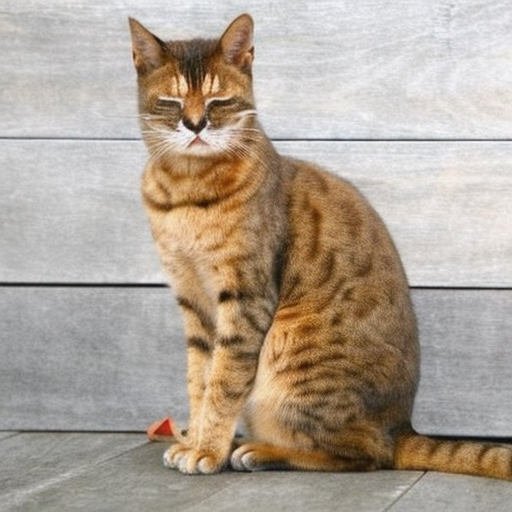}  \\

a \textcolor{red}{\sout{kitten}} \textbf{duck} walking through the grass
& \includegraphics[width=\linewidth,height=2.5cm,keepaspectratio]{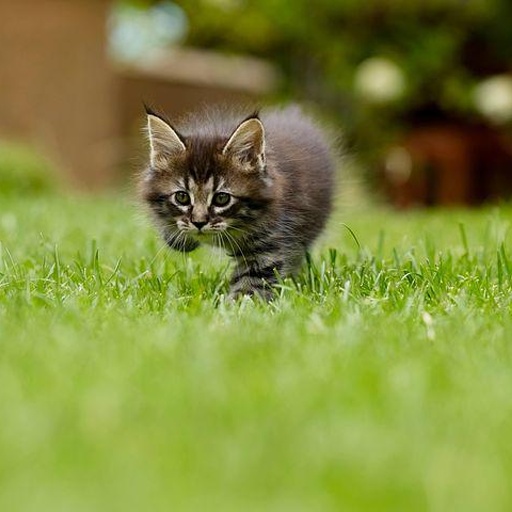} 
& \includegraphics[width=\linewidth,height=2.5cm,keepaspectratio]{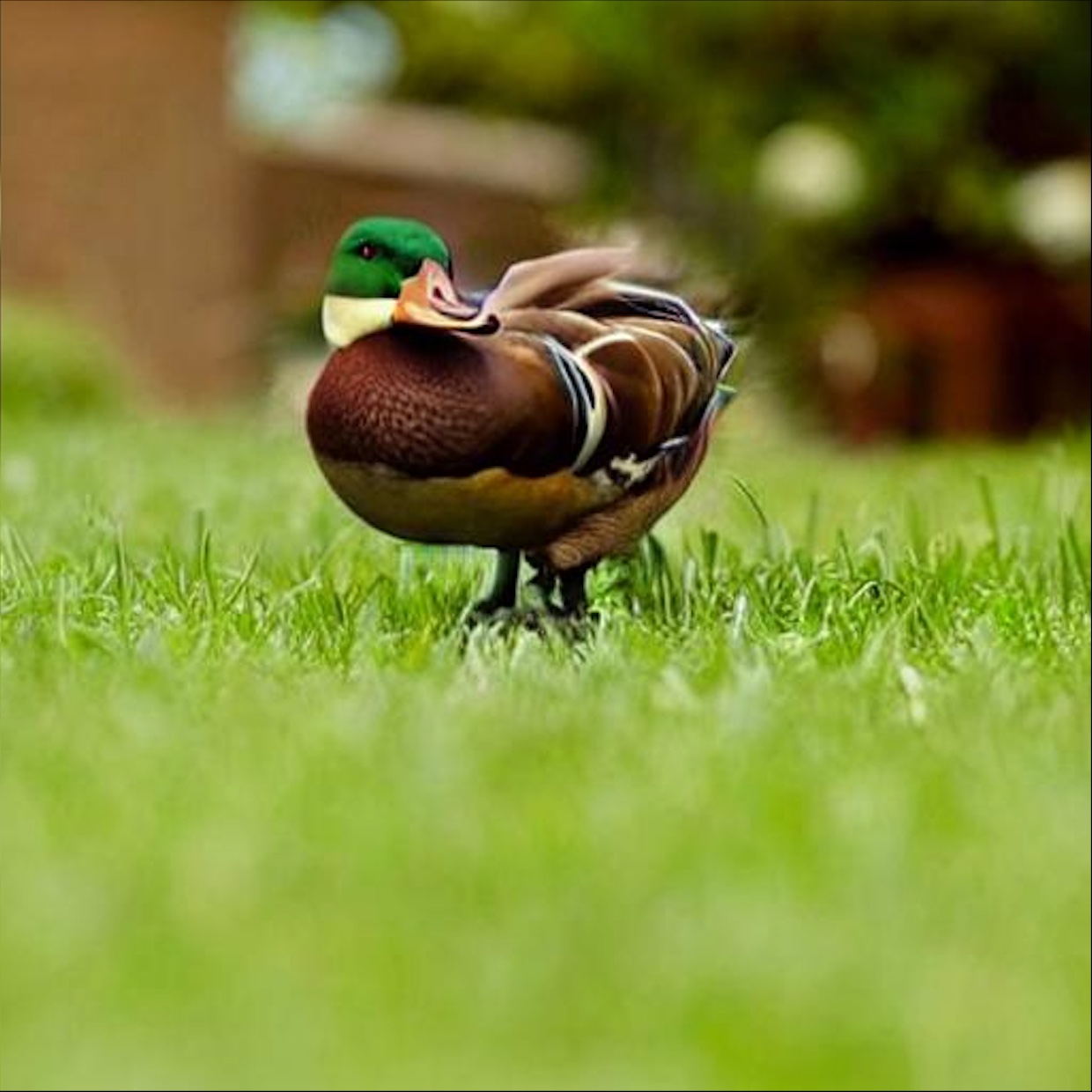} 
& \includegraphics[width=\linewidth,height=2.5cm,keepaspectratio]{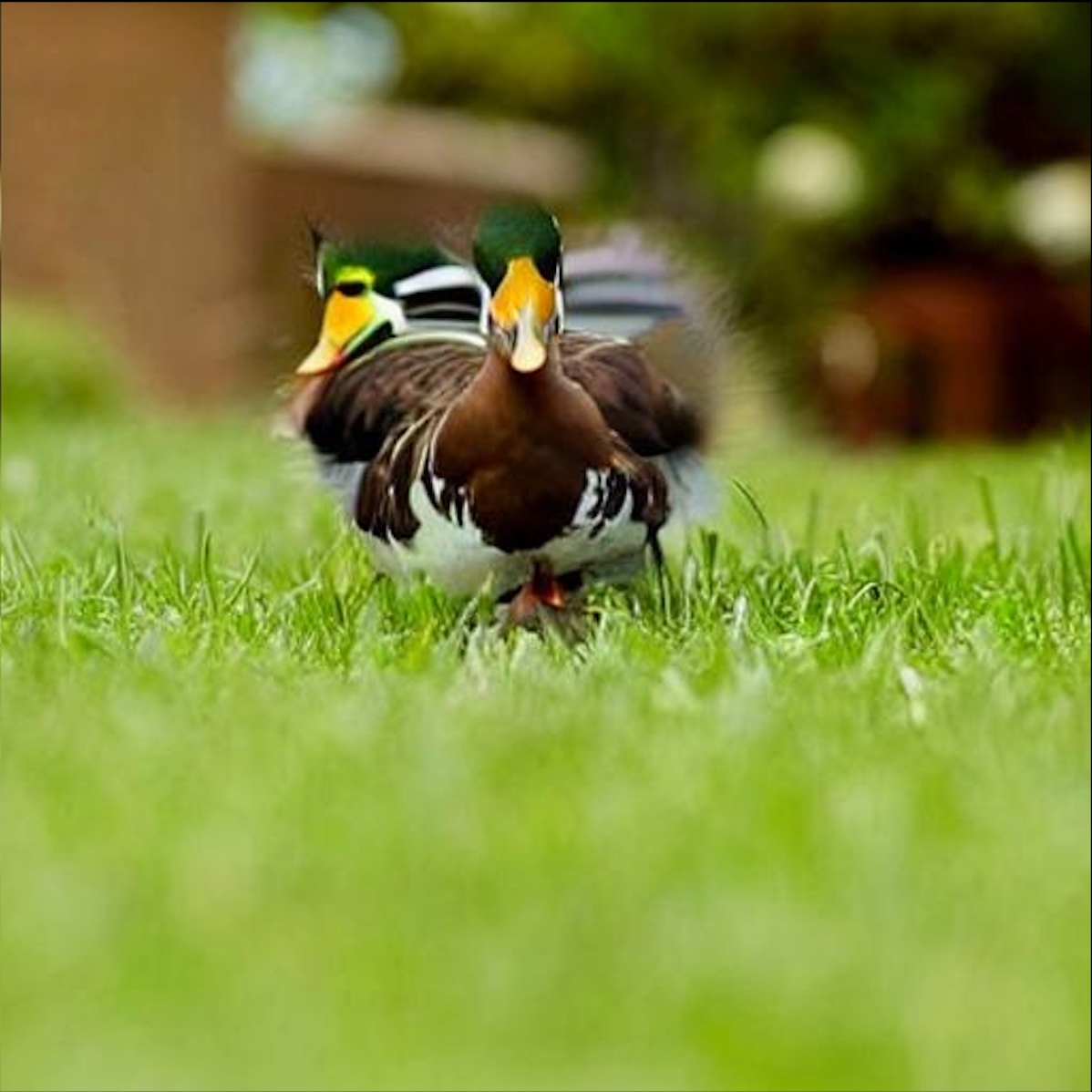} 
& \includegraphics[width=\linewidth,height=2.5cm,keepaspectratio]{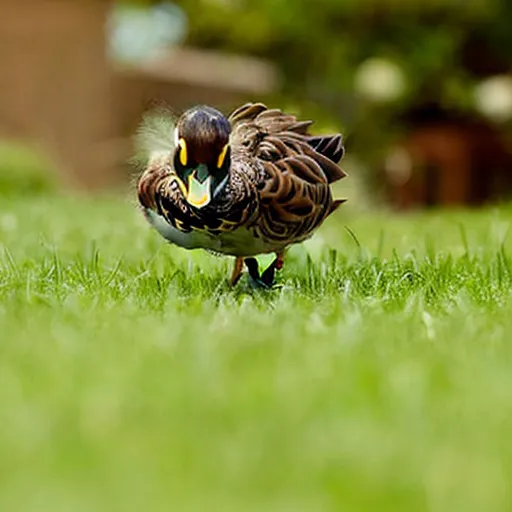} 
& \includegraphics[width=\linewidth,height=2.5cm,keepaspectratio]{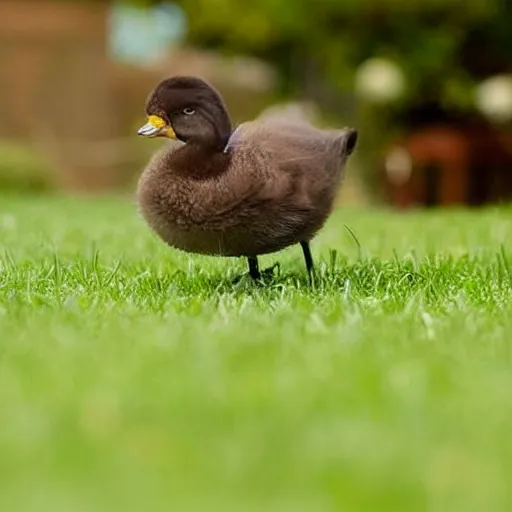} 
& \includegraphics[width=\linewidth,height=2.5cm,keepaspectratio]{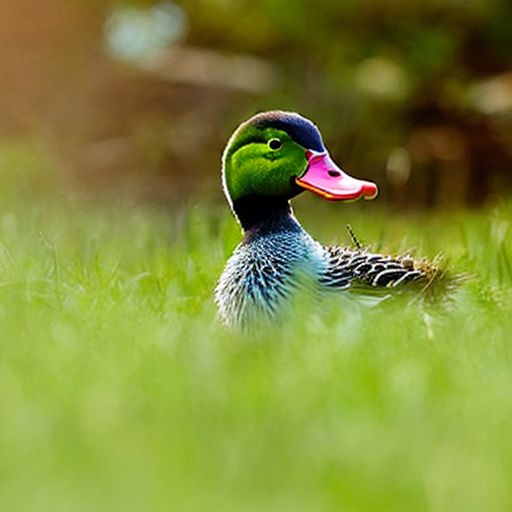} \\

a boat is docked on a lake in the \textcolor{red}{\sout{heavy fog}} \textbf{sunny day}
& \includegraphics[width=\linewidth,height=2.5cm,keepaspectratio]{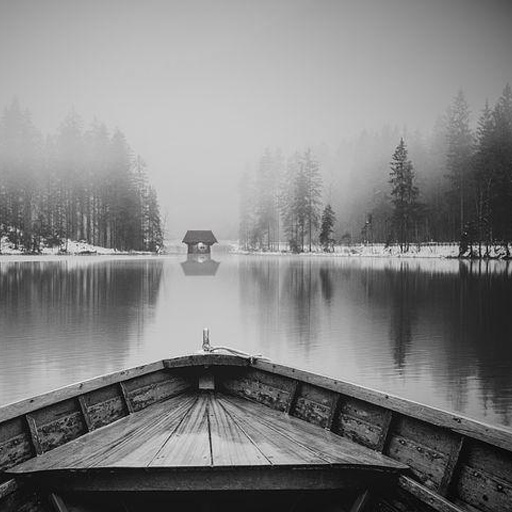} 
& \includegraphics[width=\linewidth,height=2.5cm,keepaspectratio]{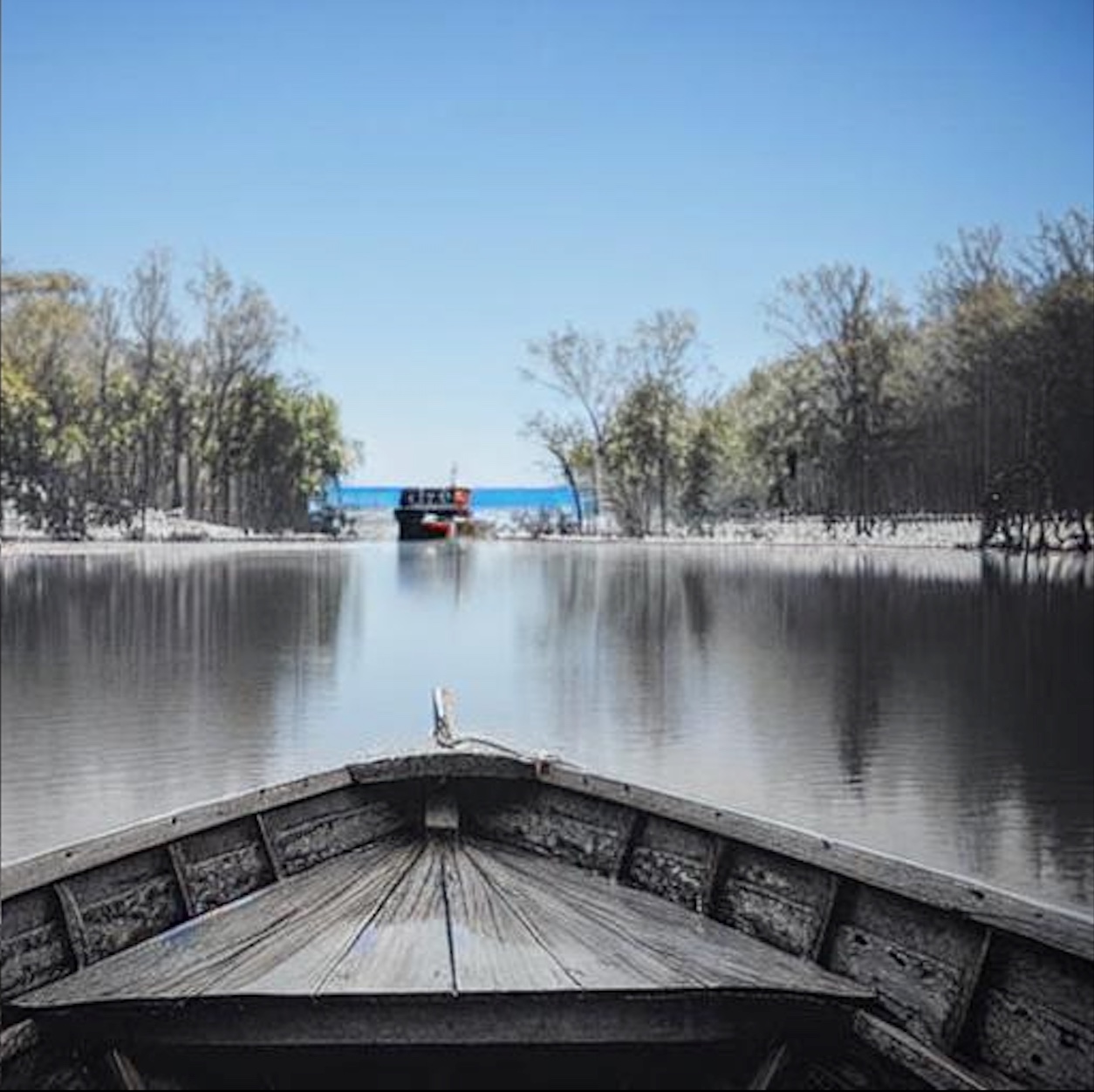} 
& \includegraphics[width=\linewidth,height=2.5cm,keepaspectratio]{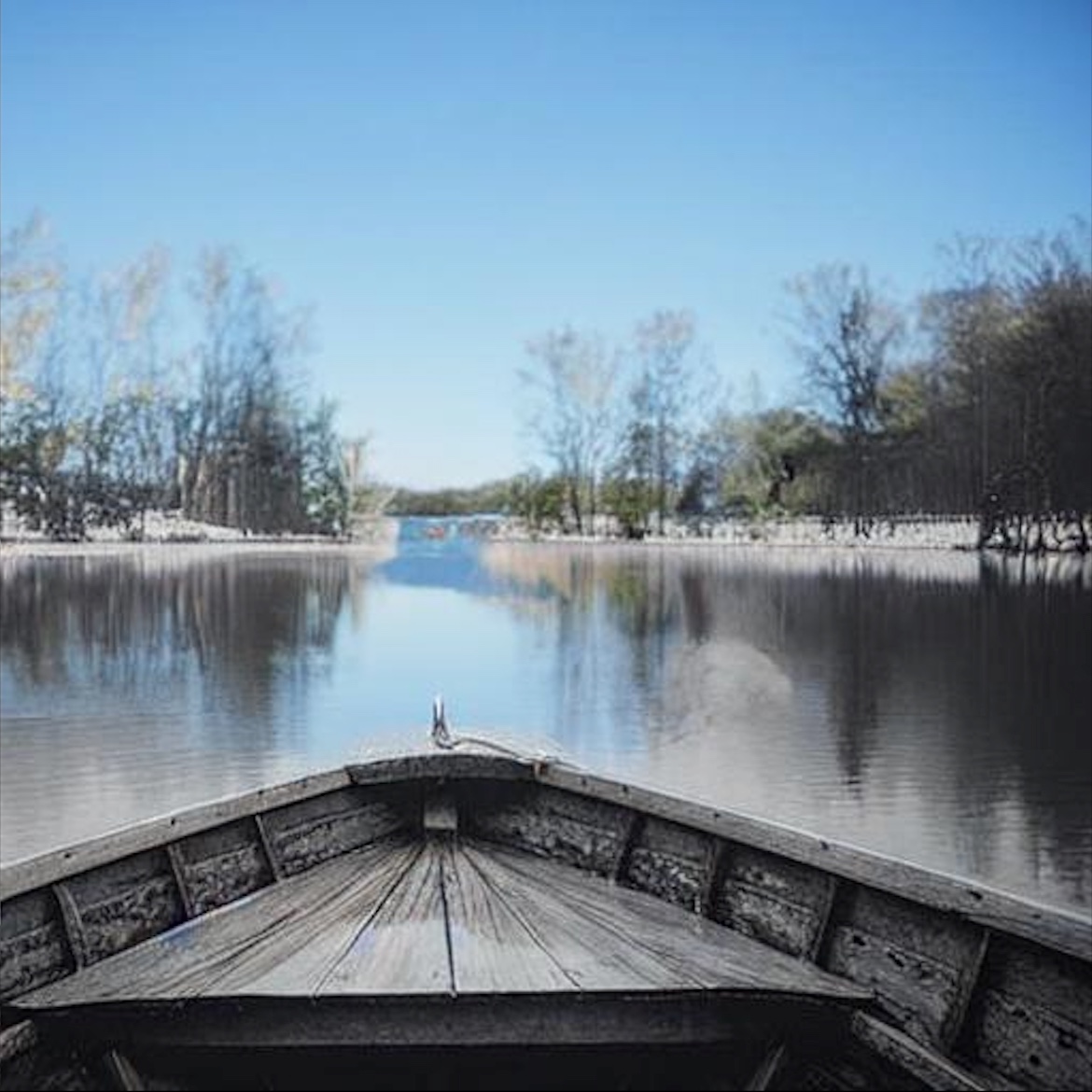} 
& \includegraphics[width=\linewidth,height=2.5cm,keepaspectratio]{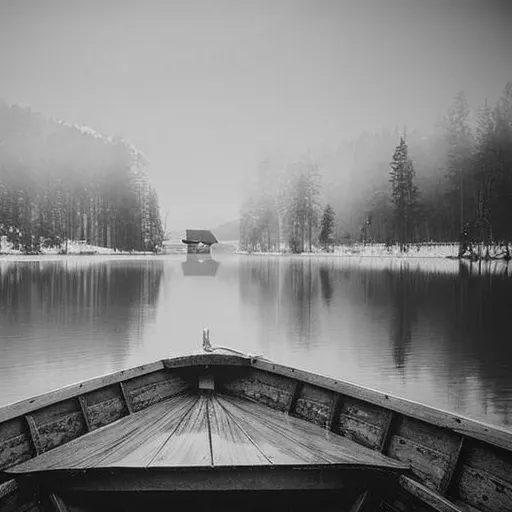} 
& \includegraphics[width=\linewidth,height=2.5cm,keepaspectratio]{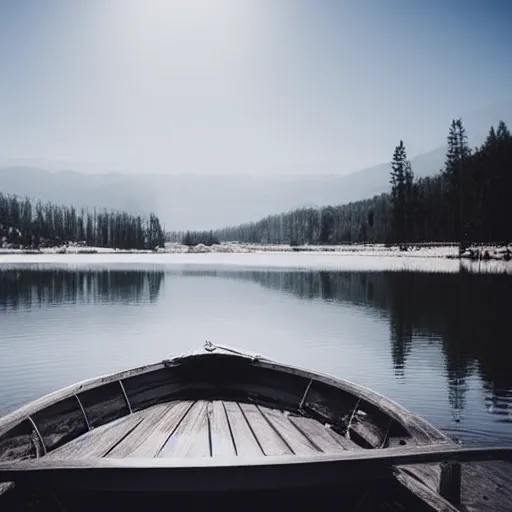} 
& \includegraphics[width=\linewidth,height=2.5cm,keepaspectratio]{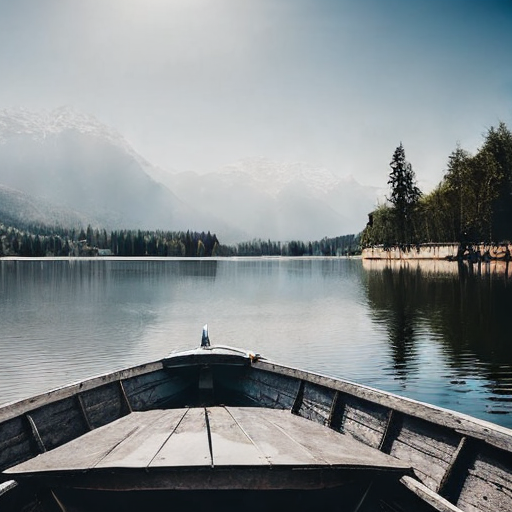} \\

a \textcolor{red}{\sout{woman}} \textbf{man}  and a horse 
& \includegraphics[width=\linewidth,height=2.5cm,keepaspectratio]{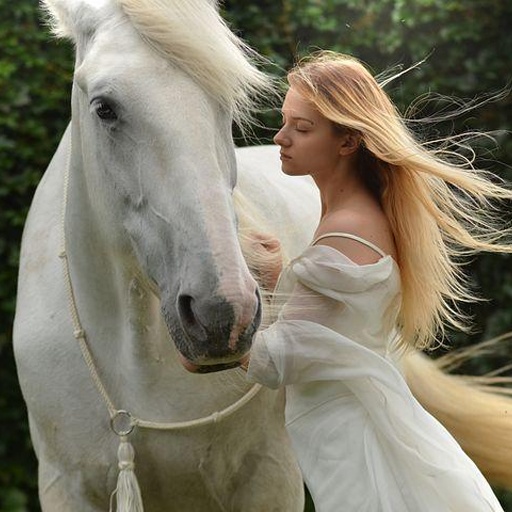} 
& \includegraphics[width=\linewidth,height=2.5cm,keepaspectratio]{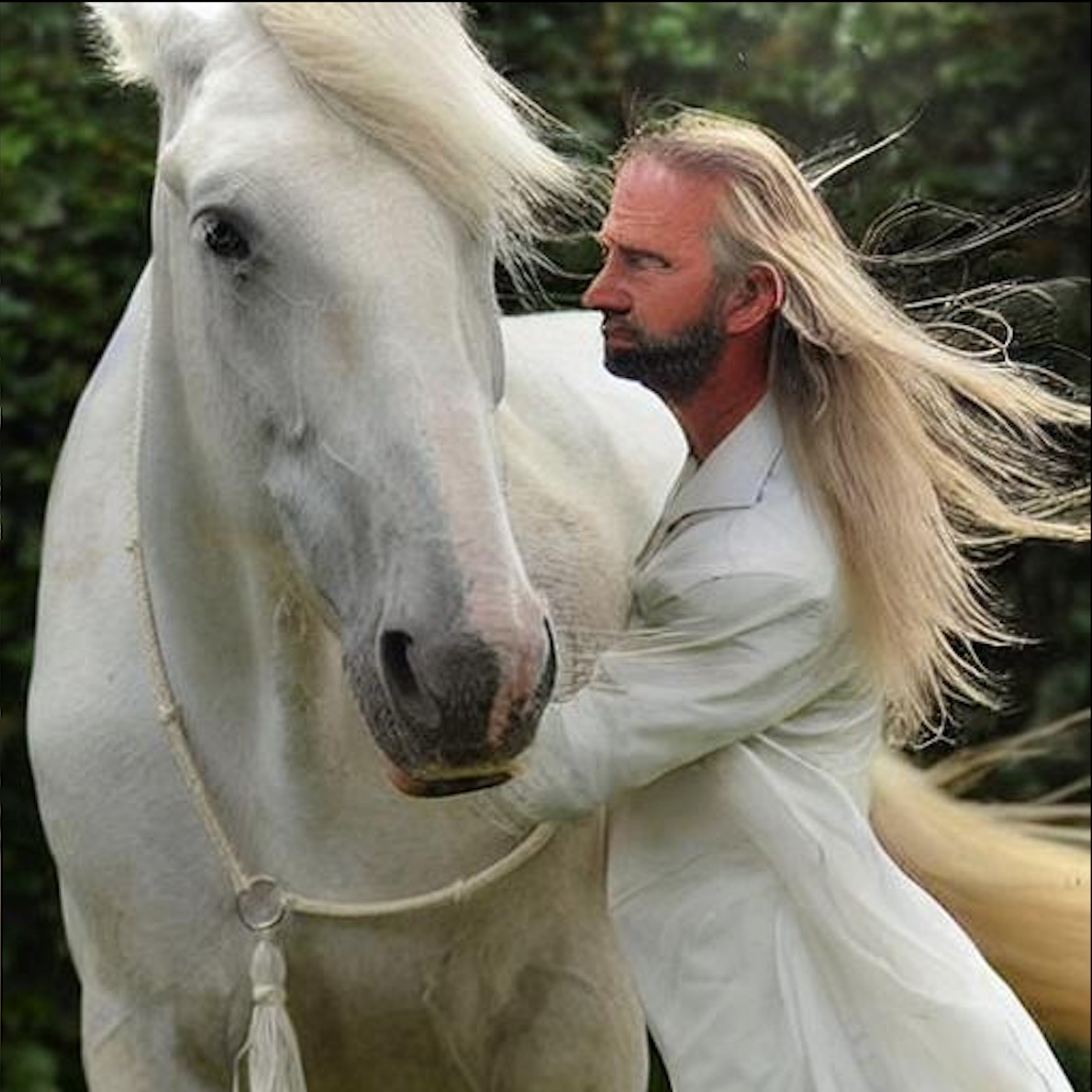} 
& \includegraphics[width=\linewidth,height=2.5cm,keepaspectratio]{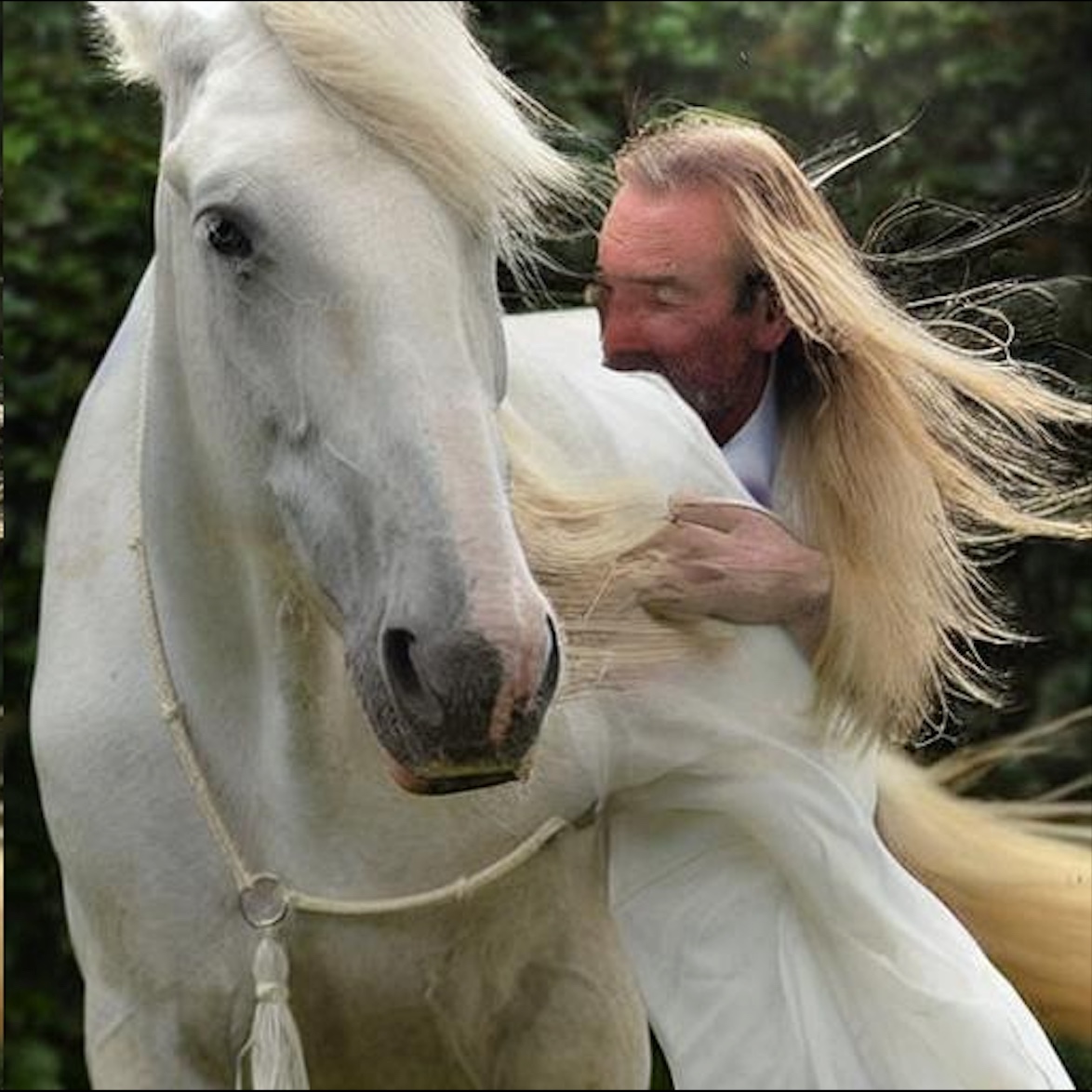} 
& \includegraphics[width=\linewidth,height=2.5cm,keepaspectratio]{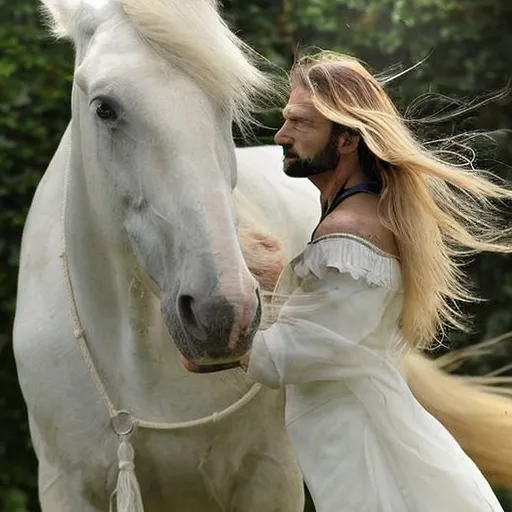} 
& \includegraphics[width=\linewidth,height=2.5cm,keepaspectratio]{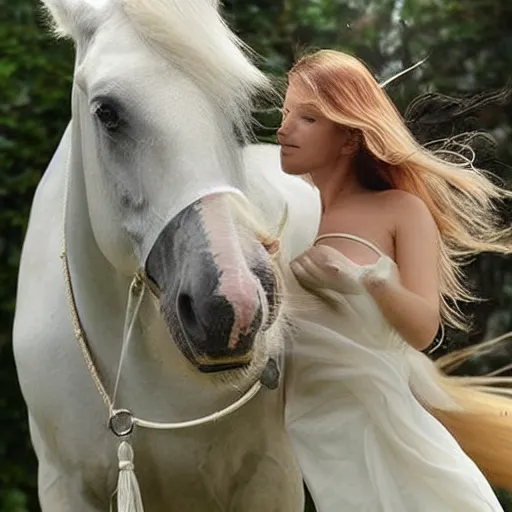} 
& \includegraphics[width=\linewidth,height=2.5cm,keepaspectratio]{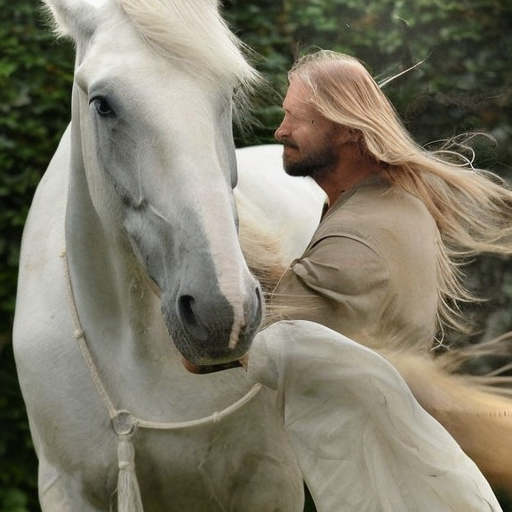} \\

\end{tabular}}

\caption{Qualitative comparisons with competing text-guided editing methods. \method yields more localized edits while preserving overall structure, outperforming baselines in both fidelity and consistency.}
\label{fig:comparison}
\end{table*}

\begingroup
\renewcommand{\arraystretch}{1} 
\begin{table*}[t]
    \centering
    \small
    \setlength{\tabcolsep}{3.5pt} 
    \resizebox{\textwidth}{!}{ 
    \begin{tabular}{c|c|c|c|cccc|cc}
        \hline
        \multicolumn{2}{c|}{\textbf{Method}} & \textbf{Editing} & \textbf{Structure} & \multicolumn{4}{c|}{\textbf{Background Preservation}} & \multicolumn{2}{c}{\textbf{CLIP Similarity}} \\
        \hline
        \textbf{Inverse} & \textbf{Sampling (steps)} &  & \textbf{Distance} $\downarrow \times 10^3$ & \textbf{PSNR} $\uparrow$ & \textbf{LPIPS} $\downarrow \times 10^3$ & \textbf{MSE} $\downarrow \times 10^4$ & \textbf{SSIM} $\uparrow \times 10^2$ & \textbf{Whole} $\uparrow$ & \textbf{Edited} $\uparrow$ \\
        \hline
        VI & DDCM(12) & InfEdit  & 13.78  & 28.51  & 47.58  & 32.09  & 85.66  & 25.03 & 22.22 \\
        VI & DDIM(50) & ViMAEdit  & 12.65  & 28.27  & 44.67  & 30.29  & 85.65  & 25.91 & 22.96 \\
        \hline
        \multirow{6}{*}{PnP-I} & \multirow{6}{*}{DDIM(50)} 
        & P2P-Zero  & 51.13  & 21.23  & 143.87 & 135.00 & 77.23  & 23.36 & 21.03 \\
        &  & MasaCtrl  & 24.47  & 22.78  & 87.38  & 79.91  & 81.36  & 24.42 & 21.38 \\
        &  & PnP  & 24.29  & 22.64  & 106.06 & 80.45  & 79.68  & 25.41 & 22.62 \\
        &  & P2P  & \underline{11.64}  & 27.19  & 54.44  & 33.15  & 84.71  & 25.03 & 22.13 \\
        &  & ViMAEdit  & 11.90  & 28.75  & 43.07  & 28.85  & 85.95  & 25.43 & 22.40 \\
        &  & \textbf{LOCATEdit (Ours)}  & 13.19  & \textbf{29.20}  & \underline{41.60}  & \underline{26.90}  & \textbf{86.53}  & \underline{25.96} & \textbf{23.02} \\
        \hline
        \multirow{4}{*}{EF} & \multirow{4}{*}{DPM-Solver++(20)}
        & LEDITS++  & 23.15  & 24.67  & 80.79  & 118.56  & 81.55  & 25.01 & 22.09 \\
        &  & P2P  & 14.52  & 27.05  & 50.72  & 37.48  & 84.97  & 25.36 & 22.43 \\
        &  & ViMAEdit  & 14.16  & 28.12  & 45.62  & 33.56  & 85.61  & 25.51 & 22.56 \\
        &  & \textbf{LOCATEdit (Ours)} & \textbf{8.71} & \underline{29.16} & \textbf{39.31} & \textbf{24.01} & \underline{86.52} & \textbf{26.07} & 22.43\\
        \hline
    \end{tabular}
    }
    \caption{Comparison of different methods based on structure, background preservation, and CLIP similarity metrics.}
    \label{tab:comparision}
\end{table*}
\endgroup

\begingroup
\renewcommand{\arraystretch}{0.3} 
\begin{table*}[t]
    \centering
    \small 
    \renewcommand{\arraystretch}{1.2} 
    \resizebox{\textwidth}{!}{ 
    \begin{tabular}{c|c|cccc|cc}
        \hline
        \textbf{Method} & \textbf{Structure} & \multicolumn{4}{c|}{\textbf{Background Preservation}} & \multicolumn{2}{c}{\textbf{CLIP Similarity}} \\
        \hline
        & \textbf{Distance} $\downarrow \times 10^3$ & \textbf{PSNR} $\uparrow$ & \textbf{LPIPS} $\downarrow \times 10^3$ & \textbf{MSE} $\downarrow \times 10^4$ & \textbf{SSIM} $\uparrow \times 10^2$ & \textbf{Whole} $\uparrow$ & \textbf{Edited} $\uparrow$ \\
        \hline
        \textbf{\method}  & 13.19  & 29.20  & 41.60  & 26.90  & 86.53  & \textbf{25.96} & \textbf{23.02} \\
        \hline
        w/o diagonal weighting matrix  & 8.68  & \textbf{29.59}  & \textbf{38.16}  & \textbf{23.17}  & \textbf{86.83}  & 25.33 & 22.34 \\
        \hline
        w/o symmetric self-attention  & \textbf{8.59}  & 29.42  & 38.75  & 24.09  & 85.66  & 25.29 & 22.26 \\
        \hline
        w/o $\alpha$-based control  & 8.86  & 29.26  & 38.96  & 24.53  & 86.60  & 25.31 & 22.28 \\
        \hline
        with high $\alpha$  & 20.37  & 24.27  & 87.37  & 60.82  & 82.22  & 26.58 & 23.22 \\
        \hline
    \end{tabular}
    }
    \caption{Comparison of different methods based on structure, background preservation, and CLIP similarity metrics.}
    \label{tab:ablation}
\end{table*}
\endgroup

\subsection{Dataset and Evaluation metrics}
We follow recent work \cite{ju2024pnp, xu2024inversion, wang2024vision} and evaluate our approach using the PIE-Bench dataset \cite{ju2024pnp}, which is currently the only established benchmark designed for prompt-based image editing. PIE-Bench contains 700 images categorized into ten different editing tasks, with each image accompanied by a source prompt, a target prompt, blend words (i.e., terms that specify the required edits), and an editing mask. Although only the source prompt, target prompt, and blend words are necessary for performing prompt-based editing, the editing mask is employed to gauge how well the method preserves the background.

To thoroughly assess our models, we adopt the evaluation strategy described in \cite{ju2024pnp}, focusing on three main criteria: 1) \textit{Structure consistency}, measured by the difference in DINO self-similarity maps \cite{tumanyan2022splicing}, 2) \textit{Background preservation}, evaluated via PSNR, LPIPS \cite{zhang2018unreasonable}, MSE, and SSIM \cite{wang2004image}, and 3) \textit{Target prompt--image alignment}, determined by CLIP similarity \cite{hessel2021clipscore}.

\subsection{Comparison with existing methods}
\method consistently yields superior spatial consistency and semantic alignment compared to state-of-the-art text-guided image editing methods as can be seen in Table \ref{tab:comparision}. Unlike P2P-Zero \cite{parmar2023zero} and PnP-based techniques \cite{tumanyan2022splicing, ju2024pnp}, which tend to induce global modifications and suffer from spatial inconsistencies, \method confines edits to intended regions, thereby preserving the source image’s structure. Mask-guided methods such as ViMAEdit \cite{wang2024vision} improve localization but can still introduce artifacts in non-target areas. Our graph Laplacian regularization refines cross-attention maps by enforcing smooth, coherent patch-to-patch relationships, addressing these issues directly.

Moreover, while approaches like Edit-Friendly Inversion \cite{huberman2024edit} and InfEdit with Virtual Inversion \cite{xu2024inversion} achieve better semantic alignment, they often struggle to disentangle editable regions from the preserved background. In contrast, our method robustly separates these regions, ensuring that modifications are both precise and localized, which can be seen in qualitative comparision we provide in Table \ref{fig:comparison}.

Overall, our experiments demonstrate that our method not only enhances the fidelity of the edited regions but also maintains the overall structural integrity of the source image. By leveraging CASA graph-based attention refinement, our approach outperforms existing techniques across multiple metrics, underscoring the importance of spatially consistent and disentangled editing for practical text-controlled image editing applications.

\subsection{Ablation Study}
To demonstrate the effectiveness of our model, we provide results for three different model ablations: 1) \textbf{w/o diagonal weighting matrix}: we use uniform $L^2$ penalty as the first term of Equation \ref{eq:optimization}, 2) \textbf{w/o symmetric self-attention}: We do not parameterize the similarity matrix $\mathbf{S}$ which is essential for the Laplacian to be positive semidefinite, and 3) \textbf{w/o $\alpha$-based control}: We keep $\alpha=1$ in Equation \ref{eq:sigmoid_weights}.

Table \ref{tab:ablation} shows that each of our contribution outperforms the baselines in terms of Structure and Background preservation. We also observe that while combining different techniques together results in a slightly worser results in Structure and Background Similarity metrics, we are able to achieve state-of-the-art CLIP Similarity. It is to be noted that even the worser results are better than all the baseline methods reported in Table \ref{tab:comparision}. Finally, when we were tuning the $\alpha$ parameter, we observed that a higher value of $\alpha$ edits images with way better CLIP similarity but significantly worsens the results for other metrics. This is to be expected because a high $\alpha$ results in ``hard thresholding" where it makes a clear distinction between areas that should be trusted and those that should be adjusted, but it also leads to abrupt transitions.

\section{Conclusion}
In this paper, we introduced a text-controlled image editing framework \method that refines cross-attention masks using graph Laplacian regularization. It leverages self-attention-derived patch relationships to enforce spatial consistency and localized, disentangled modifications while preserving the structural integrity of the source image. Extensive experiments demonstrate that our approach outperforms state-of-the-art methods in semantic alignment and background fidelity. By confining edits to intended regions, our technique avoids unwanted alterations and maintains overall coherence. Future work will extend this framework to non-symmetric regularization and more complex editing scenarios, further enhancing controllable image generation.

{
    \small
    \bibliographystyle{ieeenat_fullname}
    \bibliography{main}
}

\clearpage


\begin{onecolumn}
\setcounter{page}{1}

\maketitlesupplementary
\section{Broader Impact}
Our work advances the precision of text-guided image editing by ensuring that modifications are both spatially consistent and semantically faithful. This improvement has the potential to benefit a wide range of applications—from enhancing creative workflows in digital art and advertising to supporting critical tasks in medical imaging and scientific visualization—by reducing the need for extensive manual post-processing. At the same time, the increased reliability of automated editing tools underscores the importance of establishing robust ethical guidelines for their use, particularly in contexts where the authenticity of visual information is paramount. By delivering a method that better preserves the structural integrity of the source images, our approach paves the way for more trustworthy and accessible image editing solutions that can democratize creative technologies and support various high-stakes applications.

\section{Proof of Lemma 1}
\label{proof:lemma1}
\begin{proof}
To prove that $\mathbf{L}$ is PSD, we must show that for any $\mathbf{x} \in \mathbb{R}^n$, the quadratic form $\mathbf{x}^\top \mathbf{L} \mathbf{x}$ is nonnegative:
\[
\mathbf{x}^\top \mathbf{L} \mathbf{x} = \mathbf{x}^\top (\mathbf{D} - \mathbf{S_{\text{sym}}}) \mathbf{x}.
\]
Expanding this expression, we have:
\[
\mathbf{x}^\top \mathbf{L} \mathbf{x} = \mathbf{x}^\top \mathbf{D} \mathbf{x} - \mathbf{x}^\top \mathbf{S_{\text{sym}}} \mathbf{x}.
\]

The degree matrix $\mathbf{D}$ is diagonal, with entries $\mathbf{D}(i, i) = \sum_{j=1}^n \mathbf{S_{\text{sym}}}(i, j)$. Therefore:
\[
\mathbf{x}^\top \mathbf{D} \mathbf{x} = \sum_{i=1}^n \mathbf{D}(i, i) x_i^2 = \sum_{i=1}^n \left( \sum_{j=1}^n \mathbf{S_{\text{sym}}}(i, j) \right) x_i^2.
\]

The second term, $\mathbf{x}^\top \mathbf{S_{\text{sym}}} \mathbf{x}$, is given by:
\[
\mathbf{x}^\top \mathbf{S_{\text{sym}}} \mathbf{x} = \sum_{i=1}^n \sum_{j=1}^n \mathbf{S_{\text{sym}}}(i, j) x_i x_j.
\]

Substituting these into the quadratic form, we get:
\[
\mathbf{x}^\top \mathbf{L} \mathbf{x} = \sum_{i=1}^n \left( \sum_{j=1}^n \mathbf{S_{\text{sym}}}(i, j) x_i^2 \right) - \sum_{i=1}^n \sum_{j=1}^n \mathbf{S_{\text{sym}}}(i, j) x_i x_j.
\]

Reorganizing terms:
\[
\mathbf{x}^\top \mathbf{L} \mathbf{x} = \frac{1}{2} \sum_{i=1}^n \sum_{j=1}^n \mathbf{S_{\text{sym}}}(i, j) \left( x_i^2 + x_j^2 - 2 x_i x_j \right).
\]

This simplifies to:
\[
\mathbf{x}^\top \mathbf{L} \mathbf{x} = \frac{1}{2} \sum_{i=1}^n \sum_{j=1}^n \mathbf{S_{\text{sym}}}(i, j) (x_i - x_j)^2.
\]

Since $\mathbf{S_{\text{sym}}}(i, j) \geq 0$ (by definition of the symmetrized self-attention matrix) and $(x_i - x_j)^2 \geq 0$, every term in the summation is nonnegative. Therefore:
\[
\mathbf{x}^\top \mathbf{L} \mathbf{x} \geq 0 \quad \forall \mathbf{x} \in \mathbb{R}^n.
\]

Thus, $\mathbf{L}$ is positive semidefinite.
\end{proof}

\section{Proof of Theorem 1}\label{proof:theorem1}

\subsection{Optimization Problem}
We consider the following optimization problem:
\begin{equation}\label{eq:optimization}
    \min_{x\in\mathbb{R}^{R^2}} J(x),
\end{equation}
where the objective function is defined as
\begin{equation*}
    J(x) = (x - x^{(0)})^\top \mathbf{\Lambda} (x - x^{(0)}) + \lambda\, x^\top L\, x.
\end{equation*}
Here, the \emph{fidelity} term \((x - x^{(0)})^\top \mathbf{\Lambda} (x - x^{(0)})\) penalizes deviations from the initial mask \(x^{(0)}\) with stronger penalties in regions of higher confidence (as encoded by the diagonal weight matrix \(\mathbf{\Lambda}\)). The \emph{smoothness} term \(\lambda\, x^\top L\, x\) promotes a spatially coherent solution by enforcing that the mask varies smoothly across similar patches, as determined by the self-attention structure. The hyperparameter \(\lambda > 0\) balances the trade-off between fidelity and smoothness.

\subsection{Existence and Uniqueness of the Solution}
To obtain the refined mask, we solve the minimization problem in Equation~\eqref{eq:optimization}. The first term is strictly convex since \(\mathbf{\Lambda}\) is positive definite, and the second term is convex because \(L\) is positive semidefinite. Thus, the overall objective \(J(x)\) is strictly convex and has a unique minimizer.

Taking the gradient with respect to \(x\) yields:
\begin{equation}
    \nabla J(x) = 2\,\mathbf{\Lambda} (x - x^{(0)}) + 2\lambda\, L\, x.
\end{equation}
Setting \(\nabla J(x)=0\) gives:
\begin{equation}
    \mathbf{\Lambda} (x - x^{(0)}) + \lambda\, L\, x = 0.
\end{equation}
Rearranging, we obtain:
\begin{equation}
    \left(\mathbf{\Lambda} + \lambda\, L\right) x = \mathbf{\Lambda}\, x^{(0)}.
\end{equation}
Since \(\mathbf{\Lambda} + \lambda\, L\) is positive definite, it is invertible, and the unique solution is
\[
    x^* = \left(\mathbf{\Lambda} + \lambda\, L\right)^{-1} \mathbf{\Lambda}\, x^{(0)}.
\]
The positive semidefiniteness of \(L\) ensures the convexity of the regularization term, thereby guaranteeing the existence and uniqueness of the solution.

\subsection{Additional Qualitative Results}
This section presents qualitative results for refined masks achieved through graph Laplacian regularization and compares the editing outcomes with existing image editing methods.

\begin{figure*}
    \centering
    \captionsetup{justification=centering}
    \includegraphics[width=\textwidth]{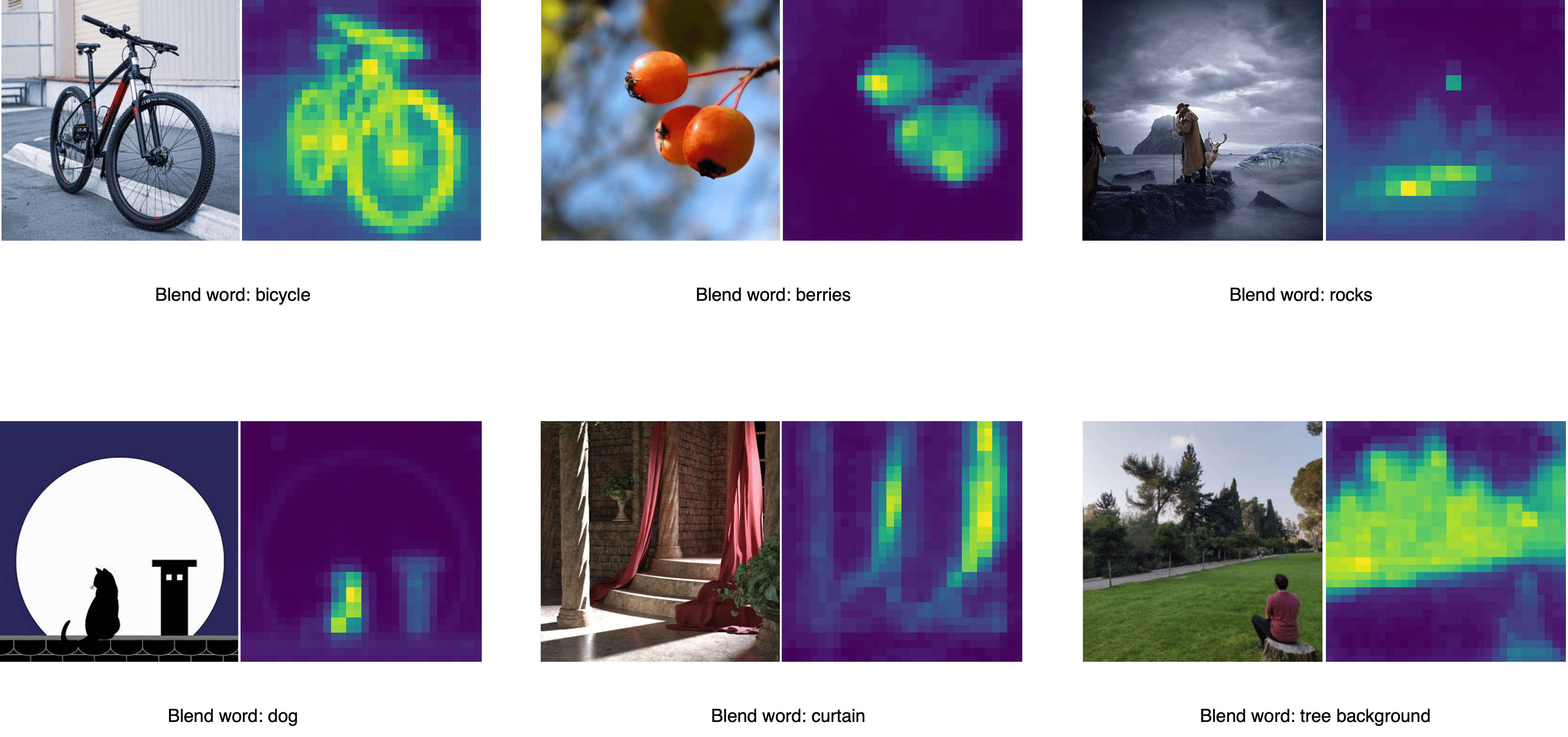}
    \caption{Refined masks after Graph Laplacian Regularization}
    \label{fig:masks}
\end{figure*}

\begin{table*}[!h]
\centering
\resizebox{\textwidth}{!}{ 
\begin{tabular}{>{\centering\arraybackslash}m{3cm}
                >{\centering\arraybackslash}m{2.5cm}:
                >{\centering\arraybackslash}m{2.5cm}
                >{\centering\arraybackslash}m{2.5cm}
                >{\centering\arraybackslash}m{2.5cm}
                >{\centering\arraybackslash}m{2.5cm}
                >{\centering\arraybackslash}m{2.5cm}}
 & \textbf{Source Image} & \textbf{LOCATEdit} & \textbf{ViMAEdit} & \textbf{InfEdit} & \textbf{MasaCtrl} & \textbf{LEDITS++} \\ 

a photo of \textcolor{red}{\sout{goat}} \textbf{horse} and a cat standing on rocks near the ocean
& \includegraphics[width=\linewidth,height=2.5cm,keepaspectratio]{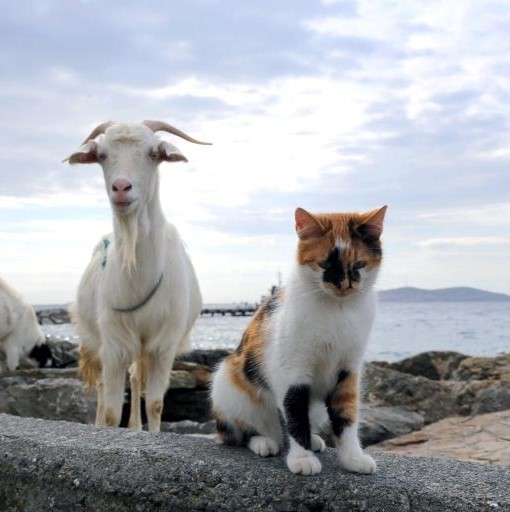} 
& \includegraphics[width=\linewidth,height=2.5cm,keepaspectratio]{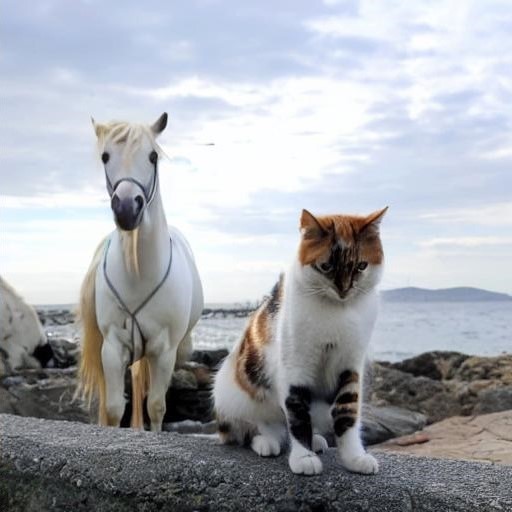} 
& \includegraphics[width=\linewidth,height=2.5cm,keepaspectratio]{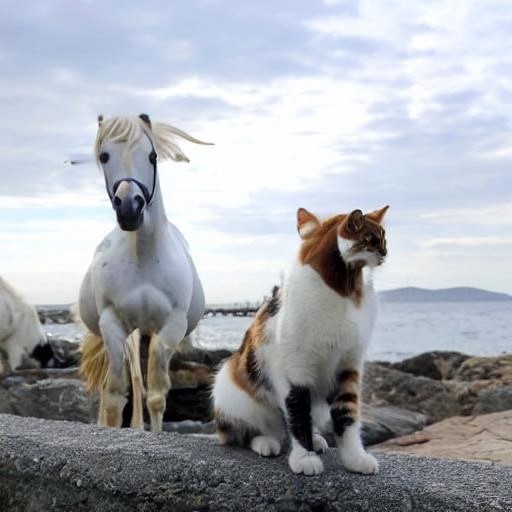} 
& \includegraphics[width=\linewidth,height=2.5cm,keepaspectratio]{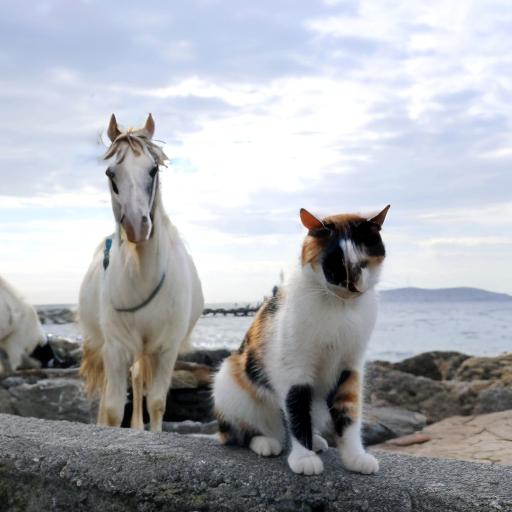} 
& \includegraphics[width=\linewidth,height=2.5cm,keepaspectratio]{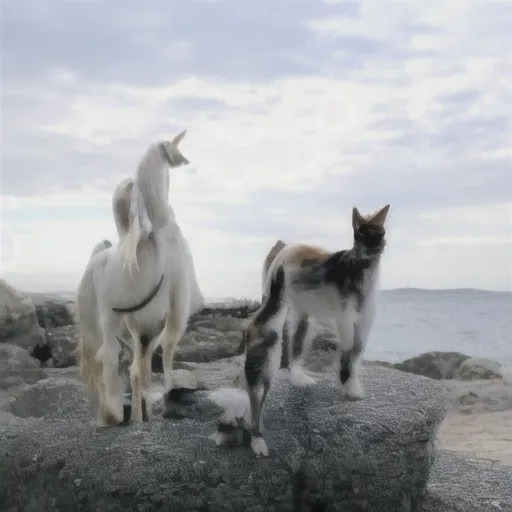} 
& \includegraphics[width=\linewidth,height=2.5cm,keepaspectratio]{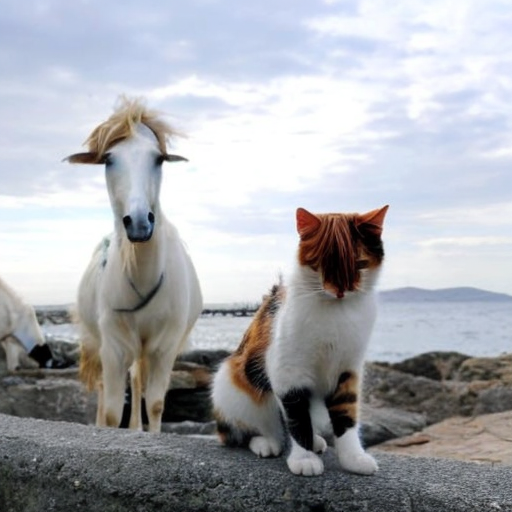} \\ 

a \textcolor{red}{\sout{brown}} \textbf{white} tea cup and a book
& \includegraphics[width=\linewidth,height=2.5cm,keepaspectratio]{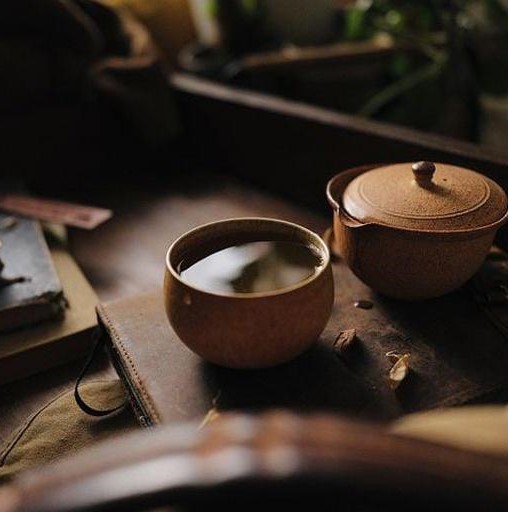} 
& \includegraphics[width=\linewidth,height=2.5cm,keepaspectratio]{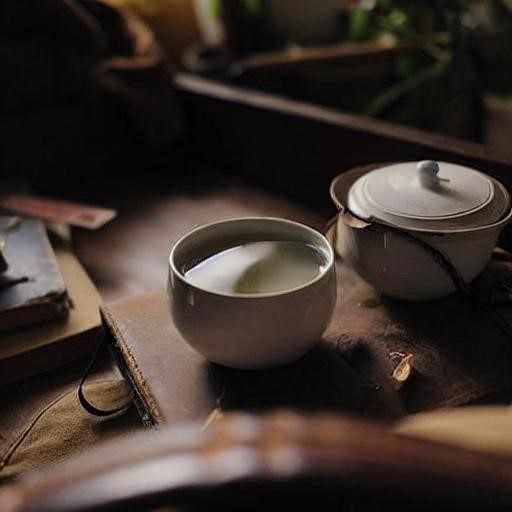} 
& \includegraphics[width=\linewidth,height=2.5cm,keepaspectratio]{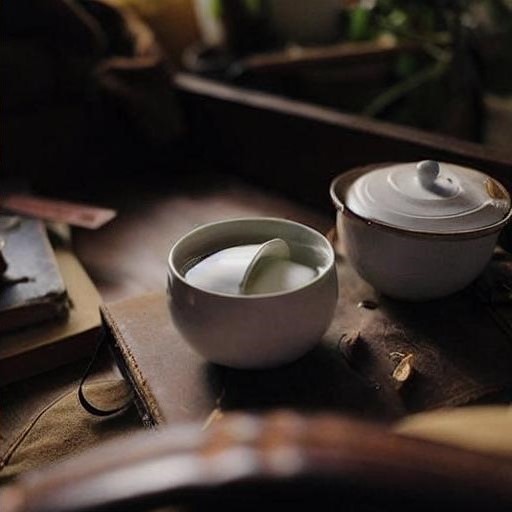} 
& \includegraphics[width=\linewidth,height=2.5cm,keepaspectratio]{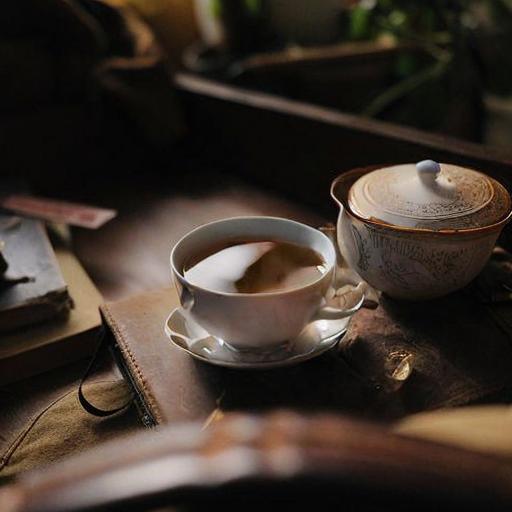} 
& \includegraphics[width=\linewidth,height=2.5cm,keepaspectratio]{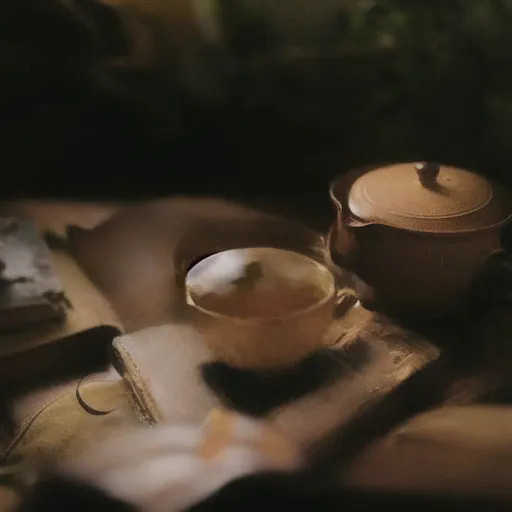} 
& \includegraphics[width=\linewidth,height=2.5cm,keepaspectratio]{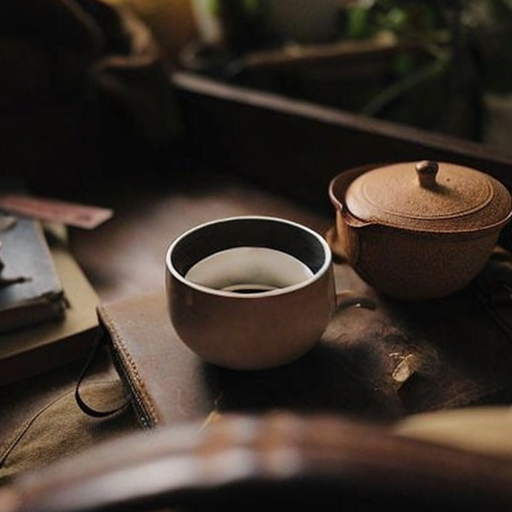}  \\ 

a cat sitting in the \textcolor{red}{\sout{grass}} \textbf{rocks}
& \includegraphics[width=\linewidth,height=2.5cm,keepaspectratio]{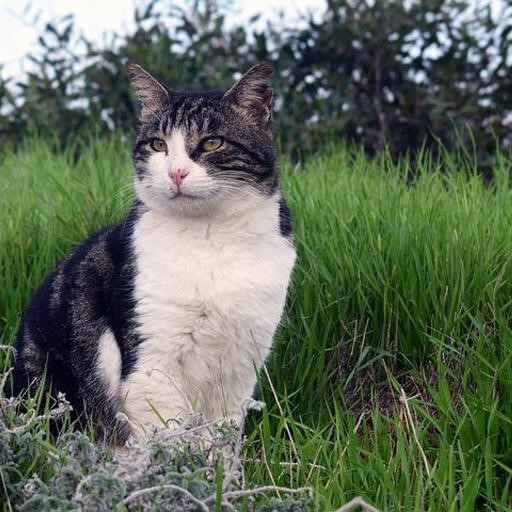} 
& \includegraphics[width=\linewidth,height=2.5cm,keepaspectratio]{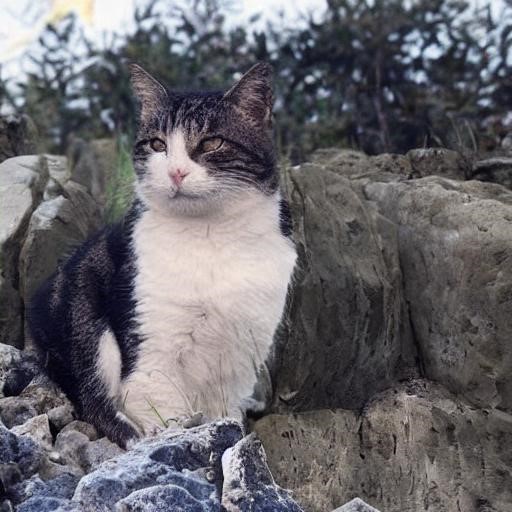} 
& \includegraphics[width=\linewidth,height=2.5cm,keepaspectratio]{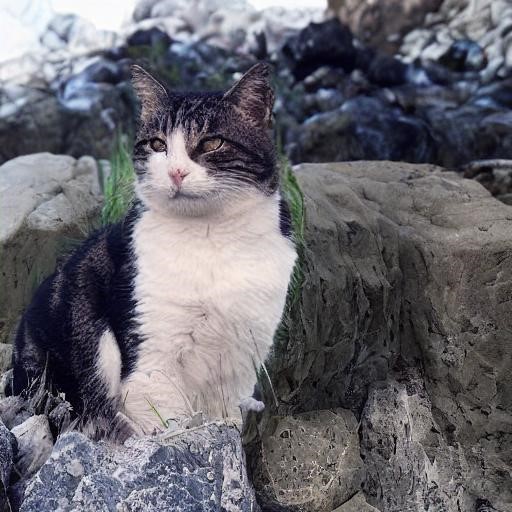} 
& \includegraphics[width=\linewidth,height=2.5cm,keepaspectratio]{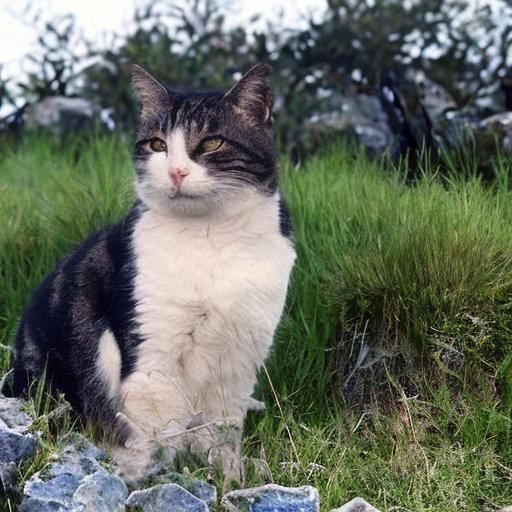} 
& \includegraphics[width=\linewidth,height=2.5cm,keepaspectratio]{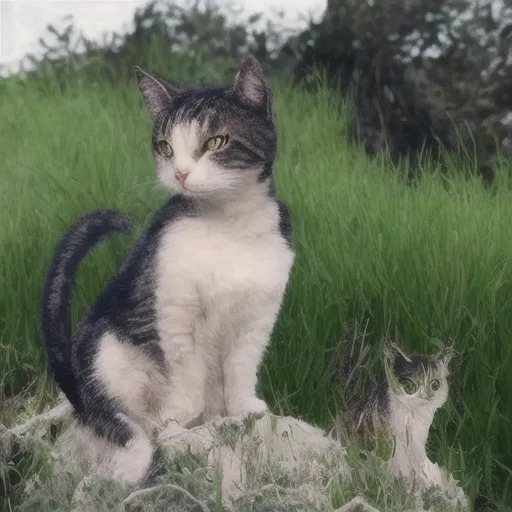} 
& \includegraphics[width=\linewidth,height=2.5cm,keepaspectratio]{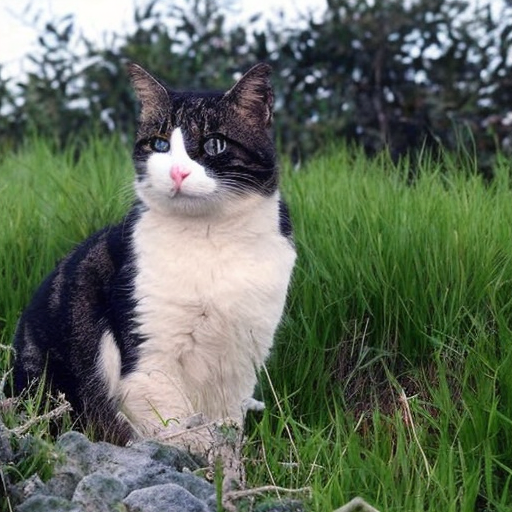}  \\ 

a woman with black hair and a white shirt is holding a \textcolor{red}{\sout{phone}} \textbf{coffee}
& \includegraphics[width=\linewidth,height=2.5cm,keepaspectratio]{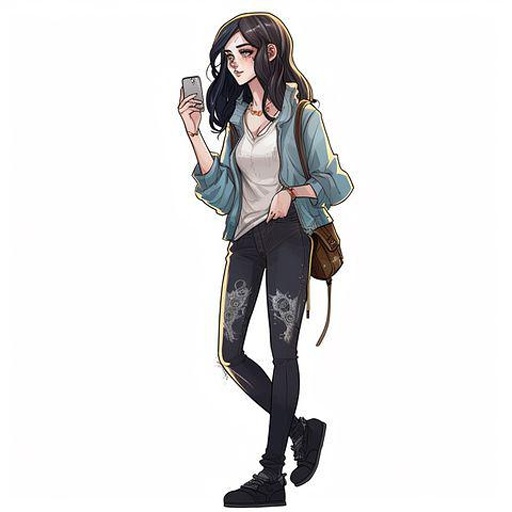} 
& \includegraphics[width=\linewidth,height=2.5cm,keepaspectratio]{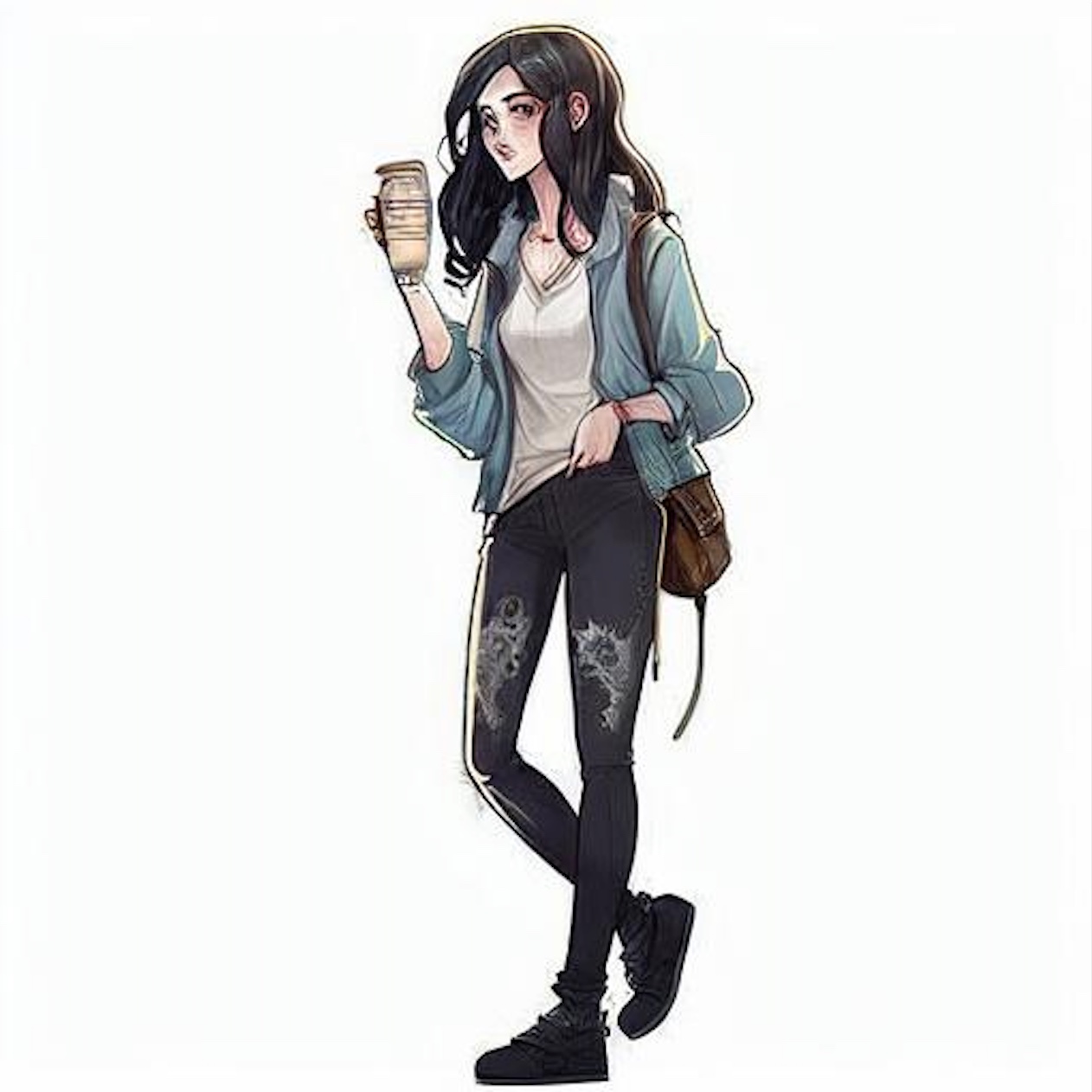} 
& \includegraphics[width=\linewidth,height=2.5cm,keepaspectratio]{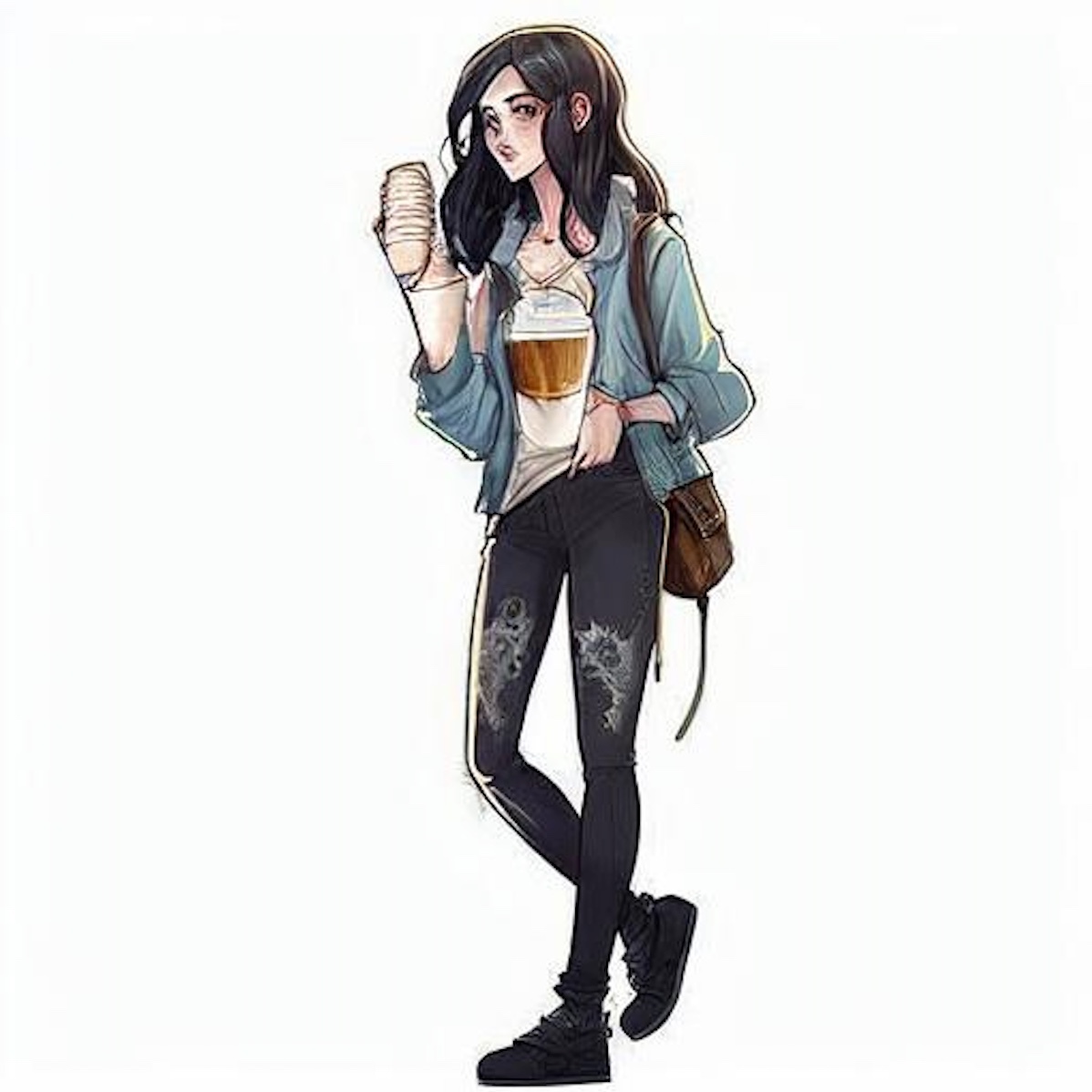} 
& \includegraphics[width=\linewidth,height=2.5cm,keepaspectratio]{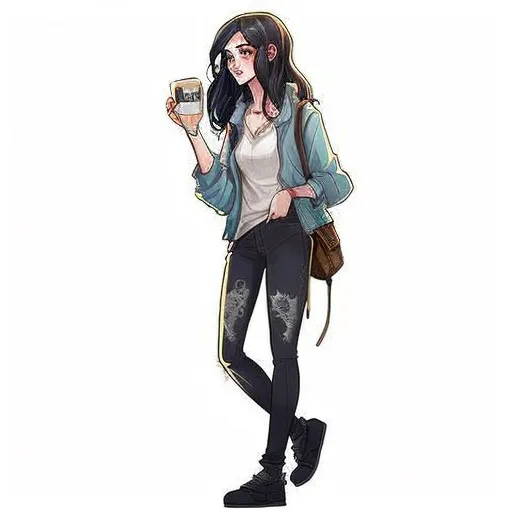} 
& \includegraphics[width=\linewidth,height=2.5cm,keepaspectratio]{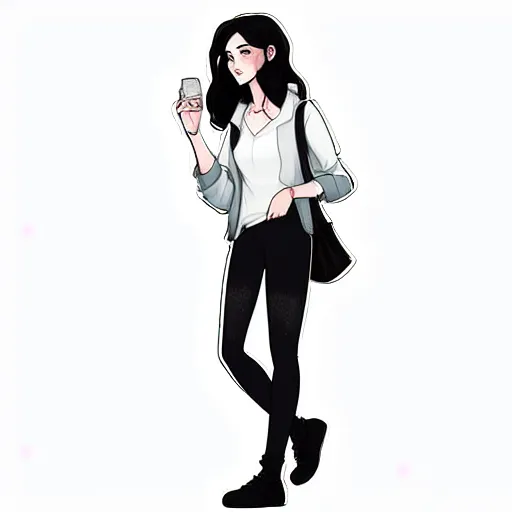} 
& \includegraphics[width=\linewidth,height=2.5cm,keepaspectratio]{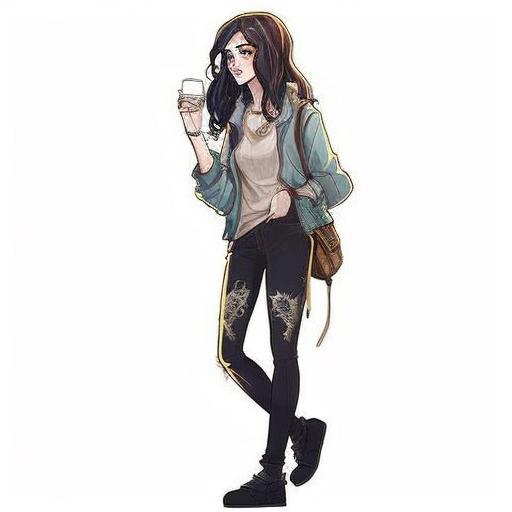}  \\ 

\textcolor{red}{\sout{sea}} \textbf{forest} and house
& \includegraphics[width=\linewidth,height=2.5cm,keepaspectratio]{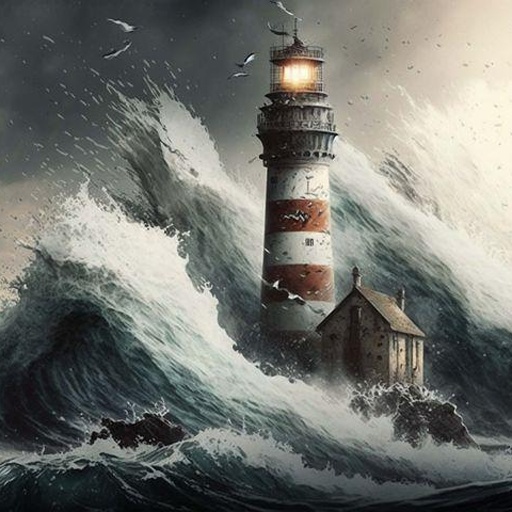} 
& \includegraphics[width=\linewidth,height=2.5cm,keepaspectratio]{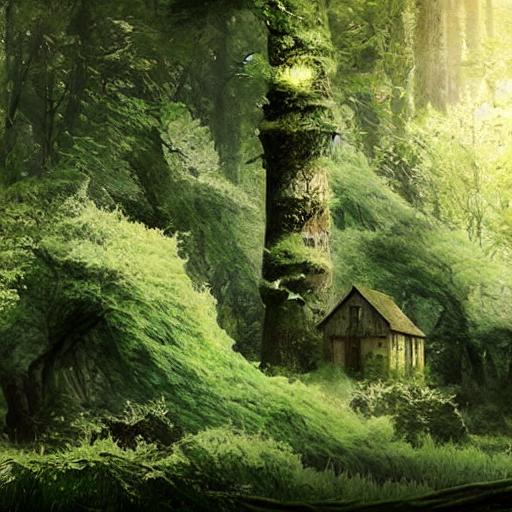} 
& \includegraphics[width=\linewidth,height=2.5cm,keepaspectratio]{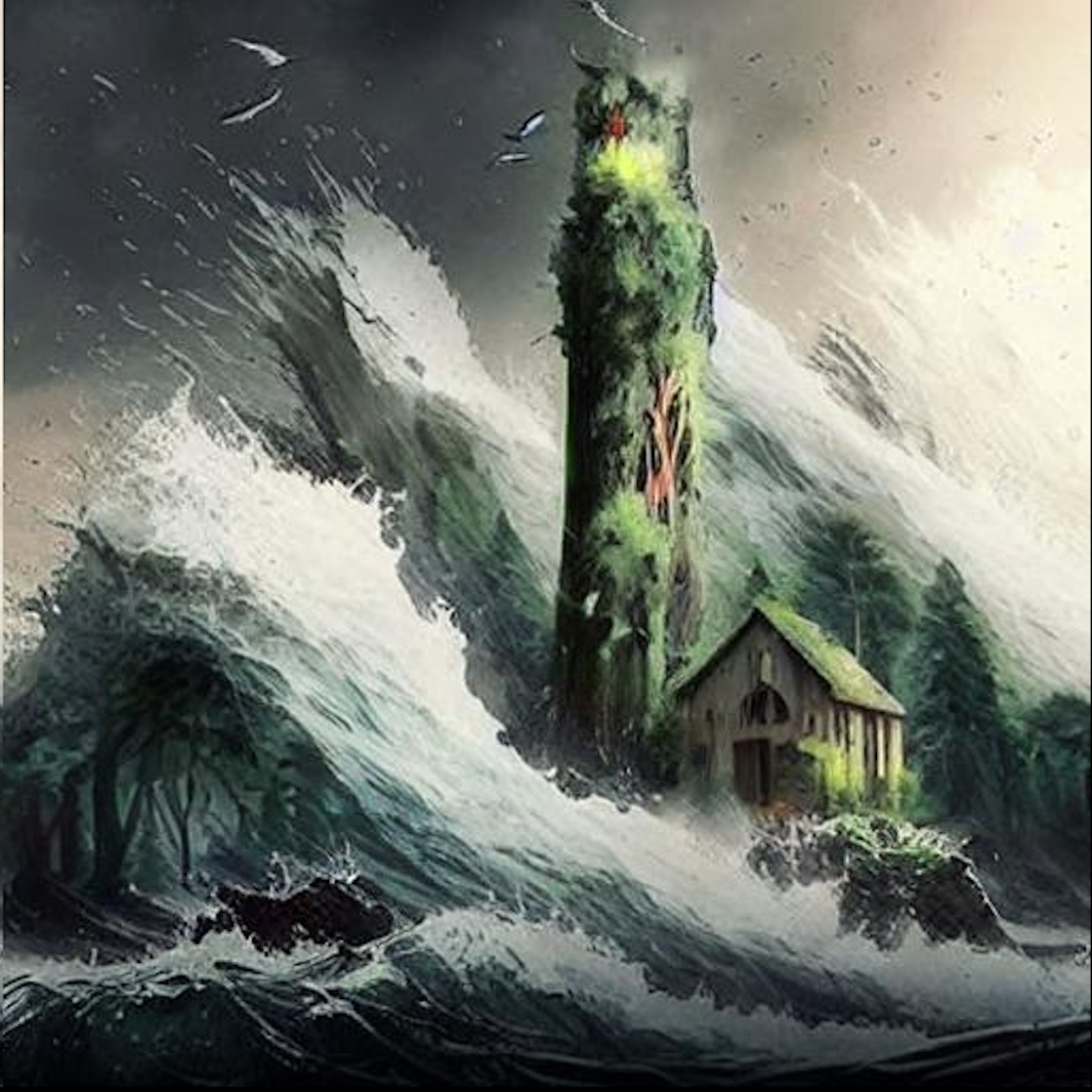} 
& \includegraphics[width=\linewidth,height=2.5cm,keepaspectratio]{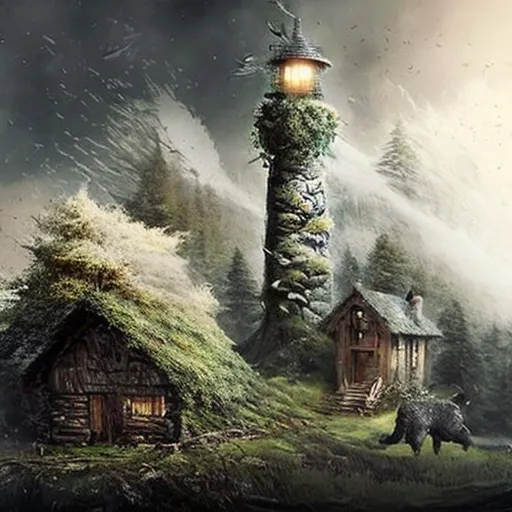} 
& \includegraphics[width=\linewidth,height=2.5cm,keepaspectratio]{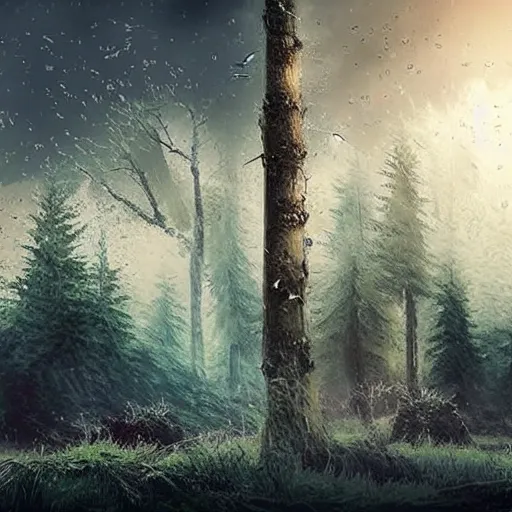} 
& \includegraphics[width=\linewidth,height=2.5cm,keepaspectratio]{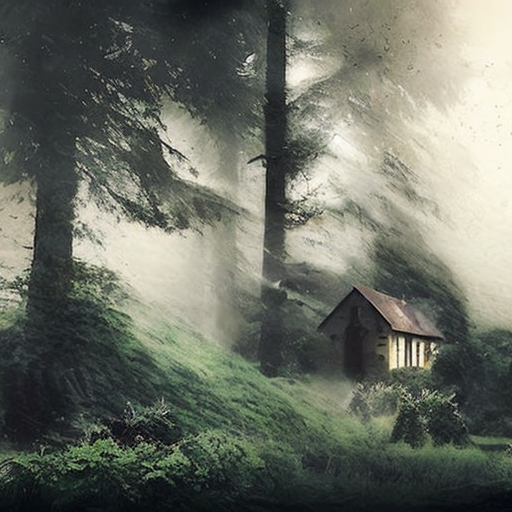}  \\ 

a cute little \textcolor{red}{\sout{duck}} \textbf{marmot} with big eyes
& \includegraphics[width=\linewidth,height=2.5cm,keepaspectratio]{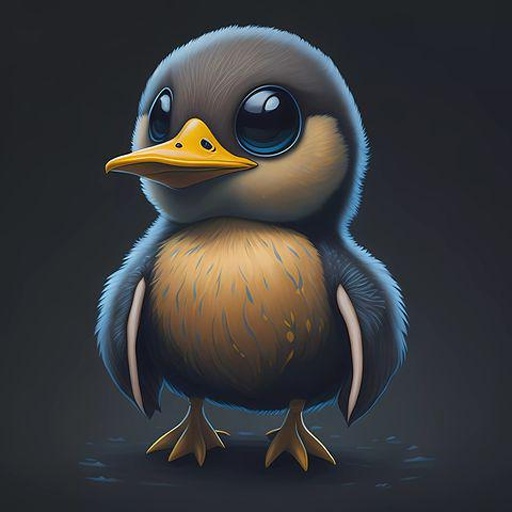} 
& \includegraphics[width=\linewidth,height=2.5cm,keepaspectratio]{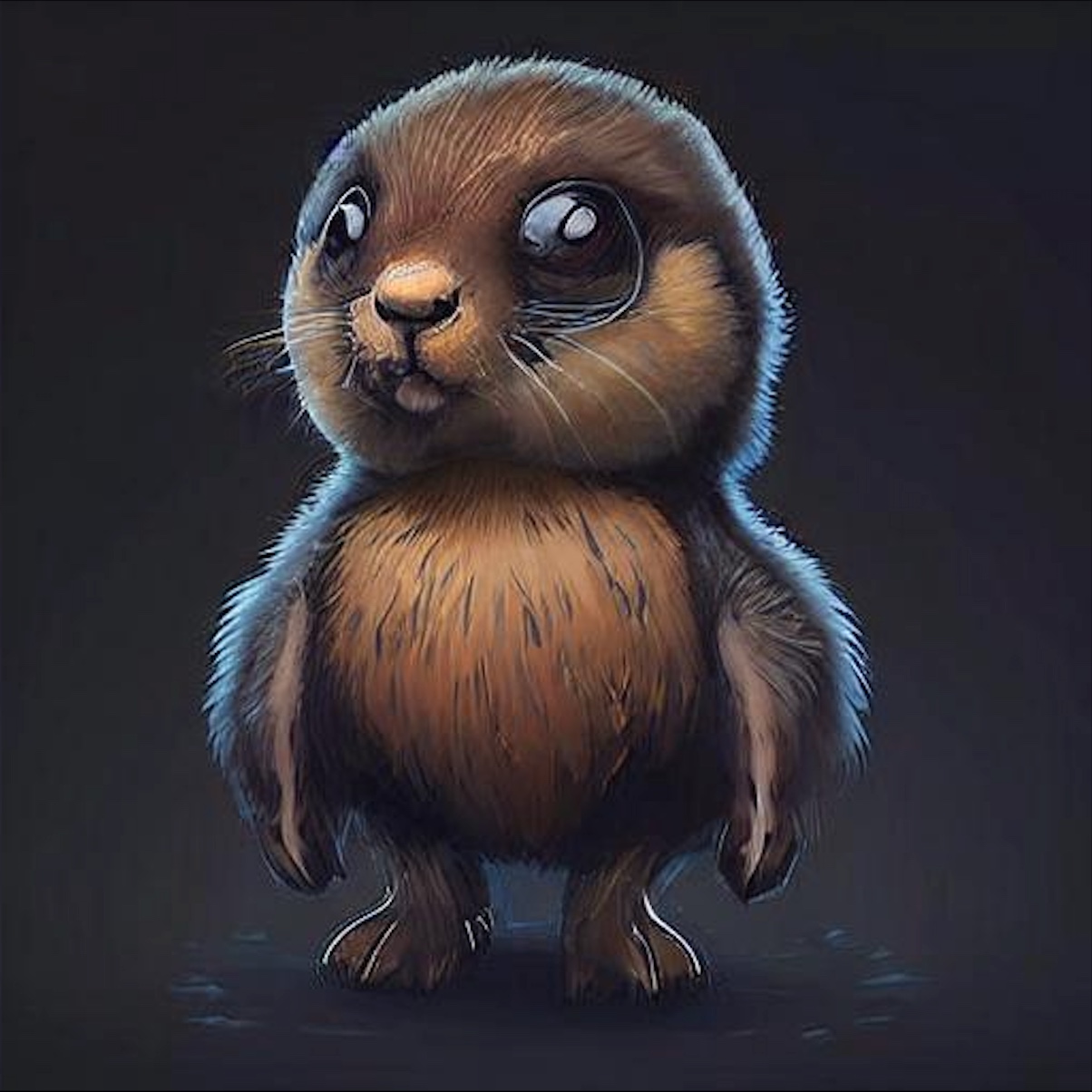} 
& \includegraphics[width=\linewidth,height=2.5cm,keepaspectratio]{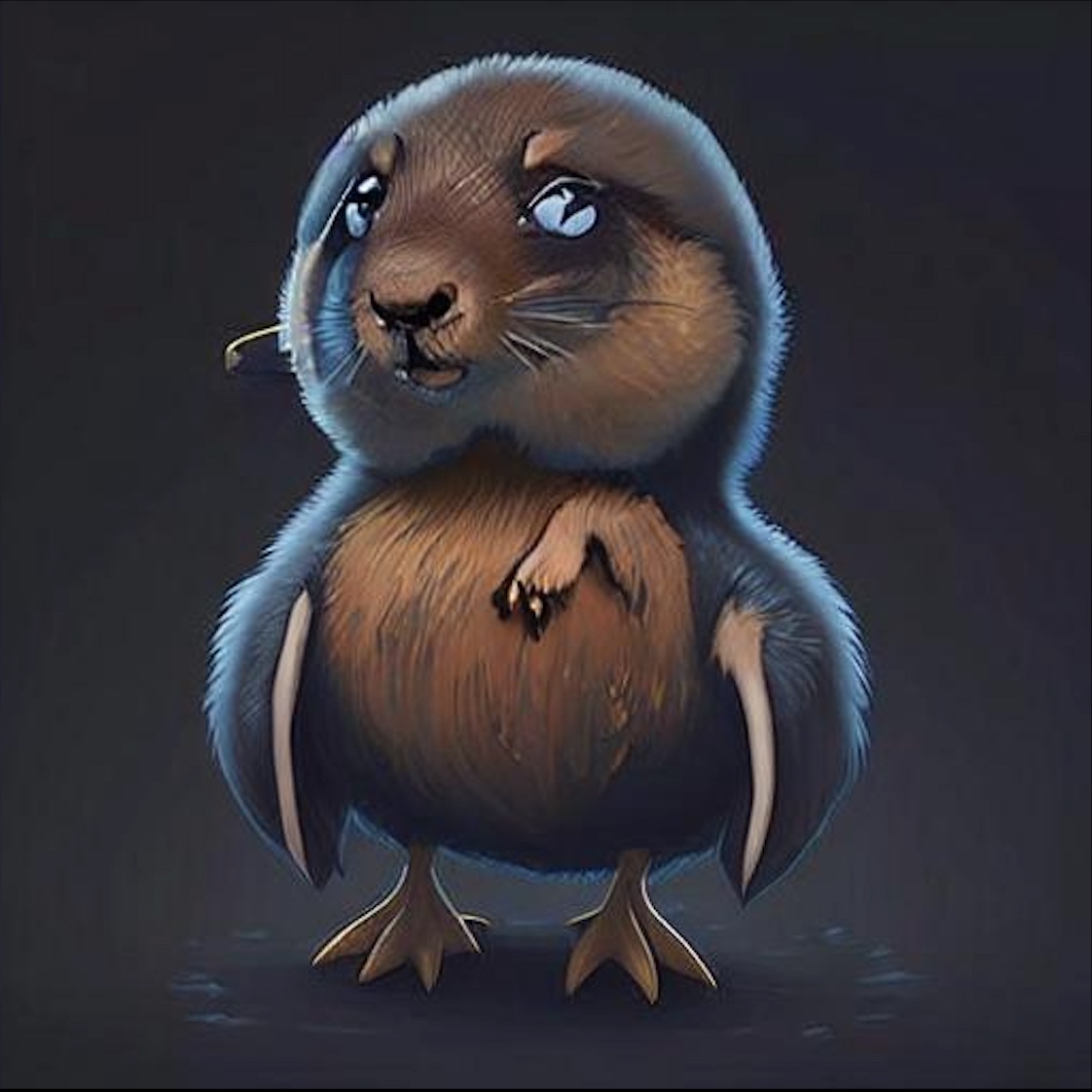} 
& \includegraphics[width=\linewidth,height=2.5cm,keepaspectratio]{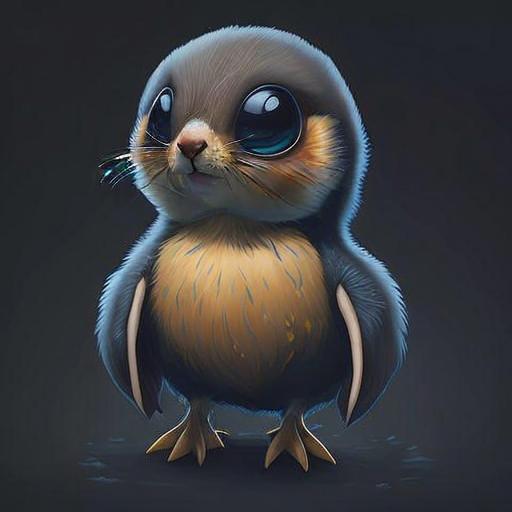} 
& \includegraphics[width=\linewidth,height=2.5cm,keepaspectratio]{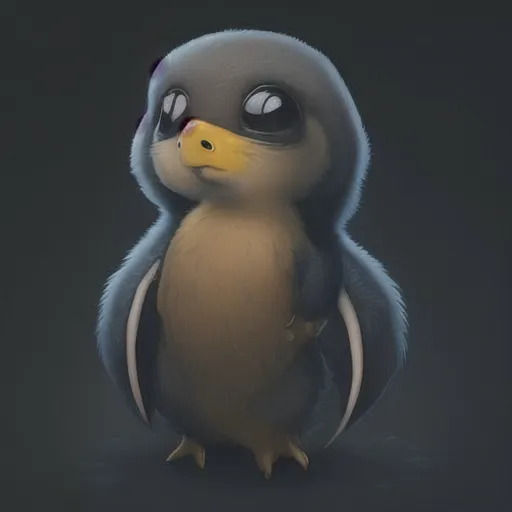} 
& \includegraphics[width=\linewidth,height=2.5cm,keepaspectratio]{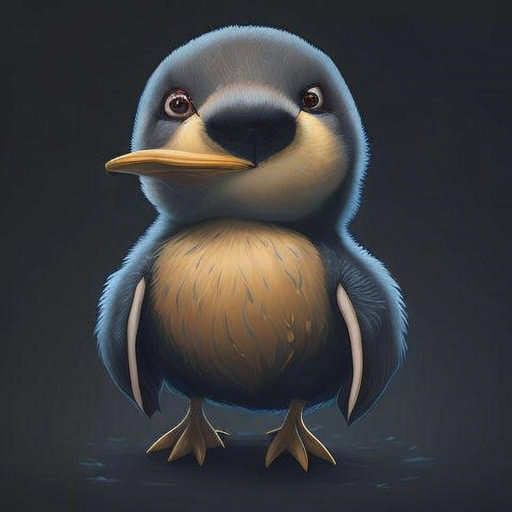}  \\ 

a woman with \textcolor{red}{\sout{flowers}} \textbf{monster} around her face
& \includegraphics[width=\linewidth,height=2.5cm,keepaspectratio]{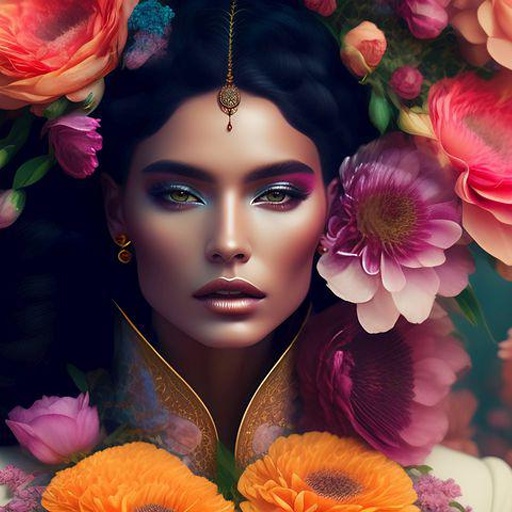} 
& \includegraphics[width=\linewidth,height=2.5cm,keepaspectratio]{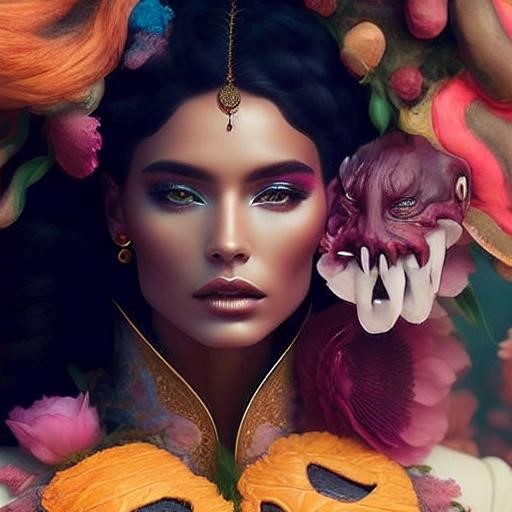} 
& \includegraphics[width=\linewidth,height=2.5cm,keepaspectratio]{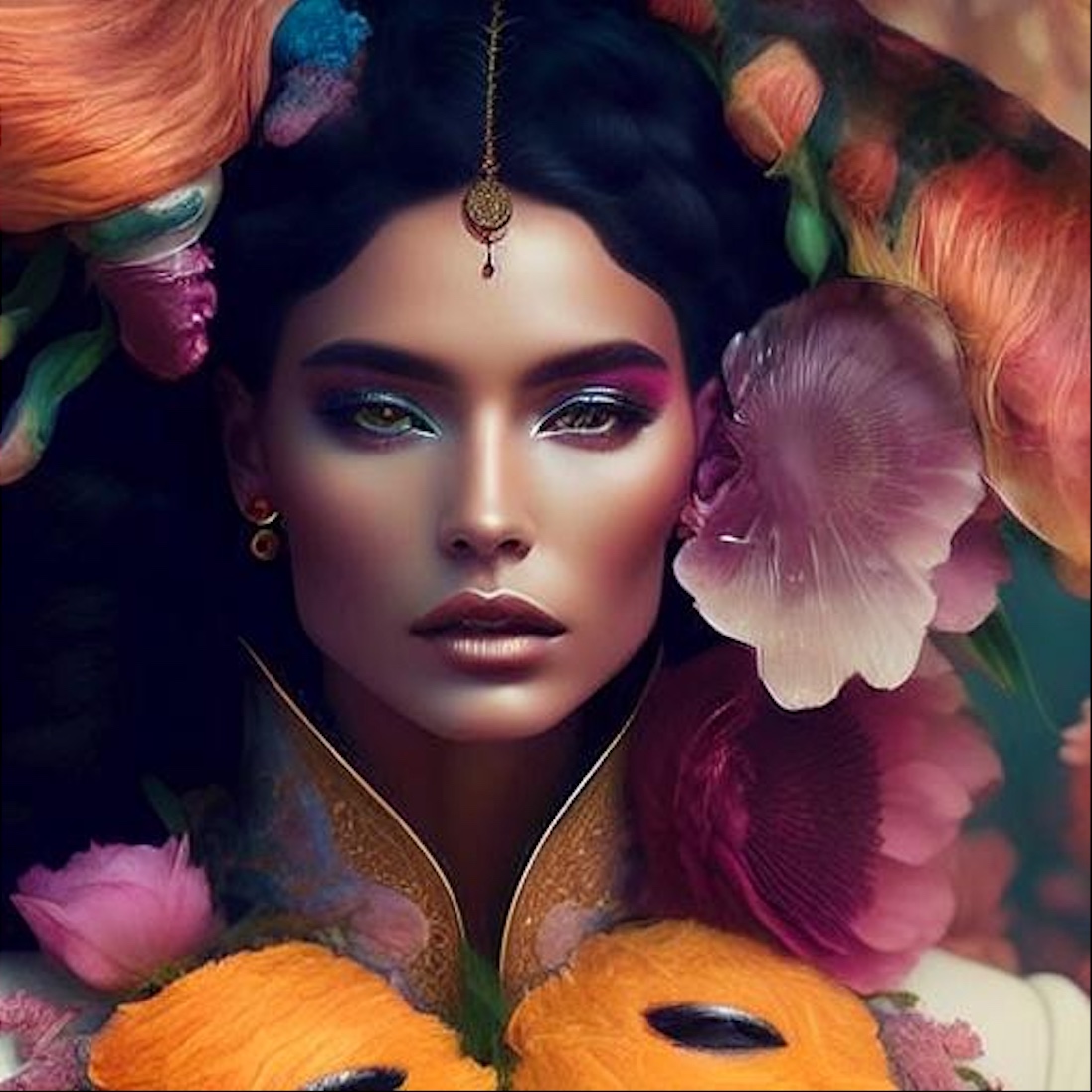} 
& \includegraphics[width=\linewidth,height=2.5cm,keepaspectratio]{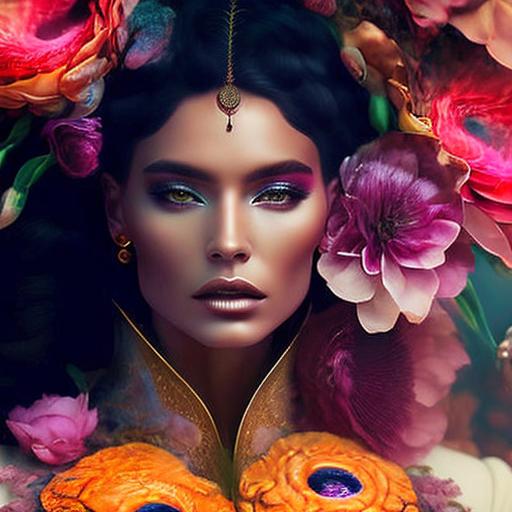} 
& \includegraphics[width=\linewidth,height=2.5cm,keepaspectratio]{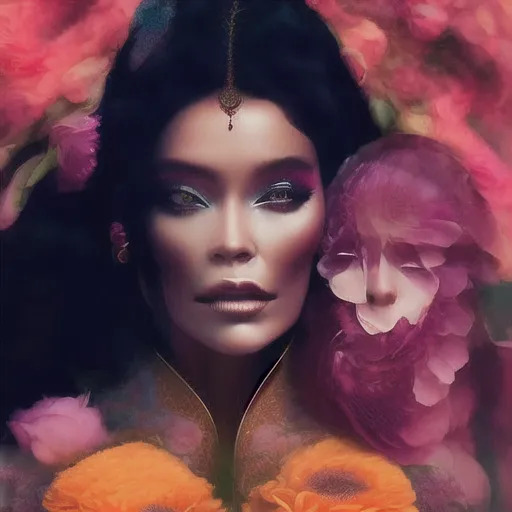} 
& \includegraphics[width=\linewidth,height=2.5cm,keepaspectratio]{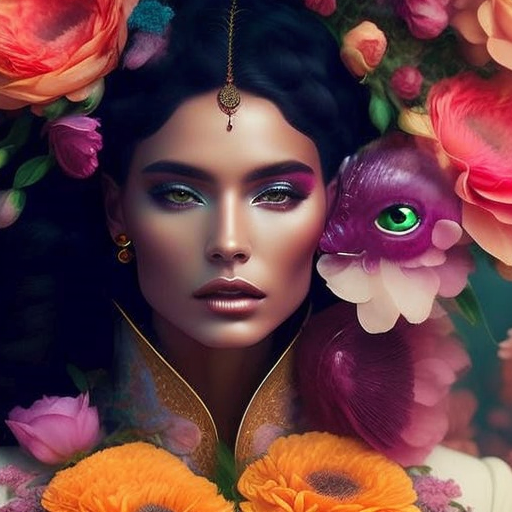}  \\ 

the two people are standing on \textcolor{red}{\sout{rocks}} \textbf{boat} with a fish
& \includegraphics[width=\linewidth,height=2.5cm,keepaspectratio]{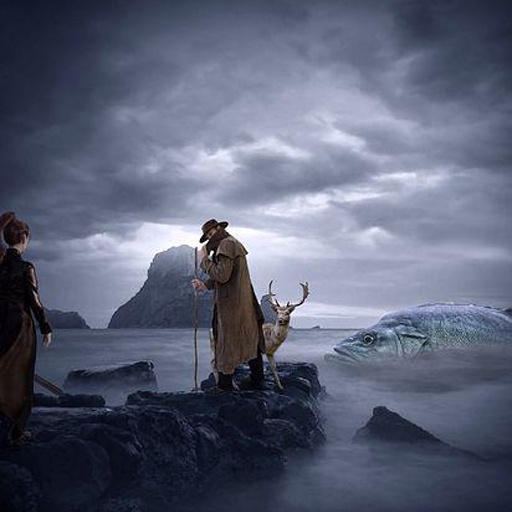} 
& \includegraphics[width=\linewidth,height=2.5cm,keepaspectratio]{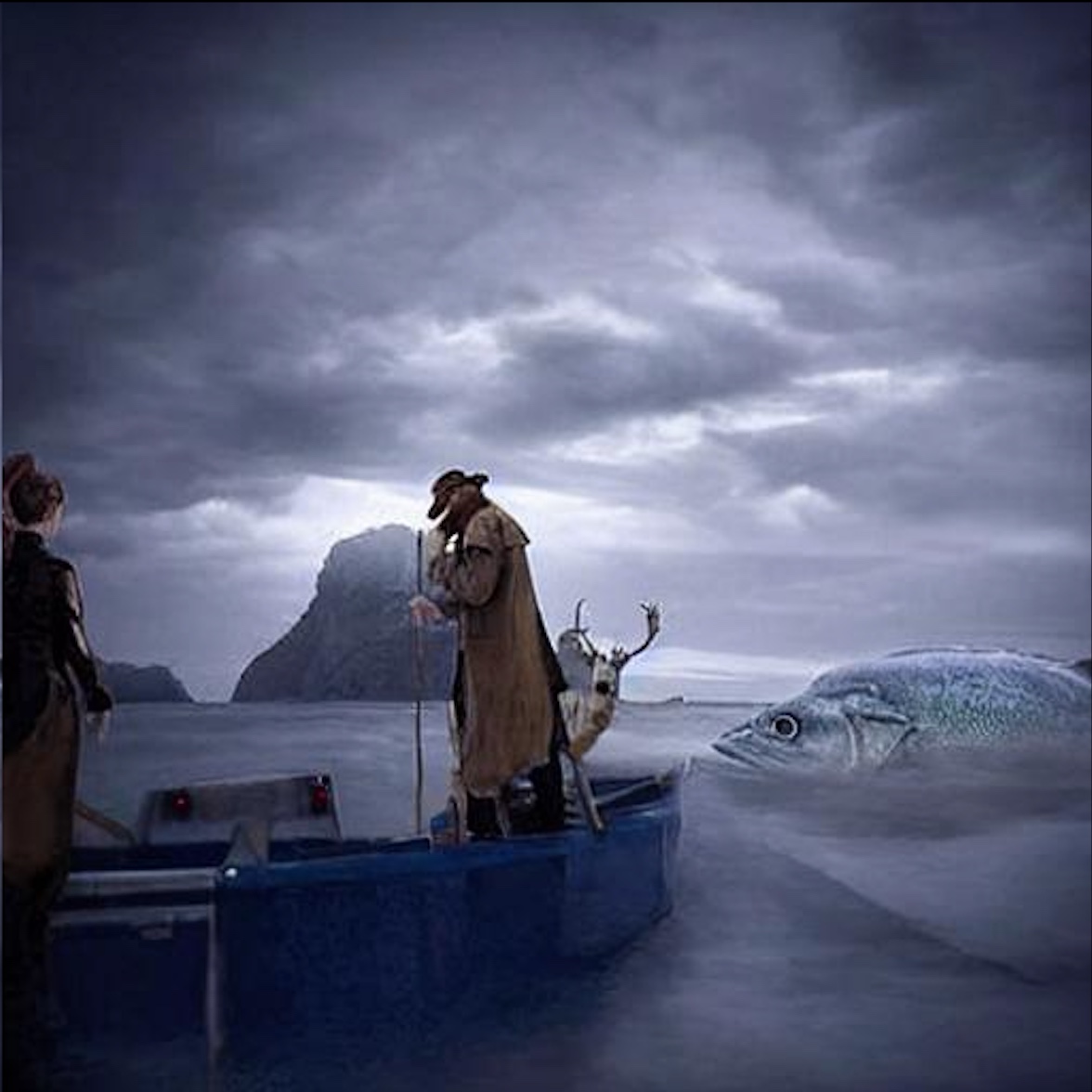} 
& \includegraphics[width=\linewidth,height=2.5cm,keepaspectratio]{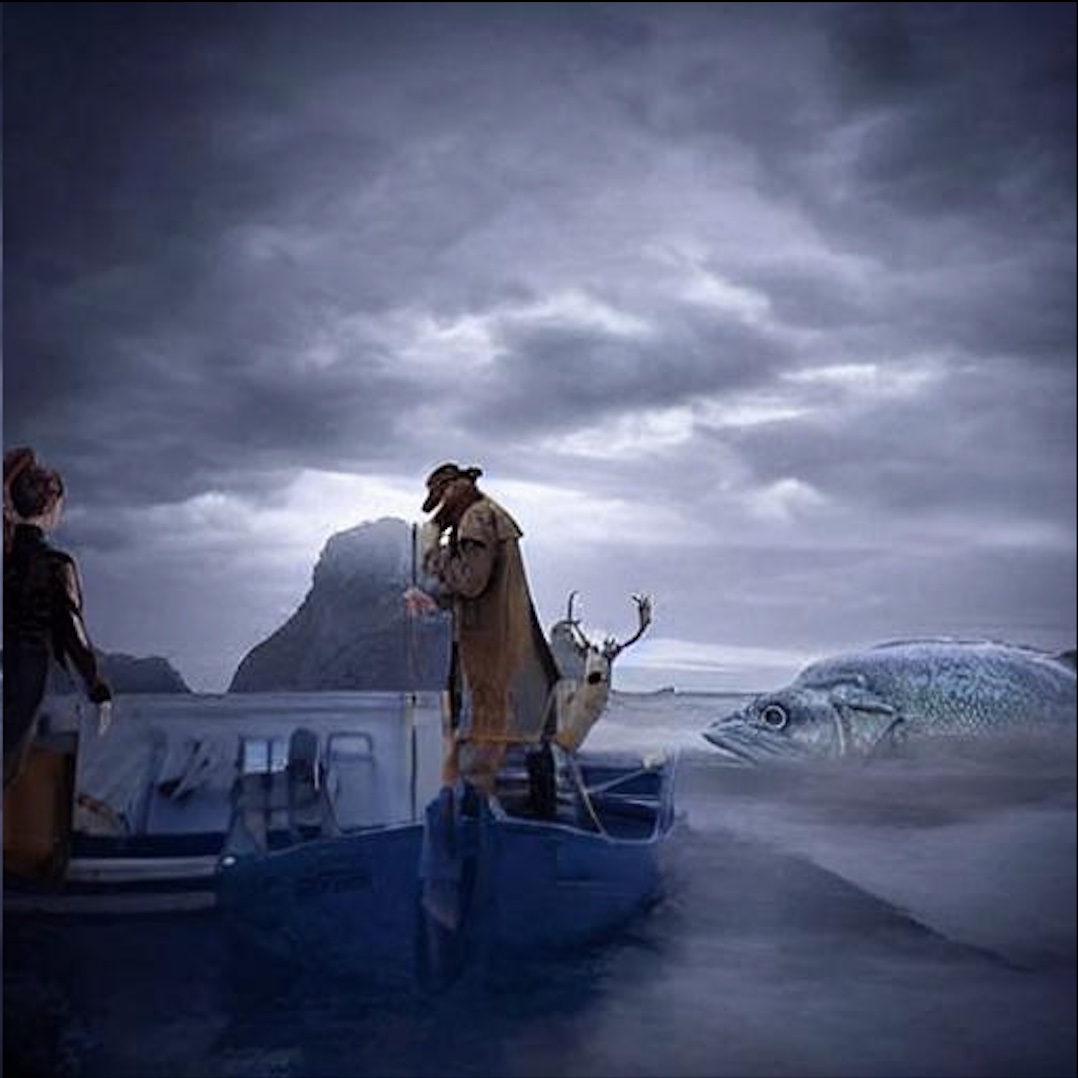} 
& \includegraphics[width=\linewidth,height=2.5cm,keepaspectratio]{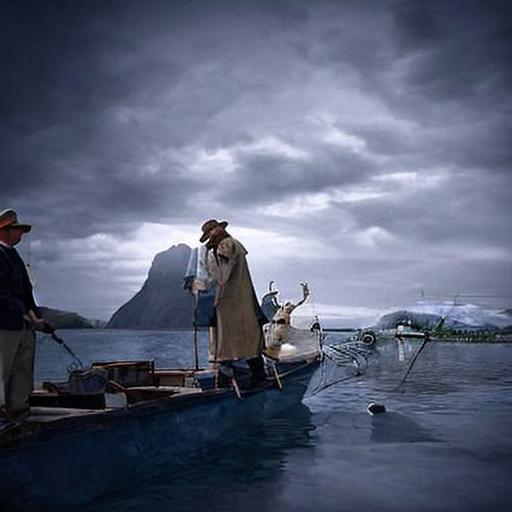} 
& \includegraphics[width=\linewidth,height=2.5cm,keepaspectratio]{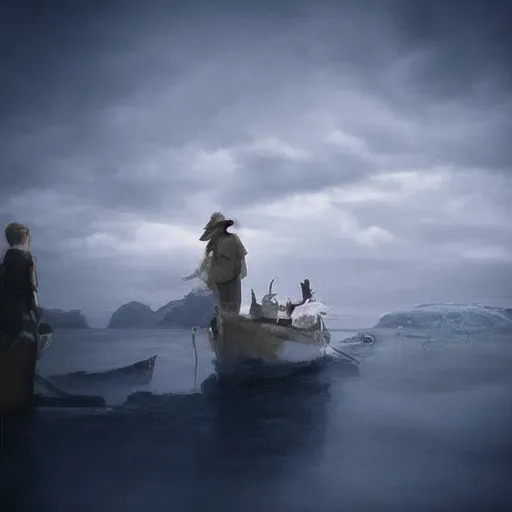} 
& \includegraphics[width=\linewidth,height=2.5cm,keepaspectratio]{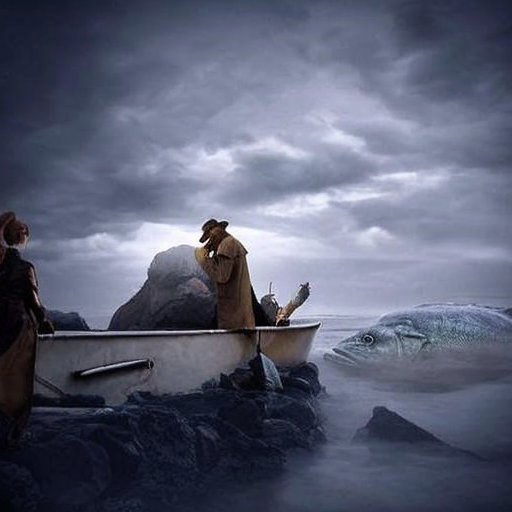}  \\ 

the \textcolor{red}{\sout{sun}} \textbf{moon} over an old farmhouse
& \includegraphics[width=\linewidth,height=2.5cm,keepaspectratio]{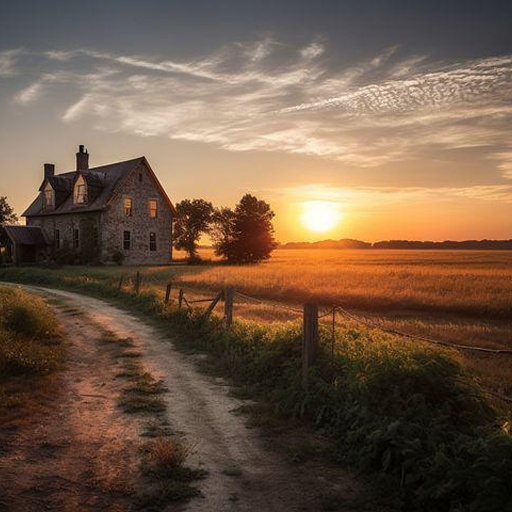} 
& \includegraphics[width=\linewidth,height=2.5cm,keepaspectratio]{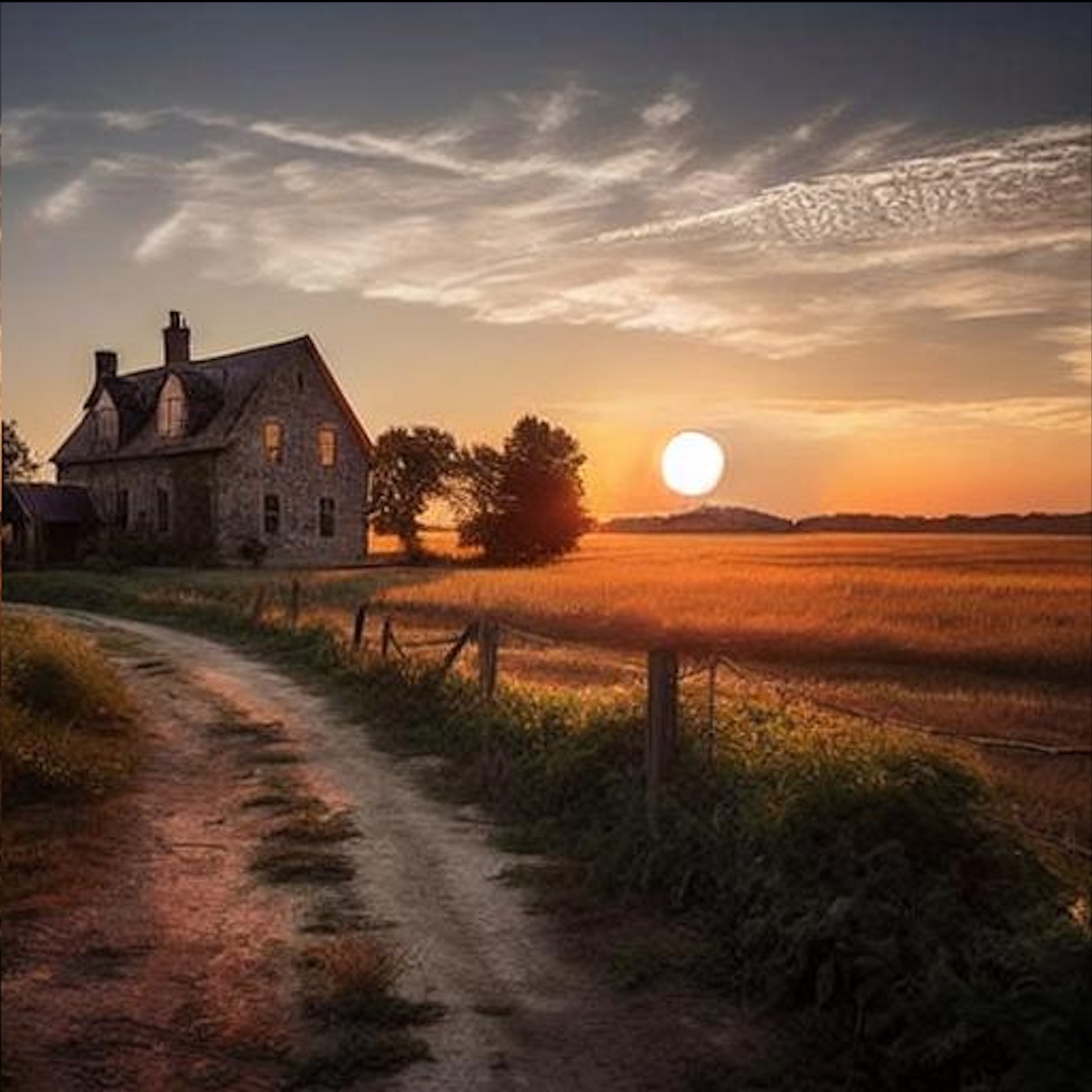} 
& \includegraphics[width=\linewidth,height=2.5cm,keepaspectratio]{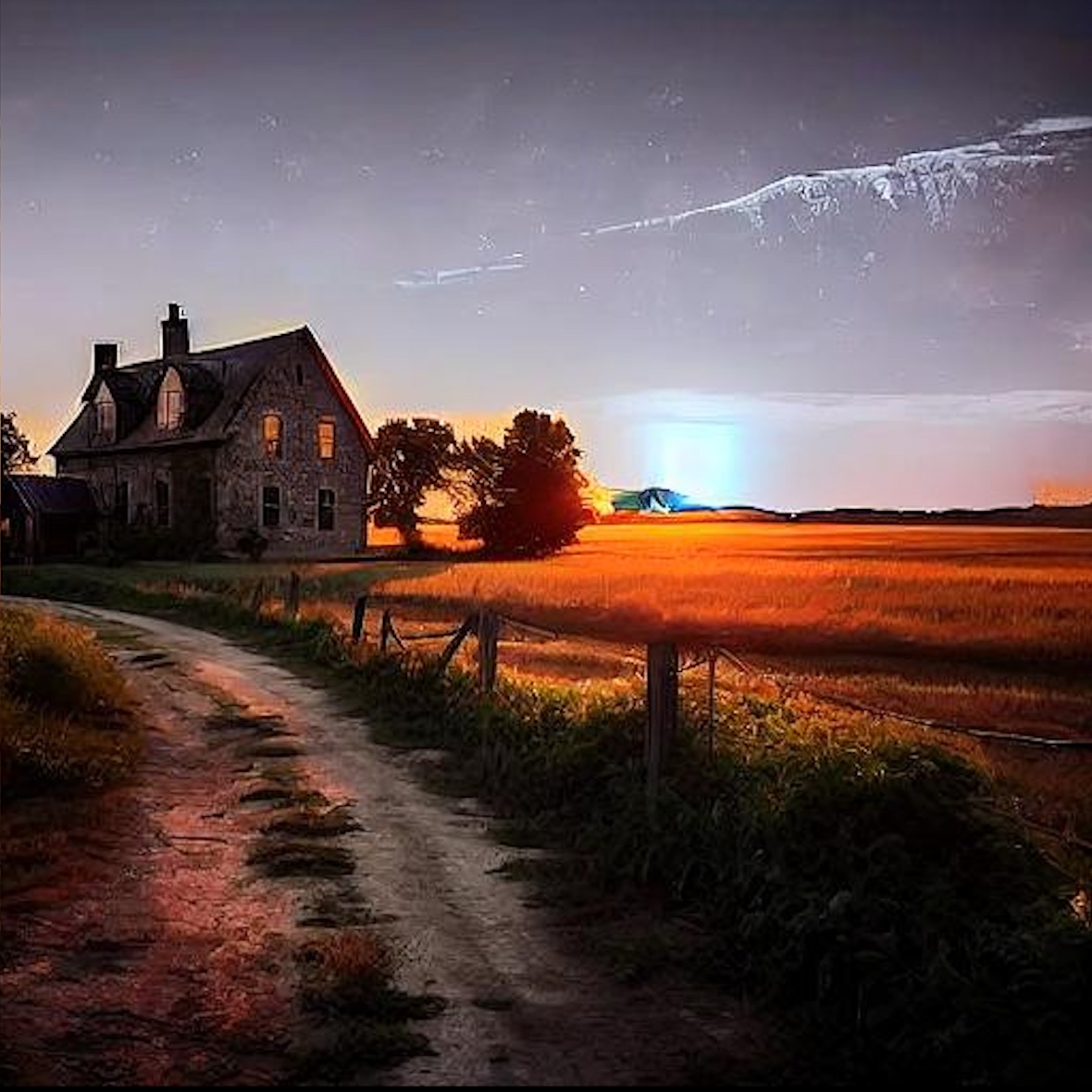} 
& \includegraphics[width=\linewidth,height=2.5cm,keepaspectratio]{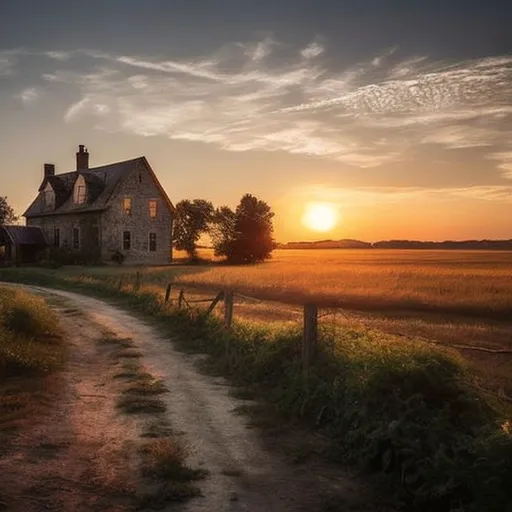} 
& \includegraphics[width=\linewidth,height=2.5cm,keepaspectratio]{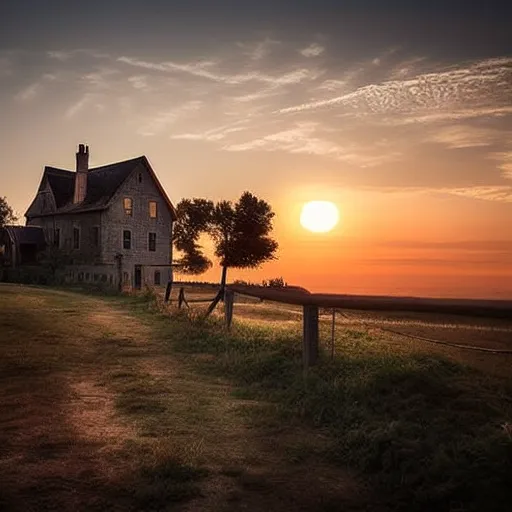} 
& \includegraphics[width=\linewidth,height=2.5cm,keepaspectratio]{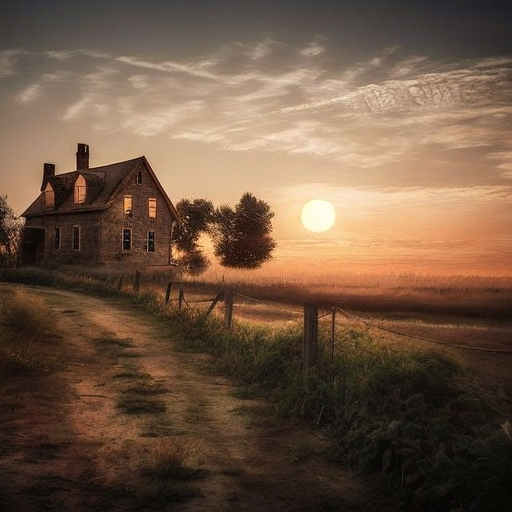}  \\

An asian woman with blue thick-lashed eyes and \textbf{flowers on her} black hair
& \includegraphics[width=\linewidth,height=2.5cm,keepaspectratio]{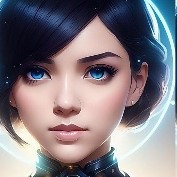} 
& \includegraphics[width=\linewidth,height=2.5cm,keepaspectratio]{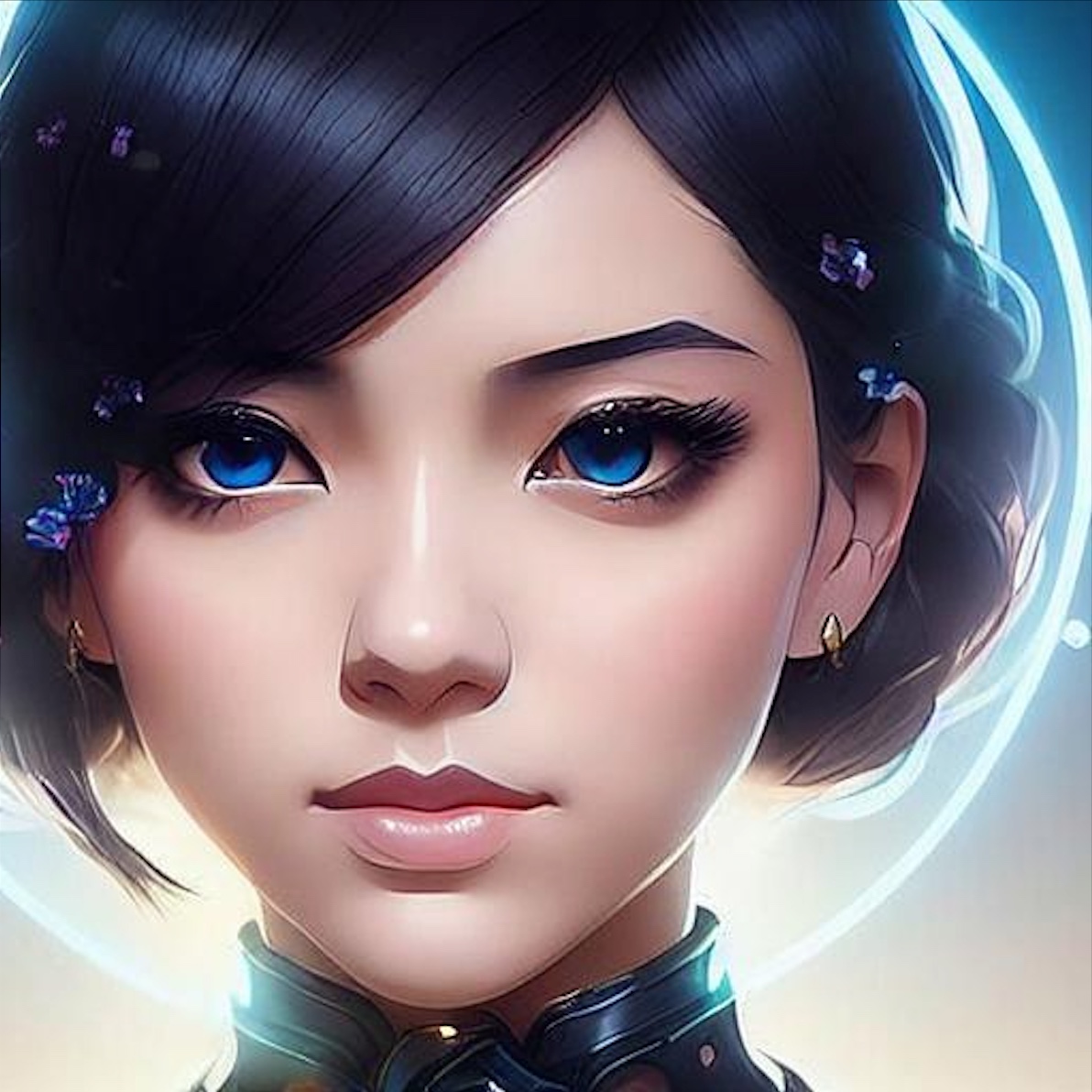} 
& \includegraphics[width=\linewidth,height=2.5cm,keepaspectratio]{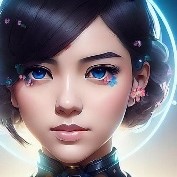} 
& \includegraphics[width=\linewidth,height=2.5cm,keepaspectratio]{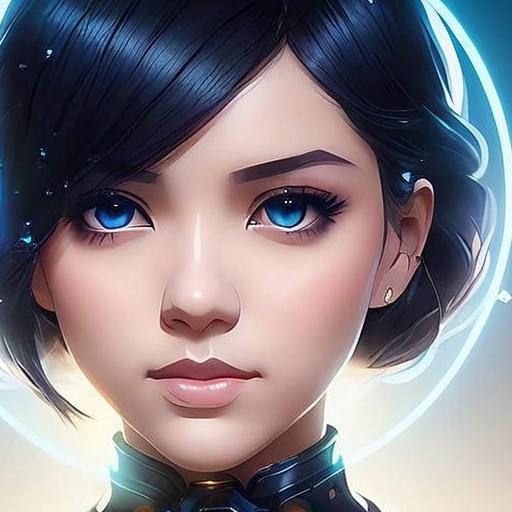} 
& \includegraphics[width=\linewidth,height=2.5cm,keepaspectratio]{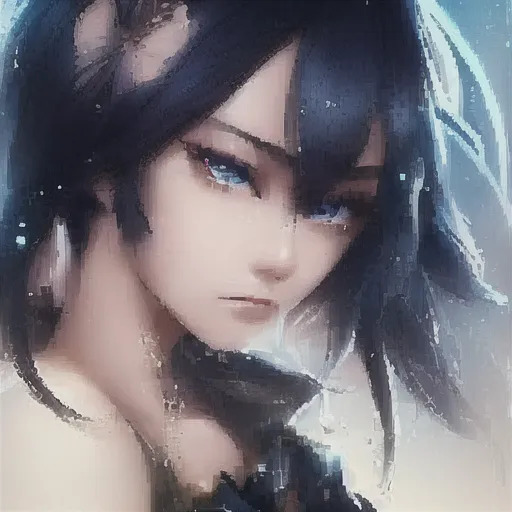} 
& \includegraphics[width=\linewidth,height=2.5cm,keepaspectratio]{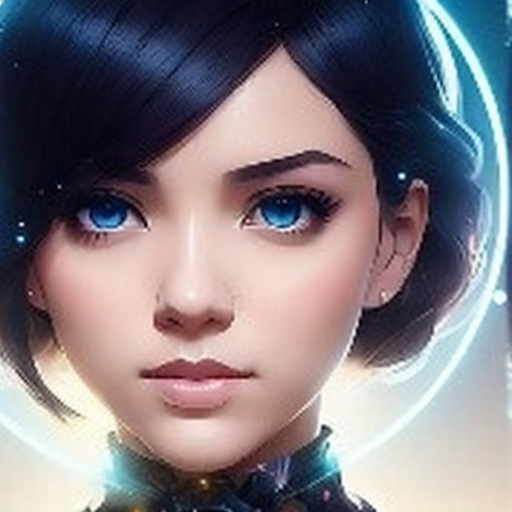}  \\ 

\end{tabular}}

\caption{Additional qualitative results on PIE-Bench}
\label{fig:additional_comp_1}
\end{table*}
\end{onecolumn}

\end{document}